\documentclass[preprint,12pt]{elsarticle}
\usepackage{url}
\usepackage[colorlinks=false, pdfborder={0 0 0}]{hyperref}

\usepackage{amssymb}
\usepackage{amsmath}

\usepackage{booktabs,tabularx}
\usepackage[font=small,labelfont=bf]{caption}
\usepackage{placeins}
\usepackage{dblfloatfix}
\usepackage{caption}
\usepackage{subcaption}
\usepackage{amsmath}
\usepackage{multirow}
\usepackage{verbatim}
\usepackage{float}
\usepackage{hyperref}
\usepackage{url} 
\usepackage{xcolor}
\usepackage[most]{tcolorbox}
\usepackage{ragged2e}
\usepackage{caption}
\usepackage{ulem}
\usepackage{longtable}
\usepackage{booktabs}
\usepackage{hyphenat}

\usepackage{graphicx}

\renewcommand{\sout}[1]{}
\makeatletter
\definecolor{red}{named}{black}
\makeatother

\newtheorem{theorem}{Theorem}

\journal{Applied Energy}
\date{}

\begin{document}

\let\emph\relax 

\begin{frontmatter}



\title{GRAFT: Grid-Aware Load Forecasting with Multi-Source Textual Alignment and Fusion}


\author[affWHU]{Fangzhou Lin\fnref{cofirst}}
\author[affNTU]{Guoshun He\fnref{cofirst}}

\author[affWHU]{Zhenyu Guo\fnref{cosecond}}
\author[affWHU]{Zhe Huang\fnref{cosecond}}

\author[affWHU]{Jinsong Tao\corref{cor1}}

\fntext[cofirst]{These authors contributed equally as co-first authors.}
\fntext[cosecond]{These authors contributed equally as co-second authors.}
\cortext[cor1]{Corresponding author.} 
\ead{jamson_tao@163.com} 
\affiliation[affWHU]{%
  organization={School of Electrical and Automation, Wuhan University},
  city={Wuhan},
  state={Hubei},
  postcode={430072},
  country={China}
}

\affiliation[affUTokyo]{%
  organization={Institute of Industrial Science, The University of Tokyo},
  addressline={4-6-1 Komaba, Meguro-ku},
  city={Tokyo},
  postcode={153-8505},
  country={Japan}
}

\tnotetext[code]{The dataset and code are publicly available at \url{https://github.com/CrimsonFZ/GRAFT-GRid-Aware-Forecasting-with-Text}.}

\begin{abstract}
Electric load is simultaneously affected across multiple time scales by exogenous
factors such as weather and calendar rhythms, sudden events, and policies.
Therefore, this paper proposes GRAFT (GRid-Aware Forecasting with Text), which
modifies and improves STanHOP to better support grid-aware forecasting and
multi-source textual interventions. Specifically, GRAFT strictly aligns
daily-aggregated news, social media, and policy texts with half-hour load, and
realizes text-guided fusion to specific time positions via cross-attention during
both training and rolling forecasting. In addition, GRAFT provides a plug-and-play
external-memory interface to accommodate different information sources in real-world
deployment. We construct and release a unified aligned benchmark covering 2019--2021
for five Australian states (half-hour load, daily-aligned weather/calendar variables,
and three categories of external texts), and conduct systematic, reproducible
evaluations at three scales---hourly, daily, and monthly---under a unified protocol
for comparison across regions, external sources, and time scales. Experimental
results show that GRAFT significantly outperforms strong baselines and reaches or
surpasses the state of the art across multiple regions and forecasting horizons.
Moreover, the model is robust in event-driven scenarios and enables temporal
localization and source-level interpretation of text-to-load effects through
attention read-out. We release the benchmark, preprocessing scripts, and forecasting
results to facilitate standardized empirical evaluation and reproducibility in power
grid load forecasting.
\end{abstract}

\begin{highlights}
\item Introduces \textbf{GRAFT}, a STanHOP-based grid-aware model that fuses multi-source texts for load forecasting
\item Releases an open, synchronized benchmark of load plus news/social/policy texts with code and forecast results
\item Provides the first 3D evaluation across Forecast-Horizon $\times$ Region $\times$ Text-Source with systematic analysis
\end{highlights}

\begin{keyword}



Electric load forecasting \sep Grid-aware modeling \sep Text-guided time series forecasting \sep Multi-source external texts (news/social/policy) \sep Cross-attention fusion \sep External memory

\end{keyword}

\end{frontmatter}



\section{Introduction}
\label{sec:intro}

\subsection{Research Background}
Electric load forecasting is a fundamental support for the safe,
economic, and low-carbon operation of power systems, and
high-accuracy load forecasting is directly related to various
decisions on the operation and market sides~\cite{hong2016pelfforecasting}.
On the one hand, forecasting results are widely used to
formulate preventive maintenance plans, evaluate the operating
status of power generation equipment and the power grid, and
arrange dispatching plans, which helps reduce equipment failure
risks and improve system security margins~\cite{xiao2015hybrid}.
On the other hand, load forecasting is also an
important input for electricity market price modeling and risk
management. By providing forward-looking information for electricity trading,
price settlement, and ancillary service allocation, it can help
dispatchers and market operators develop better action
strategies, reduce power generation costs, and enhance the
economic and social benefits of power grid
operation~\cite{song2005holidayfuzzy}.

Extensive practical experience shows that improving
forecasting accuracy can significantly enhance system economy and
reduce operational risks. For example, according to publicly
available data from the State Grid Corporation of China, it can
be inferred that reducing the forecasting error by 1\%
corresponds to avoiding annual power losses on the scale of tens of
gigawatts~\cite{xiao2015hybrid}. \sout{Meanwhile}\textcolor{red}{Additionally}, load evolution itself
is affected by various random factors such as the level of economic
development, changes in industrial structure, user energy
consumption behavior, meteorological conditions, and unexpected
events, making high-accuracy load forecasting within a given
region still a challenging task. Furthermore, actual dispatching
and resource allocation mostly take place at the regional level
(province/state, city, grid subarea, and even distribution
network nodes), and improvements only at the whole-grid average
level are insufficient to support localized demand response and
cost optimization. Therefore, achieving high-accuracy forecasting
at the subregional scale has direct engineering value and is one
of the core problems addressed in this paper.

From the perspective of time scale, load forecasting is
usually divided into long-term (annual), medium-term (monthly),
short-term (daily/intraday), and ultra-short-term (hourly)
forecasting~\cite{kang2004loadforecast}. Different scales play
different roles in the business chain: the hourly scale focuses
on rolling correction and near-term
dispatching; the daily scale supports intraday generation
planning and power-purchase strategies; the monthly scale
serves maintenance window arrangement and resource
coordination; and annual and longer-term forecasting provide
boundary conditions for power source--grid planning and fuel
planning. Therefore, a multi-scale joint forecasting framework is
needed---one that can simultaneously output hourly, daily, and
monthly forecasts, maintain cross-scale numerical consistency
through aggregation, and allow hourly deviation information to
be fed back in time to correct upstream planning.

\subsection{Research Questions}
In reality, however, the predictability of load is declining due
to multiple external factors: in short-term and ultra-short-term
scenarios, although load exhibits periodicity such as similar-day
and intra-week patterns, it is also strongly driven by sudden
weather changes, social activities, equipment maintenance, and
policy adjustments, showing nonstationary and abrupt
characteristics. In the spatial dimension, different climate
zones, industrial structures, and residential behaviors lead to
significant differences in the magnitude and lag time of
responses to the same external shock across regions. In the
temporal dimension, the influence mechanisms of the same external
source also differ across scales (e.g., weather affects hourly
patterns more, while policies affect monthly trends more). This
indicates that, to obtain higher forecasting accuracy, relying
only on historical load and structured meteorological/calendar
variables is often insufficient. It is urgently necessary to
systematically introduce external textual information capable of
characterizing event-driven and behavioral rhythms (such as news,
social media, and policy texts) and strictly align it with load
in both time and
region~\cite{reisfilho2022enrichment,li2019crudeoil,xu2024tgtsf}.

There are generally three reasons why current methods struggle to
fully benefit from external text. First, external information is
often assumed to be structured and statistically stable, lacking
mechanisms for acquiring, filtering, and strictly time-aligning
unstructured text, which leads to mismatches between the time and
position when the text influences the load. Second, most fusion
approaches remain at static concatenation or global weighting,
making it difficult to answer when, where, and with what
intensity the text affects curve patterns and peak--valley
positions, resulting in insufficient interpretability. Third, due
to the lack of unified data processing, information
categorization, and metric definitions, existing work cannot
conduct reproducible and systematic comparisons along the three
dimensions of ``region $\times$ external source $\times$ time
scale,'' making it impossible to reliably estimate the magnitude
of gains brought by external information and its applicable
scope~\cite{wang2024fromnews,wu2025dualforecaster,yao2025contextawareprobabilisticmodelingllm}.

Based on the above motivations, it is necessary to propose a
forecasting framework that can simultaneously cover subregions,
multiple time periods, and multiple external information sources.
On the one hand, it should characterize heterogeneous responses
at the regional granularity and form transferable localized
models; on the other hand, it should uniformly model the coupling
and constraints among hourly, daily, and monthly scales.
Meanwhile, at the external information level, it should achieve
strict temporal alignment and position-related fusion between
news, social media, and policy texts and the load, together with
interpretable impact attribution and a unified evaluation
protocol. Only with such a design can stable and reproducible
accuracy gains be achieved in real power grid scenarios.

\subsection{Research Status}
In recent years, research has generally evolved along three
directions:
\begin{enumerate}
  \renewcommand{\labelenumi}{(\roman{enumi})}
  \item \textbf{Numerical statistical and machine learning
  methods:} approaches such as ARIMA/ARIMAX, linear/ridge
  regression, support vector machines, and random forests are
  combined with structured exogenous variables such as weather
  and calendar, providing interpretable baseline performance but
  often remaining limited under strong nonlinearity, long-term
  dependence, and concept drift
  scenarios~\cite{hong2016pelfforecasting,torres2021deeptsf,kong2025deeptsf}.
  \item \textbf{Deep learning methods:} models based on LSTM/GRU,
  CNN/TCN, and attention-based Transformers perform well in
  short-term and intraday load forecasting, as they can learn
  nonlinear and multi-scale dependencies in an end-to-end manner
  and easily integrate multi-source structured
  features~\cite{torres2021deeptsf,kong2025deeptsf,bedi2019deeplearning}.
  \item \textbf{Decomposition--reconstruction paradigm:} this
  line of work has gradually become a common approach for
  handling nonstationarity. Signal decomposition tools such as
  EMD/EEMD/CEEMDAN and VMD decompose the original sequence into
  intrinsic mode functions (IMFs) of different frequency bands;
  each IMF is then modeled by different submodels and
  reconstructed with weights, aiming to reduce noise, alleviate
  mode mixing, and improve learnability. This idea has been
  empirically validated as effective in energy systems and
  integrated energy scenarios, especially in wind power and
  integrated energy load forecasting
  tasks~\cite{SHI2024122146,LI2024123283,zhao2016windforecast}.
\end{enumerate}

It should be noted that the above directions generally assume
that external information is structured and relatively stable,
meaning that exogenous variables enter the model in the form of
numerical or categorical features and are assumed to have
approximately consistent statistical properties during both
training and inference. This assumption usually holds in
weather/calendar-dominated situations, but when external
disturbances appear in the form of events and policies as
unstructured text and exhibit regional differences across
different time scales, relying solely on existing assumptions
becomes difficult for effective modeling. On the one hand, the
semantics, timeliness, and intensity of textual signals cannot be
sufficiently expressed by fixed features; on the other hand, text
and load are not synchronized in their temporal reference, making
precise alignment and impact localization more
challenging~\cite{reisfilho2022enrichment,li2019crudeoil,xu2024tgtsf}.

Based on the above understanding, in addition to structured
variables such as weather and calendar, unstructured external
information such as news, social media, and policy texts should
also be systematically incorporated, because they contain key
signals such as event-driven impacts, behavioral rhythms, and
institutional shocks and hold clear potential for performance
gains. Therefore, the shift from ``structured exogenous variable
fusion'' to ``text-based unstructured exogenous variable fusion''
is not only a change in feature form, but also requires
establishing systematic mechanisms in the following aspects:
\begin{enumerate}
  \renewcommand{\labelenumi}{(\roman{enumi})}
  \item strict and consistent temporal alignment;
  \item semantic relevance retrieval and filtering for the
  forecasting horizon;
  \item dynamic injection and gating that vary with temporal
  position;
  \item interpretability evaluation and attribution analysis.
\end{enumerate}

Compared with the above objectives, recent approaches based on
large language models mostly incorporate news texts directly into
the forecasting process through prompts or instructions. Related
studies have shown that appropriately introducing news can make
the forecasting trajectory closer to the actual measurements in
certain periods, but several limitations have also been exposed:
first, the method relies on prior text filtering and association,
and biases in filtering or inference can easily propagate to the
forecasting results; second, the context length constraints of
pretrained models make long windows, cross-region processing, or
parallel processing of multiple sequences subject to truncation
and performance degradation risks; third, treating the problem
simply as numerical regression and performing prompt-based
concatenation makes it difficult to achieve alignment and fusion
at the positional level, and therefore cannot clearly answer
when, where, and with what intensity the text influences the
load~\cite{wang2024fromnews,wu2025dualforecaster,yao2025contextawareprobabilisticmodelingllm}.

Overall, although existing studies have incorporated news topics,
sentiment, or word embeddings as additional features for
demand/price forecasting and have shown certain gains in
describing intraday patterns and peaks under event-driven
scenarios, current practices still have three critical gaps:
\begin{enumerate}
  \renewcommand{\labelenumi}{(\roman{enumi})}
  \item \sout{The acquisition--alignment--injection pipeline is
  structurally incomplete: effective text filtering for the
  forecasting horizon is lacking, inconsistent temporal
  references lead to cumulative alignment errors, and fusion
  mechanisms that reflect positional differences are absent;}
  \textcolor{red}{The acquisition--alignment--injection pipeline is
  structurally incomplete: \textbf{(acquisition)} effective text filtering and provenance control for the target horizon are often missing, \textbf{(alignment)} inconsistent temporal/region anchors can cause cumulative misalignment, and \textbf{(injection)} position-aware fusion mechanisms that reflect intra-day differences are frequently absent;}
  \item Fusion and interpretability are insufficient: early
  concatenation or static weighting makes it difficult to locate
  the timing and intensity of textual influence and to provide
  verifiable attribution evidence;
  \item Unified evaluation and reproducibility are lacking:
  systematic comparison and statistical significance testing
  under a unified protocol covering ``region $\times$ external
  source $\times$ time scale'' are missing, and an integrated
  open benchmark combining strictly aligned data, code, and
  results is still rare.
\end{enumerate}

\subsection{Research Goal}
\sout{To address the incompleteness of the acquisition--alignment--injection
pipeline as well as the limitations in fusion mechanisms and
interpretability, this paper proposes GRAFT (GRid-Aware
Forecasting with Text).}
\textcolor{red}{To address the incompleteness of the acquisition--alignment--injection
pipeline and the limitations in fusion and interpretability, this paper proposes GRAFT (GRid-Aware
Forecasting with Text), and \textbf{operationalizes} the three stages as follows: \textbf{(i) acquisition} collects public News/Reddit/Policy documents over 2019--2021 and constructs a day-level corpus for each state; \textbf{(ii) alignment} maps texts to the ``date $\times$ region'' anchor at the end of the input window and uses explicit masking for missing or delayed entries; and \textbf{(iii) injection} performs position-dependent fusion by retrieving and injecting text cues into each half-hour position through cross-attention in the forecasting window.}

Building upon and extending STanHOP for
grid-aware load forecasting, GRAFT constructs daily-aggregated news,
social media, and policy texts as external memory representations.
At the end of the forecasting window, cross-attention is applied
between the external memory and the half-hour load representations
to achieve position-related alignment and fusion~\cite{wu2023stanhopsparsetandemhopfield}.

On the text side, Sentence-BERT (SBERT) is used for semantic
encoding: multiple texts within the same calendar day are first
individually encoded into fixed-length vectors and normalized,
and then aggregated by weighting according to timestamps and
sources to obtain the daily text vector. All available dates are
arranged in chronological order to form the external memory
matrix, and are aligned with the anchor date at the end of the
window. For missing or delayed dates, masking is applied to
prevent pseudo-signals from entering the attention
readout~\cite{reimers2019sbert}.

The differences and innovations of GRAFT relative to traditional
STanHOP and other mainstream load forecasting models (including
statistical/machine learning, LSTM/GRU/TCN, Transformer, static
text concatenation, and LLM prompt-based injection) are
summarized as follows:
\begin{enumerate}
  \renewcommand{\labelenumi}{(\roman{enumi})}
  \item \textbf{Plug-and-play external memory interface:} news,
  social media, and policy texts are aggregated on a daily basis
  and vectorized as memory units, strictly aligned with load by
  ``date $\times$ region''; additions, missing entries, and
  delays are uniformly handled by masking.
  \item \textbf{Position-aware cross-attention fusion:} for each
  time position in the forecasting window, the load features
  serve as queries to dynamically retrieve and weight the most
  relevant textual cues from the external memory, and inject
  them in a residual manner to finely adjust curve shapes and
  peak--valley timing.
  \item \textbf{Multi-source multi-scale comparison and
  interpretability:} the same interface supports multiple
  external sources working consistently at the hourly, daily,
  and monthly scales; attention weights and source contributions
  are directly comparable.
\end{enumerate}

These points enable GRAFT, compared with traditional STanHOP
(which models only time series and struggles to leverage textual
event signals) and mainstream fusion approaches (static
concatenation, prompt-based injection, and insufficient alignment
and localization capability), to exhibit stronger sensitivity,
robustness, and interpretability in event-driven scenarios, while
maintaining plug-and-play applicability and reproducible
evaluation in engineering practice.

To translate the above design principles into an operational,
reproducible, and engineering-friendly research paradigm, this
paper develops contributions along three dimensions: methods and
framework, data and evaluation, and empirical analysis and
interpretation:
\begin{enumerate}
  \renewcommand{\labelenumi}{(\roman{enumi})}
  \item \textbf{GRAFT model:} proposes a grid-aware text-guided
  load forecasting framework based on STanHOP, which strictly
  aligns news, social media, and policy texts with half-hourly
  load data by ``date--region,'' and performs multi-source
  fusion through a position-dependent strategy.
  \item \textbf{Open-source data and parameters:} releases an
  aligned open benchmark (half-hourly load + news/social/policy
  texts), together with data processing scripts and forecasting
  results, forming an end-to-end reproducible experimental
  pipeline to facilitate subsequent comparison and extension.
  \item \textbf{Three-dimensional evaluation framework:} for the
  first time, establishes a unified evaluation protocol and
  systematic analysis along the ``Forecast-Horizon $\times$
  Region $\times$ Text-Source'' dimensions, enabling fair
  comparison and robust conclusions across regions, external
  sources, and scales.
\end{enumerate}

\section{Data and data processing}
\label{sec:data}

This study constructs and preprocesses a synchronized dataset
spanning multiple regions, multiple information sources, and
multiple time scales, covering the five Australian states from
2019--2021 (New South Wales, NSW; Queensland, QLD; South
Australia, SA; Tasmania, TAS; Victoria, VIC), where NSW contains
the Australian Capital Territory, ACT~\cite{aemo_nem_map,aemo_data_nem}.
The main time axis is the state-level total load (30-minute);
structured exogenous variables include daily weather and calendar
information; on this basis, three types of external textual
information sources are incorporated: News, Reddit, and Policy.
All information sources are strictly aligned with the load data
along the ``date--region'' dimensions, and undergo unified
cleaning, normalization, and temporal alignment before being fed
into the models, enabling consistent evaluation across hourly,
daily, and monthly forecasting horizons.

To ensure evaluation comparability and prevent information
leakage, samples are divided according to the forecast end date
($T_{end}$): the training set spans 2019-01-01 to 2020-09-30; the
validation set spans 2020-10-01 to 2020-12-31; and the test set
spans 2021-01-01 to 2021-12-31. Sample allocation is determined
by the date of $T_{end}$ to ensure no duplication of samples.
When constructing validation/test samples, their input windows may
reference earlier historical observations, but those observations
are not included in training to avoid information leakage. All
normalization and standardization parameters are fitted only on
the training set; \sout{model selection and hyperparameter tuning are
conducted on the validation set, while the test set is used
solely for reporting final results.}
\textcolor{red}{the validation set is used only for offline monitoring and diagnostic analysis (and is not used for early stopping, model selection, or hyperparameter tuning), while the test set is used solely for reporting final results.}

\subsection{Power load data with weather/calendar}
\label{subsec:load_weather_calendar}

\textbf{Load data.}
The state-level load data with a 30-minute resolution are sourced
from the Australian Energy Market Operator (AEMO) for the period
2019--2021~\cite{aemo_data_nem}. Preprocessing includes:
identification and light correction of missing and anomalous
points; unification to Australian Eastern Standard Time and
Australian Eastern Daylight Time (AEST/AEDT); construction of an
aligned backbone time axis; and de-duplication and sorting of
timestamps. These steps yield an aligned National Electricity
Market (NEM) load dataset. Such publicly available NEM load data
have been widely used in recent hybrid decomposition and deep
learning--based load forecasting
studies~\cite{SHI2024122146,LI2024123283}.\\

\textbf{Weather and calendar.}
Weather variables are collected at a daily frequency (temperature,
humidity, air pressure, wind speed, precipitation, etc.), and the
calendar includes working-day and public-holiday indicators as
well as seasonal tags. Daily values are mapped to the 30-minute
resolution via same-day broadcasting, while ensuring consistency
across years and daylight-saving transitions. This configuration
of structured exogenous variables---primarily weather and
calendar---follows common practice in traditional load forecasting
studies~\cite{hong2016pelfforecasting}.

\begin{figure*}[p]
  \centering
  \includegraphics[width=\textwidth,height=0.85\textheight,keepaspectratio]{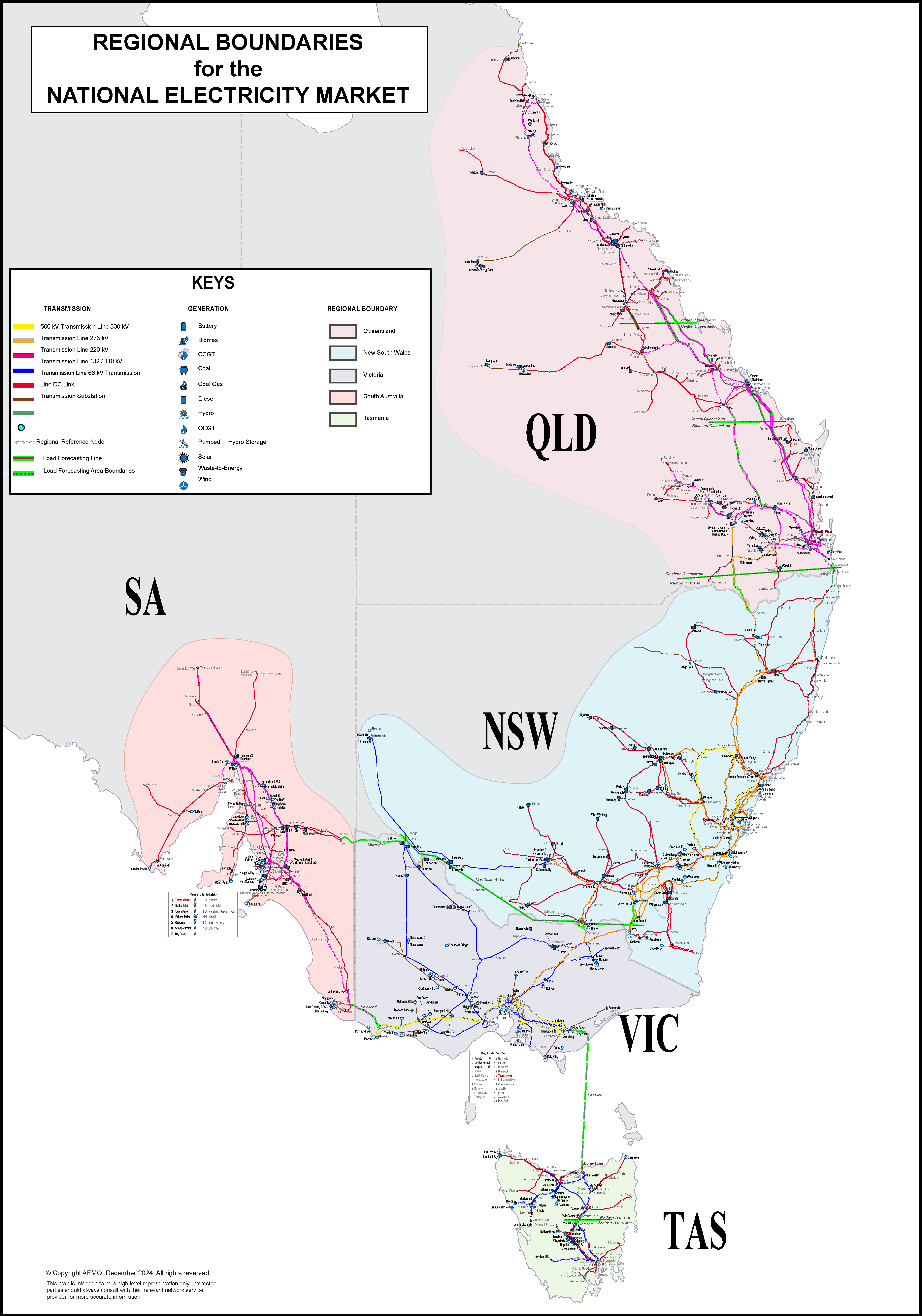}
  \caption{Regional boundaries, major interconnectors, and
  representative generation mix in the National Electricity
  Market (NEM) (schematic map from AEMO~\cite{aemo_nem_map}).}
  \label{fig:nem-map}
\end{figure*}

\subsubsection{Regional load characteristics}
\label{subsubsec:regional_patterns}

To facilitate understanding of the coupling between regional load
patterns and grid structure, Figure~\ref{fig:nem-map} presents an
illustration of the regional boundaries, major transmission
corridors, and representative power supply distribution of the
Australian National Electricity Market (NEM), as officially
released by AEMO. The figure shows the five settlement regions
(QLD, NSW, VIC, SA, TAS). The main interconnectors across states
include the Queensland--New South Wales Interconnector (QNI),
the Victoria--New South Wales Interconnector (VNI), the Heywood
interconnector (AC), Murraylink (high-voltage direct current,
HVDC), and Basslink (HVDC). It also shows the spatial
distribution of transmission networks of different voltage levels
(66/110/132/220/275/330/500~kV), together with wind, solar, and
hydro generation, coal- and gas-fired units, pumped storage, and
battery energy storage systems~\cite{aemo_nem_map}.

From a spatial perspective, the QLD--NSW--VIC east-coast corridor
is connected by a backbone of 330/500~kV transmission lines. SA
is structured around a 275~kV network and is linked to VIC via
the two interconnectors Heywood and Murraylink, while TAS is
connected to the mainland through Basslink
(Figure~\ref{fig:nem-map}). This strongly coupled cross-state
network provides the physical basis for inter-regional balancing
under heatwaves or cold spells. For example, during heatwaves in
QLD and NSW, QNI and VNI support north--south power transfers;
whereas on clear spring or summer midday periods in SA, the
uplift of distributed and utility-scale PV generation often leads
Heywood/Murraylink to be in a net-exporting state, before
reverting to net-importing in the evening as solar irradiance
declines. This pattern aligns with the previously observed
``midday deep trough--steep evening peak'' load
profile~\cite{aemo_msl_sa}.

In terms of generation mix, TAS is dominated by hydropower
supplemented by pumped storage and wind power, which determines
its ``winter peak--summer trough'' seasonal load pattern and
seasonal exchanges with VIC. SA and western VIC have a high
density of wind farms with multiple battery storage facilities,
which explains their sensitivity to daytime renewable output
fluctuations and inter-regional support. NSW/QLD/VIC exhibit a
more diversified structure consisting of coal, gas, and increasing
shares of wind and solar, with generating units located along
coal bases and load centers being tightly coupled with the
330/500~kV backbone grid, facilitating large-scale power flow
reconfiguration under extreme weather conditions. Together, these
topologies and generation layouts shape the differences across
states in seasonal amplitude, intraday patterns, and weekly
rhythms, and provide the basis for subsequently introducing
external variables such as ``interconnector available
capacity/maintenance information'' and ``regional wind--solar
output proxies.''

\begin{figure*}[!htbp]
  \centering
  \setlength{\tabcolsep}{6pt} 
  \renewcommand{\arraystretch}{1.0}

  \begin{tabular}{cc}
    \begin{subfigure}{0.48\textwidth}
      \centering
      \includegraphics[width=\linewidth]{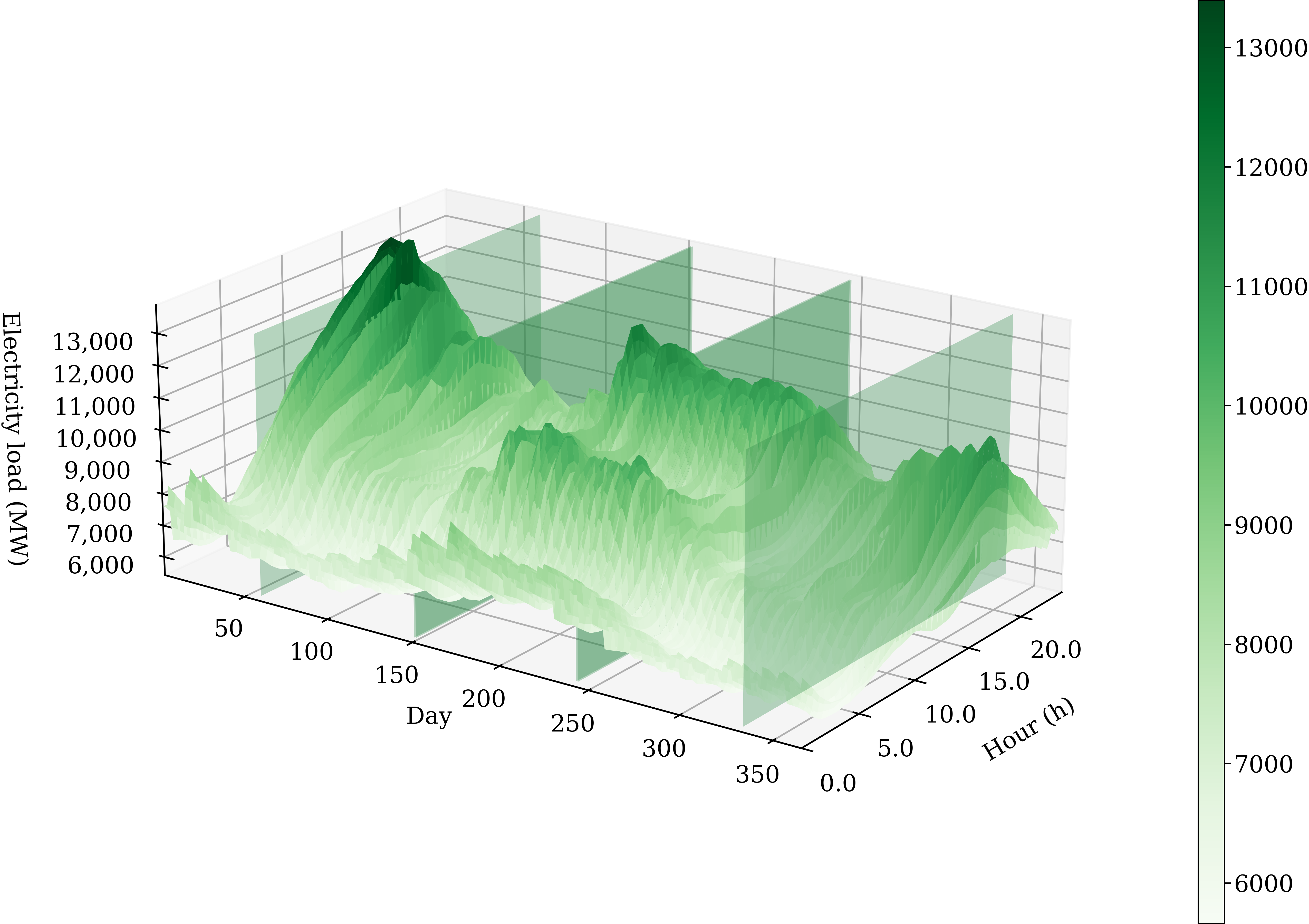}
      \caption{NSW (including ACT)}
    \end{subfigure}
    &
    \begin{subfigure}{0.48\textwidth}
      \centering
      \includegraphics[width=\linewidth]{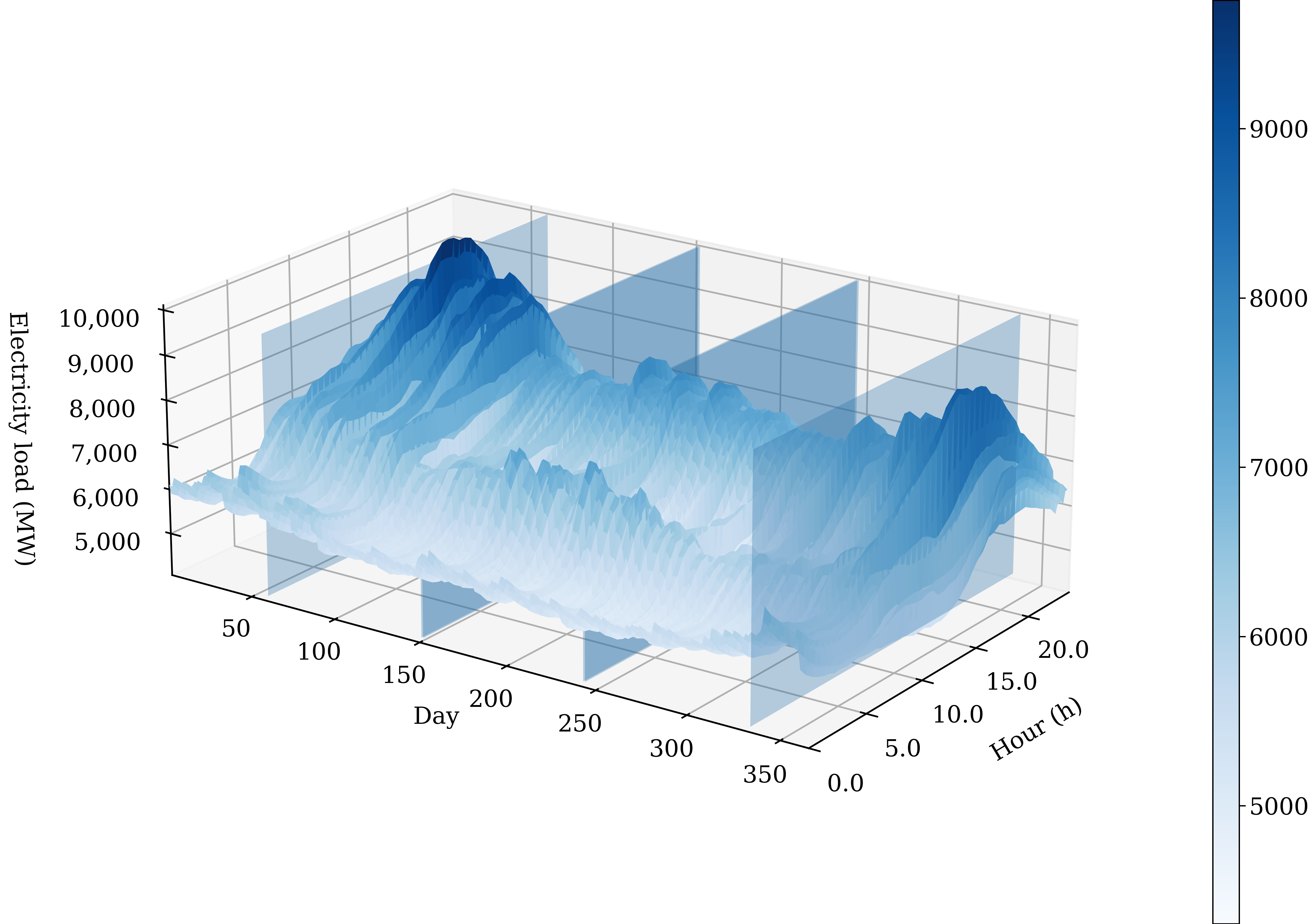}
      \caption{QLD}
    \end{subfigure}
    \\[4pt]

    \begin{subfigure}{0.48\textwidth}
      \centering
      \includegraphics[width=\linewidth]{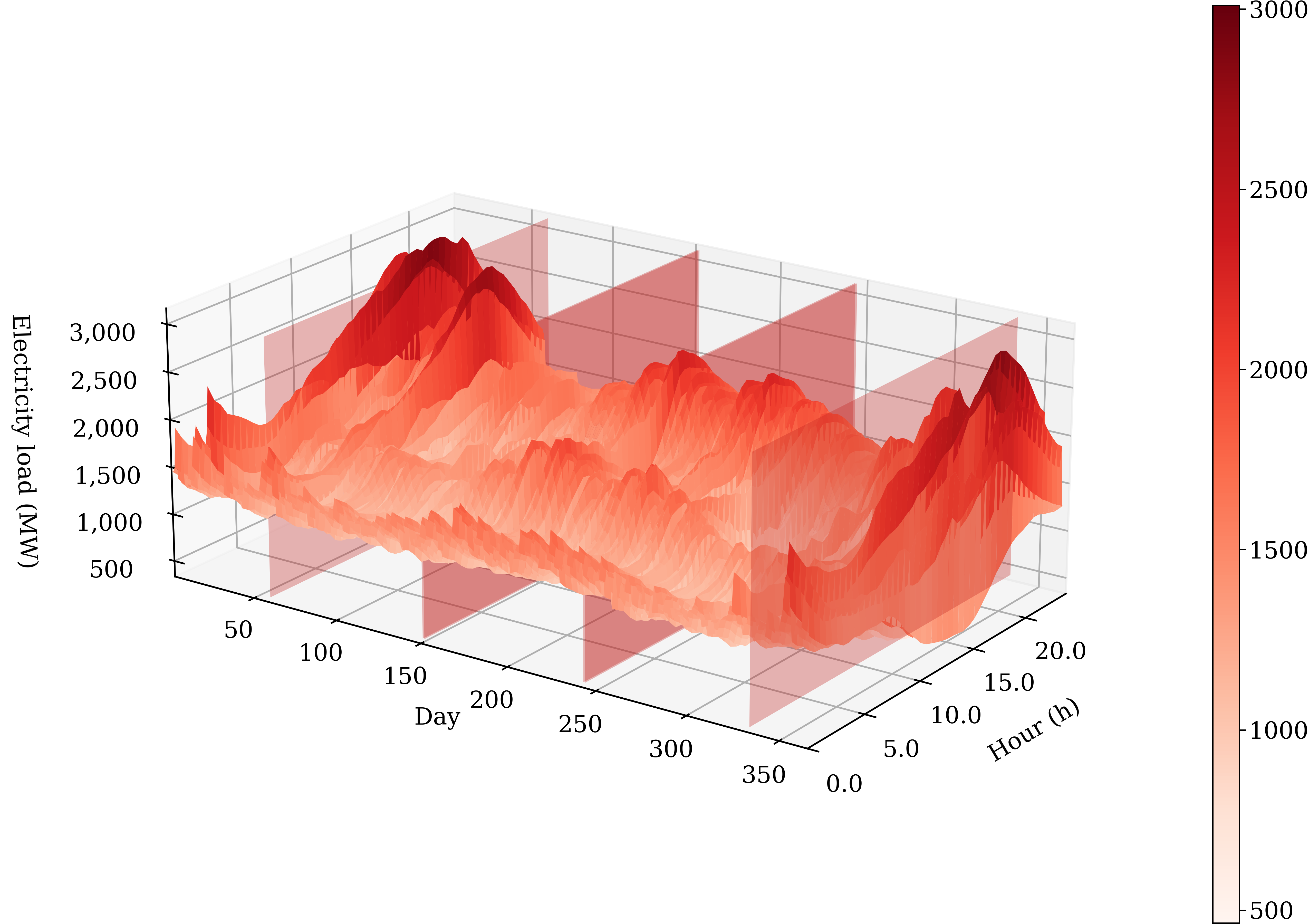}
      \caption{SA}
    \end{subfigure}
    &
    \begin{subfigure}{0.48\textwidth}
      \centering
      \includegraphics[width=\linewidth]{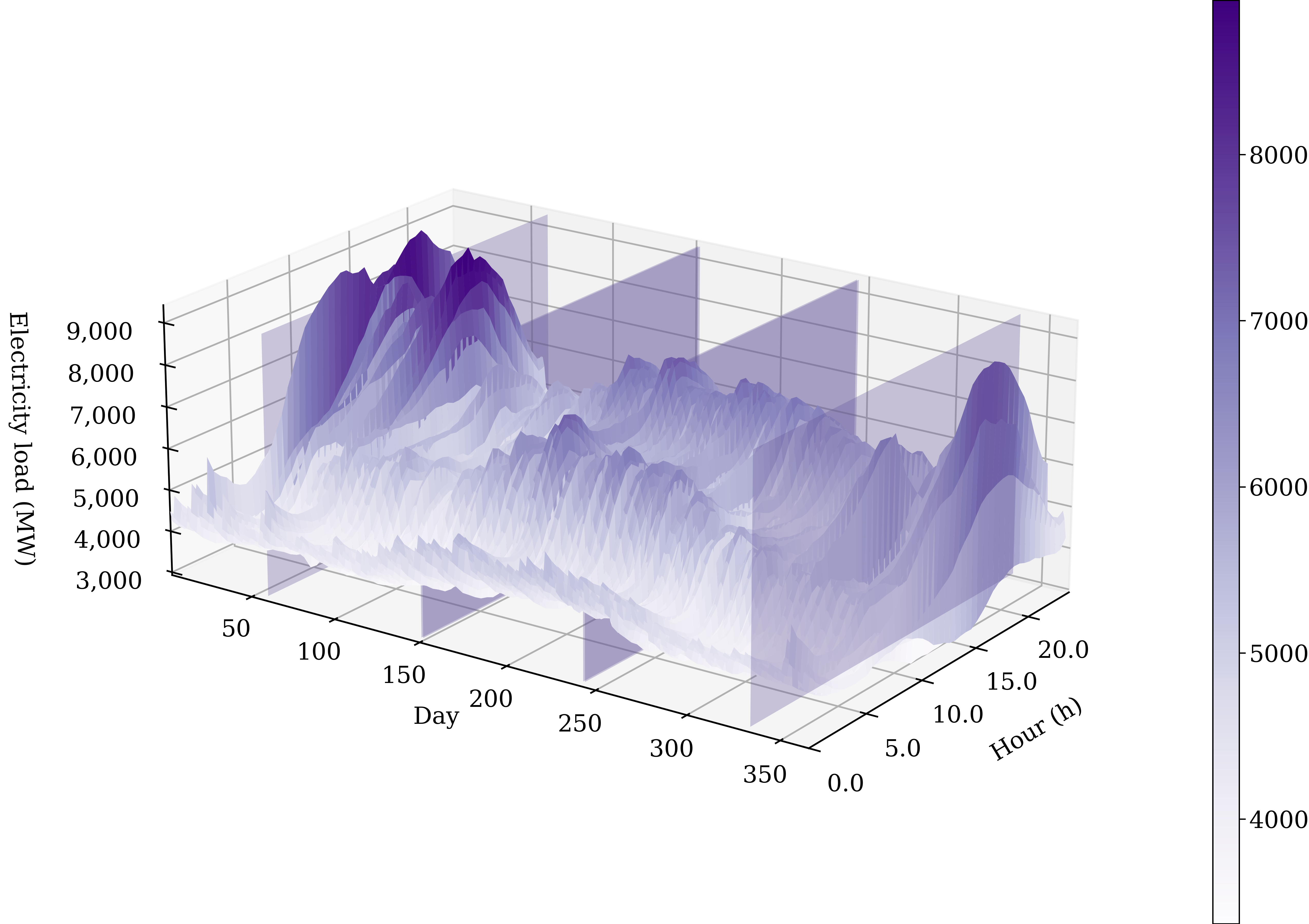}
      \caption{VIC}
    \end{subfigure}
    \\
  \end{tabular}

  \vspace{6pt}

  \begin{subfigure}{0.60\textwidth}
    \centering
    \includegraphics[width=\linewidth]{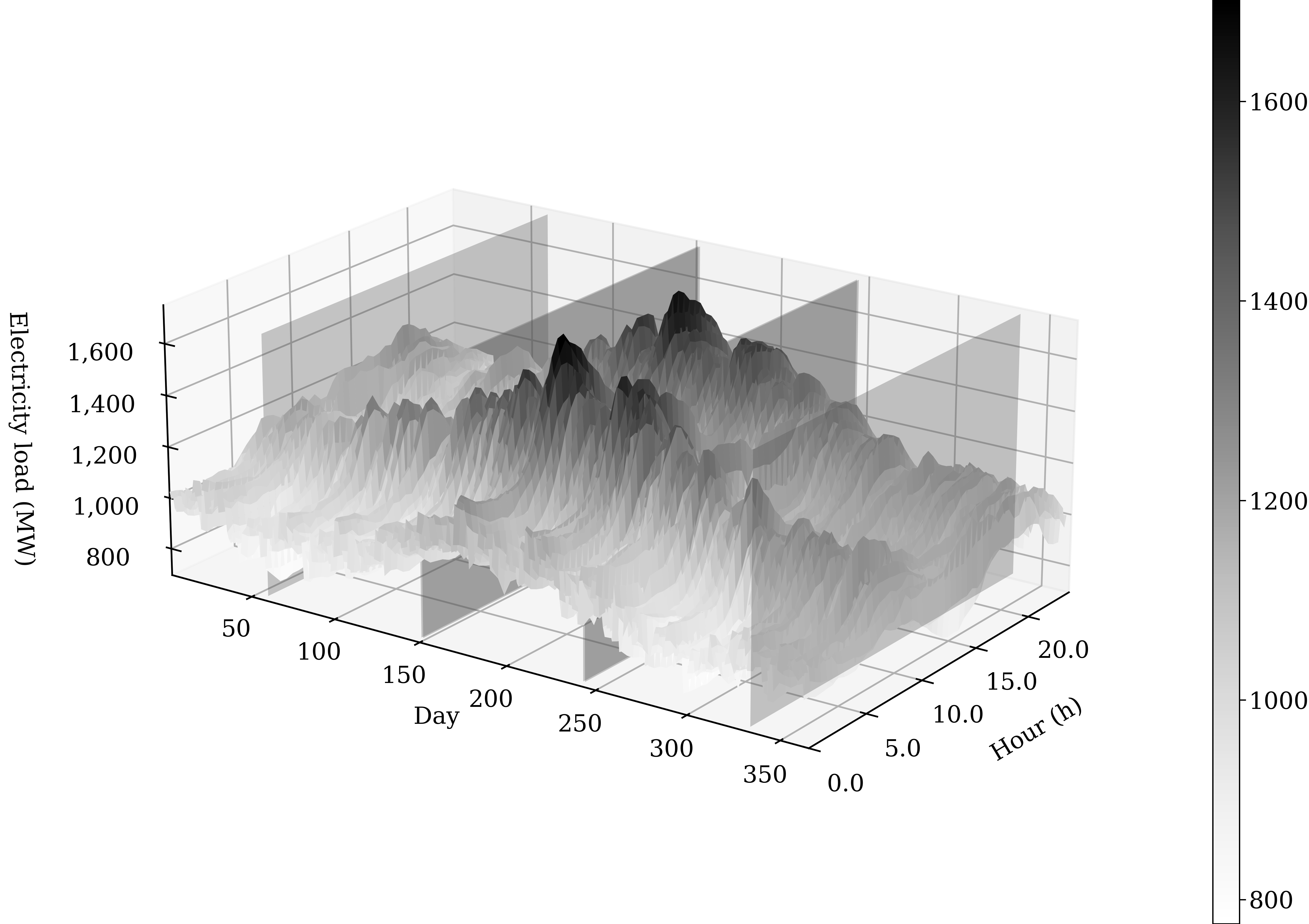}
    \caption{TAS}
  \end{subfigure}

  \caption{Annual electrical load surfaces in 2019 for NEM states.}
  \label{fig:nem-2019-surfaces}
\end{figure*}

To visually illustrate regional differences,
Figure~\ref{fig:nem-2019-surfaces} presents the three-dimensional
``Intraday (Hour)--Full Year (Day)--Load Value'' surface plots for
the five states in 2019, with seasonal backdrops
(Spring/Summer/Fall/Winter) indicating seasonality. Based on this
figure together with the grid structure facts, each state is
analyzed individually, followed by an integrated comparison and a
summary of underlying causes.

\paragraph{NSW (New South Wales)}
\begin{enumerate}
  \renewcommand{\labelenumi}{(\roman{enumi})}
  \item Seasonal characteristics: the annual load amplitude is large,
  with both summer heat and winter heating demand triggering
  pronounced peaks. The ``shoulder seasons'' of spring and autumn
  are relatively stable, but secondary peaks caused by alternating
  warm and cold spells remain visible.
  \item Intraday characteristics: on weekdays, the load exhibits a
  ``morning--evening dual-peak'' pattern, with the morning peak
  synchronized with employment/commuting activities and a sharper
  evening peak. On weekends, the peaks are blunter and shift
  downward overall. On extremely hot days, the evening peak
  increases substantially.
  \item Cause analysis: the temperate coastal climate combined with
  a highly concentrated population and industry (Sydney metropolitan
  area) results in a high share of residential + commercial demand.
  Rooftop photovoltaics (Rooftop PV) continue to rise, but their
  peak-shaving effect is weaker compared with SA. QNI with QLD and
  VNI with VIC form the principal interconnectors, and cross-regional
  balancing under peak and abnormal operating conditions enhances
  supply--demand flexibility.
\end{enumerate}

\paragraph{QLD (Queensland)}

\begin{enumerate}
\renewcommand{\labelenumi}{(\roman{enumi})}
\item Seasonal characteristics: influenced by subtropical/tropical
climate, cooling demand dominates in summer, leading to prominent
summer peaks with a long persistence within the season; winter
peaks are comparatively mild.
\item Intraday characteristics: on hot days, the load exhibits a
``late\hyp{}afternoon\hyp{}evening single-peak uplift,'' with evening
air-conditioning and lighting jointly forming a high ridge. The
difference between weekdays and weekends is partially ``washed
out'' during prolonged heatwaves, and the peak timing remains
relatively stable.
\item Cause analysis: higher temperature and humidity levels lead
to a stronger temperature--demand elasticity; widespread household air\hyp{}conditioning  
and longer usage duration reinforce peak
rigidity; the QNI interconnection with NSW provides cross-regional
support; the absence of daylight saving time in the state further
concentrates peak load from late afternoon to evening.
\end{enumerate}

\paragraph{SA (South Australia)}
\begin{enumerate}
\renewcommand{\labelenumi}{(\roman{enumi})}
\item Seasonal characteristics: the annual amplitude is significantly
modulated by distributed photovoltaics (PV); on clear days in late
spring/early summer, the midday trough deepens; both summer
evening peaks and winter evening peaks may exhibit sharp spikes.
\item Intraday characteristics: a distinct midday trough appears,
while the evening peak is sharp with a steep rising edge; on
working days, the dual-peak structure becomes more asymmetric
after the daytime portion is ``cut down'' by PV output.
\item Cause analysis: the nation-leading penetration of rooftop PV
substantially depresses the midday ``operational load'' and
triggers operational constraints during extremely low-load
periods~\cite{aemo_msl_sa}; the relatively small local load base
makes marginal volatility more pronounced; SA is connected with
VIC through the Heywood and Murraylink interconnectors, with a
higher proportion of daytime ``export'' and evening ``import'' on
event days, making cross-regional power flows critical for
stability.
\end{enumerate}

\paragraph{VIC (Victoria)}
\begin{enumerate}
\renewcommand{\labelenumi}{(\roman{enumi})}
\item Seasonal characteristics: both summer cooling and winter heating create dual pressure, and either extreme heat or cold waves may form candidates for the annual peak; fluctuations during the shoulder seasons occur more frequently due to alternating warm and cold conditions.
\item Intraday characteristics: with increasing Rooftop PV penetration, the daytime curve follows a three-stage pattern of ``rise--decline--evening peak''; the weekday dual-peak structure is evident, and the evening peak aligns more closely with residents' after-work hours.
\item Cause analysis: the temperate oceanic climate and the high share of urban and manufacturing sectors lead to both cooling and heating demand drivers; bidirectional interconnection with NSW (VNI) and SA (Heywood/Murraylink) enhances the system's ability to balance between peak and off-peak periods; Rooftop PV and large-scale renewable integration have strengthened the daytime peak-shaving effect year by year.
\end{enumerate}

\paragraph{TAS (Tasmania)}
\begin{enumerate}
  \renewcommand{\labelenumi}{(\roman{enumi})}
  \item Seasonal characteristics: a typical ``winter peak--summer trough'' pattern, with winter heating demand dominating and relatively mild summer load; the annual amplitude is strongly driven by cold-air processes.
  \item Intraday characteristics: the daytime load remains relatively flat, followed by an increase from evening to nighttime; weekday--weekend differences are smaller than in NSW/VIC; the evening peak becomes more prominent on extremely cold days.
  \item Cause analysis: the high-latitude oceanic climate combined with a high share of electric heating leads to winter peaks; the generation mix is dominated by hydropower, and the overall load scale is small; interconnection with VIC via Basslink enables seasonal and peak--off-peak energy balancing and supports system security~\cite{aemo_nem_map}.
\end{enumerate}

\subsubsection{Overview of interstate similarities, differences, and causes}

\begin{enumerate}
  \renewcommand{\labelenumi}{(\roman{enumi})}
  \item \textbf{Seasonal characteristics:}
  QLD exhibits the most stable and longest-lasting summer
  temperature-driven peaks; TAS shows the most pronounced winter
  heating peaks; NSW/VIC experience the largest annual amplitude
  due to dual drivers of cooling and heating, making the causes of
  annual peaks non-unique; SA is strongly influenced by Rooftop PV
  penetration, resulting in a prominent ``midday trough'' during
  spring and summer~\cite{aemo_msl_sa}.

  \item \textbf{Intraday characteristics:}
  NSW/VIC display the most typical ``morning--evening dual-peak''
  structure, reflecting residential--commercial temporal rhythms
  and commuting patterns; QLD tends toward a ``late-afternoon to
  evening single-peak ramp'' during hot periods; SA shows sharp
  midday troughs and evening spikes due to the combined effect of
  ``PV suppressing midday load and demand pushing the evening
  peak''; TAS exhibits a flat daytime profile followed by an
  evening rise, consistent with heating and home-energy usage
  patterns.

  \item \textbf{Causes and mechanisms:}
  latitude and climate zones (tropical/subtropical/temperate/cold),
  industrial structure and population distribution
  (residential--commercial share, manufacturing weight),
  distributed PV penetration and degree of electrification
  (especially the PV-induced midday peak shaving in SA and VIC),
  differences in daylight-saving policies (QLD not adopting DST
  leading to different peak--solar alignment), and the buffering
  effects of the NEM interconnection network (QNI, VNI, Heywood,
  Murraylink, Basslink) collectively shape interstate differences
  in load curves~\cite{aemo_nem_map,aemo_msl_sa}. For modeling
  purposes, these factors determine the necessity of region-specific
  modeling, the importance of exogenous variables (particularly
  temperature/solar/distributed generation and event information),
  and the value of cross-horizon consistency constraints within
  the scheduling chain.
\end{enumerate}

\FloatBarrier

\subsubsection{Guidance for the model}

Facing the above regional differences, modeling should adhere to
the principles of ``state-specific adaptation, cross-temporal
consistency, and event interpretability.'' The specific contents
are as follows:
\begin{enumerate}
  \renewcommand{\labelenumi}{(\roman{enumi})}
  \item \textbf{Architecture and regionalization:}
  adopt a multi-task or hierarchical architecture, where the
  shared backbone learns common seasonal--weekly--intraday
  rhythms. State-specific embeddings together with
  state-specific output heads (or gating) are introduced so that
  interstate characteristics---such as the dual cooling--heating
  drivers of NSW/VIC, the high-temperature rigidity of QLD, the
  deep midday valley of SA, and the winter peak of TAS---can be
  modeled and parameterized in a dedicated manner.
  \item \textbf{External data and textual information:}
  build targeted external datasets from mechanistic and
  geographical perspectives, and exploit the complementarity of
  multi-source textual information. On the demand side, collect
  meteorological data (temperature, humidity, wind,
  precipitation, etc.); on the supply side, incorporate
  operation and maintenance announcements, maintenance/fault
  records, and information on interconnection capacity and
  congestion; on the institutional side, include policy and
  price-mechanism adjustments; on the social side, integrate
  news and social media. All textual data must be strictly
  aligned along the date and regional dimensions, undergo
  deduplication and correlation filtering, and be assigned
  reasonable effective lags and temporal decay. On the modeling
  side, location-dependent selection and weighting mechanisms
  are adopted to dynamically determine when and where, and which
  type of source, should be activated, so as to amplify valid
  signals while suppressing noise propagation.
  \item \textbf{Multi-scale forecasting and rapid deployment:}
  for different forecasting horizons such as hourly, daily, and
  monthly, a unified model is applied to maintain cross-scale
  consistency; on the engineering side, lightweight and
  parallelizable interfaces are built to enable convenient
  invocation and low-latency outputs. During online operation,
  periodic rolling updates are carried out to monitor
  distribution drift and peak-position migration, and, when
  necessary, robust/quantile losses and safety fallback
  mechanisms are triggered to ensure stability under extreme
  conditions. Interfaces and configurations remain unified,
  allowing one-click switching of forecasting horizons and
  regions to satisfy operational practicality.
\end{enumerate}

\subsection{External information sources for text}
\label{subsec:text_sources}

This study targets three categories of textual external data
sources---News, Social Media (Reddit), and Policy---all covering
the period 2019--2021 and mainly consisting of English materials,
with the sources listed as follows.
\textcolor{red}{This step corresponds to \textbf{acquisition}, where we collect public documents/posts from heterogeneous channels and form the raw text corpora before any temporal/region anchoring.}

\textbf{News.}
The news sources include two major categories: the first is
global/regional news indexes and archives (such as internationally
used news index platforms), which are used to batch-retrieve
Australia-related reports by keywords and dates; the second is
publicly released articles and column pieces on the main sites of
Australian media and the official websites of local media
(including ABC News, The Australian, The Sydney Morning Herald,
The Age, The Guardian Australia, Australian Financial Review,
etc.). These sources can continuously cover ``system-level /
regional-level'' information such as emergencies, planned
maintenance, policy releases, extreme weather, and market price
fluctuations, with typical data entry points including the news
and announcement pages published by institutions such as AER and
AEMO~\cite{aer2025news,aemo_market_notices}.
\textcolor{red}{To address data licensing and redistribution constraints, we do not redistribute full copyrighted news texts in our released benchmark; instead, we provide document-level metadata (e.g., date, source outlet, and URL when applicable) together with derived representations (e.g., embeddings and aggregated statistics) and the preprocessing scripts, so that interested readers can re-acquire the content from the original publishers under their respective terms.}

\textbf{Reddit (social media).}
The social media sources mainly consist of public subforums and
topic posts on Reddit, where keyword searches are conducted around
themes such as energy, electricity markets, and travel activities,
with attention paid to communities with state/city identifiers
(such as r/australia, r/sydney, r/melbourne, r/brisbane,
r/Adelaide, r/Tasmania, r/energy, r/solar, etc.). This source is
closer to the daily energy use, travel, and activity rhythms of
residents and enterprises, and can reflect the short-term impact
of events, festivals, and heatwaves on electricity load
curves~\cite{reddit2025energyforum}.
\textcolor{red}{Data governance and ethics: we only collect publicly accessible Reddit posts (no private messages or closed communities), apply rate limiting, and comply with Reddit's platform policies/terms for research use. We do not perform user-level profiling or individual inference, and we remove or avoid storing personally identifiable fields (e.g., usernames, profile links) in the processed dataset.}
\textcolor{red}{Licensing and release form: since Reddit content remains subject to the platform's and authors' rights, we do not redistribute raw Reddit text in our released benchmark. Instead, we release post identifiers and non-identifying metadata (e.g., timestamps and subreddit names when permissible), together with derived features (e.g., embeddings) and the full preprocessing pipeline, enabling reproducible re-collection by readers under the same policies.}

\textbf{Policy.}
Policy sources come from the official websites and announcement
channels of Australia's federal and state energy and market
institutions, including AEMO's Market Notices, ESOO/GSOO/ISP and
other reports, AER (Australian Energy Regulator)'s regulatory
decisions and annual reviews, AEMC (Australian Energy Market
Commission)'s rule change consultations and final decisions, as
well as state-level regulatory and pricing institutions (such as
IPART of NSW, ESC of VIC, QCA of QLD, ESCOSA of SA, and OTTER of
TAS) and state government gazettes/ministerial statements. This
type of text has a low publication frequency but high impact
intensity, and often corresponds to institutional changes in grid
connection rules, capacity and reliability standards, price
mechanisms, and network planning~\cite{aemo_market_notices,dcceew2025strategies,aemc2025overview}.
\textcolor{red}{These policy documents are obtained from official public channels; in the released benchmark we provide the source links and derived representations, and we preserve the original provenance information to support traceability and compliant reuse.}

\begin{figure*}[t]
  \begin{subfigure}{\textwidth}
    \centering
    \includegraphics[width=\textwidth]{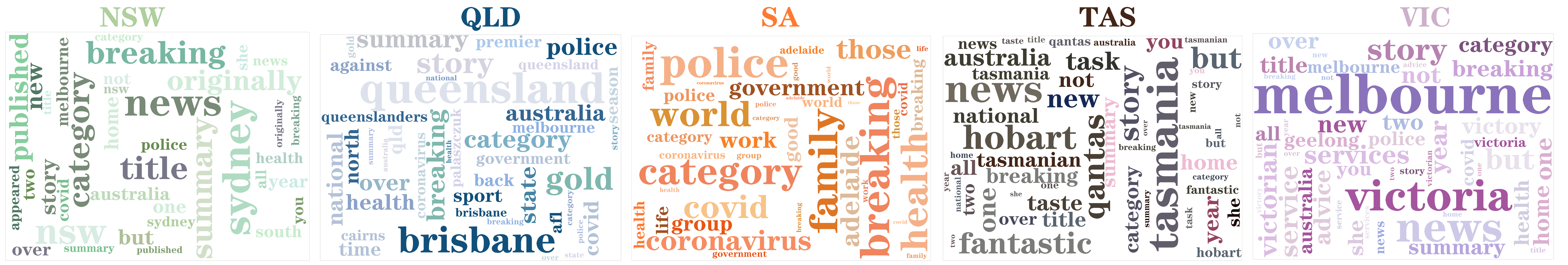}
    \caption{State-wise word clouds of the general news corpus
    (News). Each panel corresponds to one NEM region (NSW, QLD, SA,
    TAS, VIC); font size is proportional to term frequency after
    preprocessing.}
    \label{fig:wordcloud-news}
  \end{subfigure}

  \vspace{0.8em}

  \begin{subfigure}{\textwidth}
    \centering
    \includegraphics[width=\textwidth]{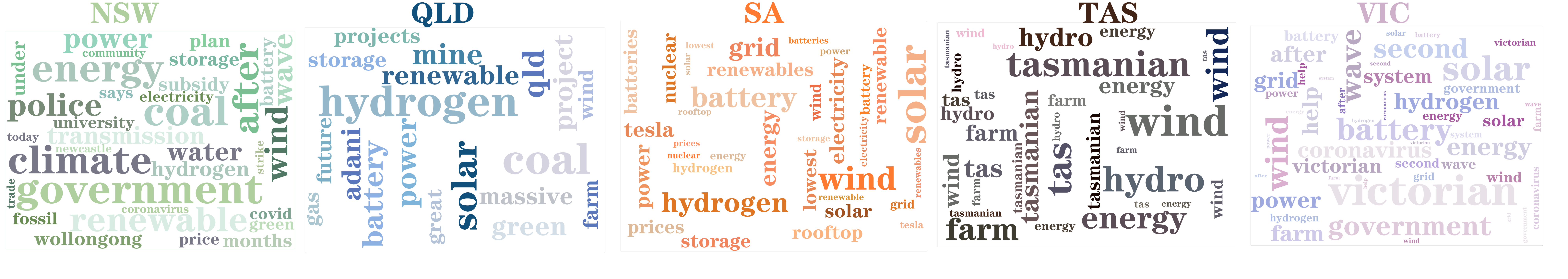}
    \caption{State-wise word clouds of the Reddit social-media
    corpus (Reddit). High-frequency terms highlight everyday events,
    public concerns and behavioural patterns that may influence
    electricity demand.}
    \label{fig:wordcloud-reddit}
  \end{subfigure}

  \vspace{0.8em}

  \begin{subfigure}{\textwidth}
    \centering
    \includegraphics[width=\textwidth]{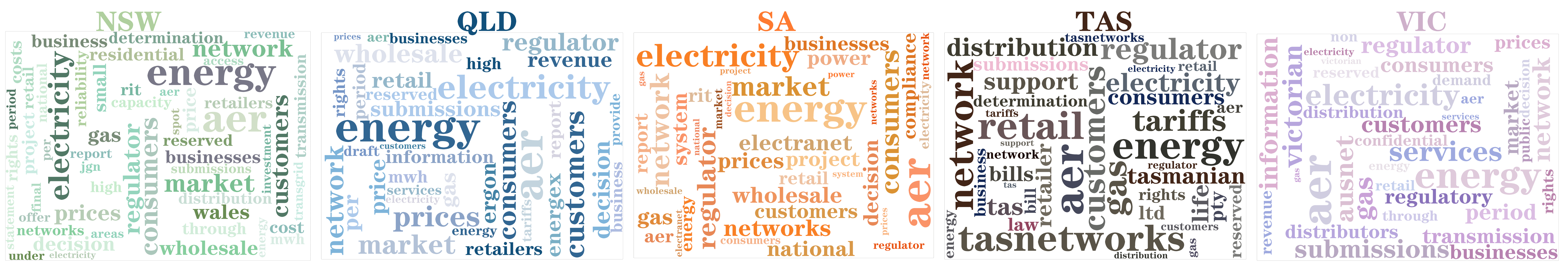}
    \caption{State-wise word clouds of the regulatory and policy
    corpus (Policy), mainly sourced from AER, AEMO and network
    businesses. High-frequency terms emphasise tariff, price and
    network regulation topics.}
    \label{fig:wordcloud-policy}
  \end{subfigure}

  \caption{Word cloud visualizations of the three text corpora across NEM states.}
  \label{fig:wordclouds-all}
\end{figure*}

To more intuitively show the differences of the three types of
external texts in the semantic space, we aggregate the corpora by
state, count the word frequencies of the cleaned terms, and
provide representative word cloud visualizations in
Figures~\ref{fig:wordcloud-news}--\ref{fig:wordcloud-policy}.
Figure~\ref{fig:wordcloud-news} shows the word clouds of the News
corpus in the five states of NSW, QLD, SA, TAS, and VIC, where
each state is dominated by city names such as Sydney, Brisbane,
and Melbourne, as well as high-frequency words such as
\emph{breaking}, \emph{police}, and \emph{coronavirus}, reflecting
general news themes such as emergencies, public security, and the
pandemic. Figure~\ref{fig:wordcloud-reddit} corresponds to the
social media source (Reddit), where keywords are more related to
daily life, public sentiment, and travel behaviour, reflecting
information such as residents' activity rhythms and demand-side
responses. Figure~\ref{fig:wordcloud-policy} shows regulatory and
policy (Policy) documents, where high-frequency words are
concentrated on \emph{energy}, \emph{electricity}, \emph{network},
\emph{prices}, \emph{tariffs}, \emph{AER}, \emph{customers}, etc.,
indicating that this source mainly records content such as
electricity market regulation, network tariffing, and retail
electricity price adjustments. These visualization results verify,
from the semantic level, the potential correlation between
external text and electricity load, and also show that News,
social media, and Policy texts provide complementary information
for GRAFT along the ``events--behaviour--regulation/policy''
dimension.
\textcolor{red}{Throughout data collection and release, we follow a privacy-preserving and license-compliant principle: raw third-party texts are not redistributed when restricted by platform/publisher terms; instead, we provide provenance, identifiers (when permissible), derived representations, and reproducible scripts to support transparent and compliant research reuse.}

\tcbset{
  evcard/.style={
    enhanced,
    breakable,
    width=\textwidth,
    colback=white,
    colframe=black!25,
    colbacktitle=black!12,
    coltitle=black,
    fonttitle=\bfseries\scriptsize,
    boxrule=0.4pt,
    arc=1.2pt,
    left=4pt,right=4pt,top=2.5pt,bottom=2.5pt, 
    boxsep=1pt
  }
}

\newcommand{\kw}[1]{%
  \tcbox[
    on line,
    colback=black!8,
    colframe=black!25,
    boxrule=0.3pt,
    arc=1.2pt,
    left=3pt,right=3pt,top=1pt,bottom=1pt
  ]{\scriptsize\textbf{#1}}%
}

\newcommand{\badge}[1]{%
  \tcbox[
    on line,
    colback=black!12,
    colframe=black!35,
    boxrule=0.3pt,
    arc=1.0pt,
    left=4pt,right=4pt,top=1pt,bottom=1pt
  ]{\scriptsize\textbf{#1}}%
}

\newcommand{\event}[6]{%
  \noindent
  \badge{#1}\hspace{0.55em}%
  \textbf{#2}~(\textbf{#3}) --- #4\par
  \hspace*{2.35em}%
  \textbf{Keywords:}~#5\hspace{0.9em}%
  \textit{Source:}~#6\par
  \vspace{1.5pt}
}

To facilitate understanding of the correspondence between various
types of texts and electricity-related events,
Figure~\ref{fig:text-events-cards} provides several representative
examples of text events from 2019--2021. Grouped into the three
categories of News, Reddit, and Policy, the figure lists the event
occurrence time, the region where it occurred, the content summary,
the keywords, and the source.

\begin{center}
\scriptsize

\begin{tcolorbox}[evcard,breakable,
  title={News text events (2019--2021)~\cite{aer2025news,aemo_market_notices}},
  before skip=4pt,after skip=4pt]
\RaggedRight
\event{N1}{2019-12-01}{NSW}{Thousands without power in Sydney after severe storms.}{\kw{storm}\ \kw{blackout}}{ABC News}
\event{N2}{2019-12-05}{QLD}{Brisbane property market strengthens under low interest rates and population growth.}{\kw{economy}\ \kw{demand context}}{The Courier-Mail}
\event{N3}{2020-01-01}{NSW}{Around 50{,}000 homes lose power during New South Wales bushfire crisis.}{\kw{bushfire}\ \kw{outage}}{ABC News}
\event{N4}{2020-01-02}{VIC}{Victoria declares State of Disaster due to bushfires and extreme risk conditions.}{\kw{bushfire}\ \kw{extreme risk}}{The Age}
\event{N5}{2021-05-25}{QLD}{Explosion at Callide Power Station triggers widespread blackouts across Queensland.}{\kw{Callide}\ \kw{blackout}}{ABC News}
\event{N6}{2021-06-09}{VIC}{Severe storms leave tens of thousands without power and damage network assets.}{\kw{storm}\ \kw{infrastructure}}{ABC News}
\end{tcolorbox}

\begin{tcolorbox}[evcard,breakable,
  title={Reddit (social media) text events (2019--2021)~\cite{reddit2025energyforum}},
  before skip=4pt,after skip=4pt]
\RaggedRight
\event{R1}{2019-10-18}{TAS}{Posts on wind generation surge in Tasmania and its impact on demand.}{\kw{wind generation}\ \kw{demand shift}}{r/tasmania}
\event{R2}{2019-12-14}{NSW}{Thread on Sydney heatwave pushing the grid to its limits.}{\kw{heatwave}\ \kw{high demand}}{r/sydney}
\event{R3}{2020-01-08}{NSW/VIC}{Reports that bushfire smoke reduces solar generation across NSW and Victoria.}{\kw{bushfire}\ \kw{solar loss}}{r/australia}
\event{R4}{2020-04-10}{National}{Lockdown discussions note reduced commercial electricity demand during COVID-19.}{\kw{COVID-19}\ \kw{demand drop}}{r/australia}
\event{R5}{2021-02-07}{QLD}{Posts about record Queensland heat pushing electricity demand up.}{\kw{heatwave}\ \kw{high demand}}{r/brisbane}
\event{R6}{2021-06-15}{VIC}{Users share photos of storms damaging power lines and causing outages.}{\kw{storm}\ \kw{outage}}{r/melbourne}
\end{tcolorbox}

\begin{tcolorbox}[evcard,breakable,
  title={Policy text events (2019--2021)~\cite{aemo_market_notices,dcceew2025strategies,aemc2025overview}},
  before skip=4pt,after skip=4pt]
\RaggedRight
\event{P1}{2019-11-29}{National}{AER issues summer readiness requirements for market participants across the NEM.}{\kw{summer readiness}\ \kw{reliability}}{AER}
\event{P2}{2019-12-18}{National}{AER publishes final decision on Value of Customer Reliability (VCR) methodology.}{\kw{VCR}\ \kw{reliability valuation}}{AER}
\event{P3}{2020-01-31}{NSW}{AER triggers the NSW Retailer Reliability Obligation (RRO) for upcoming summers.}{\kw{RRO}\ \kw{reliability}}{AER}
\event{P4}{2020-03-11}{NSW/VIC/SA}{Report investigates price spikes in New South Wales, Victoria and South Australia.}{\kw{price spikes}\ \kw{volatility}}{AER}
\event{P5}{2021-10-19}{National}{Guidelines released for Wholesale Demand Response participation in the NEM.}{\kw{WDR}\ \kw{demand response}}{AEMO}
\event{P6}{2021-12-16}{SA/QLD}{Regulatory note on price spikes above \$5000/MWh in South Australia and Queensland.}{\kw{high price event}}{AER}
\end{tcolorbox}

\end{center}

\begin{figure}[t]
\centering
\caption{Representative external text events (2019--2021) from news, social media, and policy sources.}
\label{fig:text-events-cards}
\end{figure}

For any source $s\in\{\text{News},\text{Reddit},\text{Policy}\}$,
state $r$, and calendar day $t$, denote the cleaned text collection
of that day as
$\mathcal{D}_{s,r,t}=\{d_1,\dots,d_{n_{s,r,t}}\}$.
For a single document $d$, the Sentence-BERT (SBERT) encoder
$E(\cdot)$~\cite{reimers2019sbert} is used to obtain the vector
representation and perform unit normalization:
\begin{equation}
\mathbf{e}_{d}
\;=\;
\frac{E(d)}{\lVert E(d)\rVert_2}
\in\mathbb{R}^{d_s}.
\label{eq:sbert_encode}
\end{equation}

Based on this, daily aggregation is performed according to
relevance weights to obtain state--day level text vectors:
\begin{equation}
\mathbf{x}^{(s)}_{r,t}
\;=\;
\sum_{d\in\mathcal{D}_{s,r,t}} w_d\,\mathbf{e}_d,
\qquad
\sum_{d\in\mathcal{D}_{s,r,t}} w_d = 1,\;
w_d\ge 0,
\label{eq:daily_agg}
\end{equation}
where $w_d$ is the normalized weight of document $d$, which can be
obtained by applying softmax to scores based on keywords, named
entities, or source credibility.
For policy texts, in order to reflect cross-day effects, a
time-decaying validity weight $\rho^{\Delta t}$
($\rho\in(0,1]$, $\Delta t$ is the number of days since the
release date) is introduced to modify the daily aggregation:
\begin{equation}
\mathbf{x}^{(\text{Policy})}_{r,t}
\;=\;
\sum_{d\in\mathcal{D}_{\text{Policy},r,\le t}}
\underbrace{\frac{\tilde{w}_d\,\rho ^{\,t-t_d}}
{\sum_{d'} \tilde{w}_{d'}\,\rho ^{\,t-t_{d'}}}}_{\displaystyle
w_d(t)}
\,\mathbf{e}_d,
\label{eq:policy_decay}
\end{equation}
where $t_d$ is the document release date, and $\tilde{w}_d$ is the
relevance weight before decay.
If a text can only be located at the national level, it is
broadcast to all states through a fixed mapping; if a state-level
geographic label can be parsed, it is injected only into the
corresponding state.

Define an availability mask if no text is available on that day:
\begin{equation}
m^{(s)}_{r,t}
\;=\;
\mathbb{I}\!\left[\,\lvert\mathcal{D}_{s,r,t}\rvert>0\,\right]
\in\{0,1\},
\label{eq:mask_def}
\end{equation}
Instead of using zero vector placeholders for missing sources,
masks are retained to prevent spurious signals from entering
subsequent readout paths.
To facilitate multi-source collaboration and dimensionality
unification, the three types of sources are concatenated at the
hierarchical level and then linearly projected to obtain a unified
representation:
\begin{equation}
\mathbf{z}_{r,t}
=\mathbf{P}\Big[
  m^{(\mathrm{News})}_{r,t}\mathbf{x}^{(\mathrm{News})}_{r,t};\allowbreak\,
  m^{(\mathrm{Reddit})}_{r,t}\mathbf{x}^{(\mathrm{Reddit})}_{r,t};\allowbreak\,
  m^{(\mathrm{Policy})}_{r,t}\mathbf{x}^{(\mathrm{Policy})}_{r,t}
\Big]\in\mathbb{R}^{d}.
\label{eq:proj_concat}
\end{equation}
where $[\cdot;\cdot]$ denotes vector concatenation,
$\mathbf{P}\in
\mathbb{R}^{d\times(d_{\text{News}}+d_{\text{Reddit}}+d_{\text{Policy}})}$
is a learnable or predefined linear projection matrix, and $d$ is
the unified external text feature dimension.

The text is at daily resolution and aligned to the half-hour
axis using a \sout{``same-day broadcasting''} \textcolor{red}{\emph{within-day broadcasting} strategy on the \emph{historical} calendar day}:
\begin{equation}
\mathbf{z}_{r,t,\ell} \;=\; \mathbf{z}_{r,t}, \qquad
\forall\,\ell\in\{1,\dots,48\},
\label{eq:broadcast}
\end{equation}
and the corresponding availability mask is broadcast in the
same way $m^{(s)}_{r,t,\ell} = m^{(s)}_{r,t}$.
\textcolor{red}{Importantly, in the day-ahead setting with fixed $T_{\text{out}}=48$, when forecasting day $d$ we only use historical load and external texts up to the end of day $d\!-\!1$; thus no forecast-day text is used and no information leakage occurs.}
Training samples are constructed using a sliding window:
given a historical window length $L$ and a \sout{prediction end day}
\textcolor{red}{last historical day}
$T$, the calendar day $T$ at the end of the input window is used as
the alignment anchor, forming
\begin{align}
\mathcal{X}_{r,T}
&= \big\{\,\mathbf{y}_{r,t,\ell}\ \big|\ 
t\in[T-L+1,\,T],\ \ell\in\{1,\dots,48\}\,\big\},
\\
\mathcal{Z}_{r,T}
&= \big\{\,\mathbf{z}_{r,t}\ \big|\ 
t\in[T-L+1,\,T]\,\big\},
\label{eq:window_pack}
\end{align}
and $\mathcal{Z}_{r,T}$ is then broadcast to half-hour
resolution according to equation~(5) to synchronize with
$\mathcal{X}_{r,T}$ for subsequent position-dependent reading
and injection.

\textcolor{red}{This procedure corresponds to \textbf{alignment}: texts are anchored to the window end day $T$ on the ``date $\times$ region'' axis and synchronized to the half-hour timeline via Eq.~\eqref{eq:broadcast} with explicit masking for missing days.}

To maintain inference consistency, the masks
$\{m^{(s)}_{r,t}\}$ and the time-decay factors share the same
computation rules in both training and testing phases, and
missing sources are not replaced by zero vectors.

For the sake of reproducibility and fast retrieval, we next summarize the collection
specifications, time granularity, and the preprocessing steps corresponding to
equations~(1)--(7) for each data source---including time standardization, same-day
broadcasting, state-level mapping, and (for the policy source) time-decay processing.
We also specify a unified strategy for missing and abnormal values, facilitating
consistent data entry across different experimental configurations.

\begin{tcolorbox}[title={Data sources and preprocessing overview}, colback=white, colframe=black!30,
  boxrule=0.4pt, arc=1pt, left=4pt, right=4pt, top=3pt, bottom=3pt, breakable]
\scriptsize
\begin{itemize}\setlength{\itemsep}{2pt}\setlength{\parskip}{0pt}

\item \textbf{Load (target).}
Source: AEMO regional load.
Coverage: 2019--2021.
Granularity: 30\,min.
Preprocessing: outlier/missing-value correction; timezone and DST standardization~\cite{aemo_data_nem}.

\item \textbf{Weather / calendar.}
Source: weather data; holiday calendar.
Coverage: 2019--2021.
Granularity: daily.
Preprocessing: same-day broadcast; workday/holiday/season flags~\cite{hong2016pelfforecasting}.

\item \textbf{News.}
Source: news indexes and AU media.
Coverage: 2019--2021.
Granularity: document.
Preprocessing: cleaning; deduplication; SBERT; daily alignment; state mapping or national broadcast~\cite{aer2025news,aemo_market_notices}.

\item \textbf{Reddit.}
Source: public subreddits.
Coverage: 2019--2021.
Granularity: post.
Preprocessing: cleaning; deduplication; SBERT; daily alignment; state mapping or national broadcast~\cite{reddit2025energyforum}.

\item \textbf{Policy.}
Source: government/regulator sites.
Coverage: 2019--2021.
Granularity: notice.
Preprocessing: cleaning; SBERT; daily alignment with time decay; mapping to applicable regions~\cite{aemo_market_notices,aer2025news,aemc2025overview,dcceew2025strategies}.

\end{itemize}
\end{tcolorbox}

\FloatBarrier

\section{GRAFT: GRid-Aware Forecasting with Text}
\label{sec:graft}

\begin{figure*}[t]
  \centering
  \includegraphics[width=\textwidth,
                   height=0.48\textheight,
                   keepaspectratio]{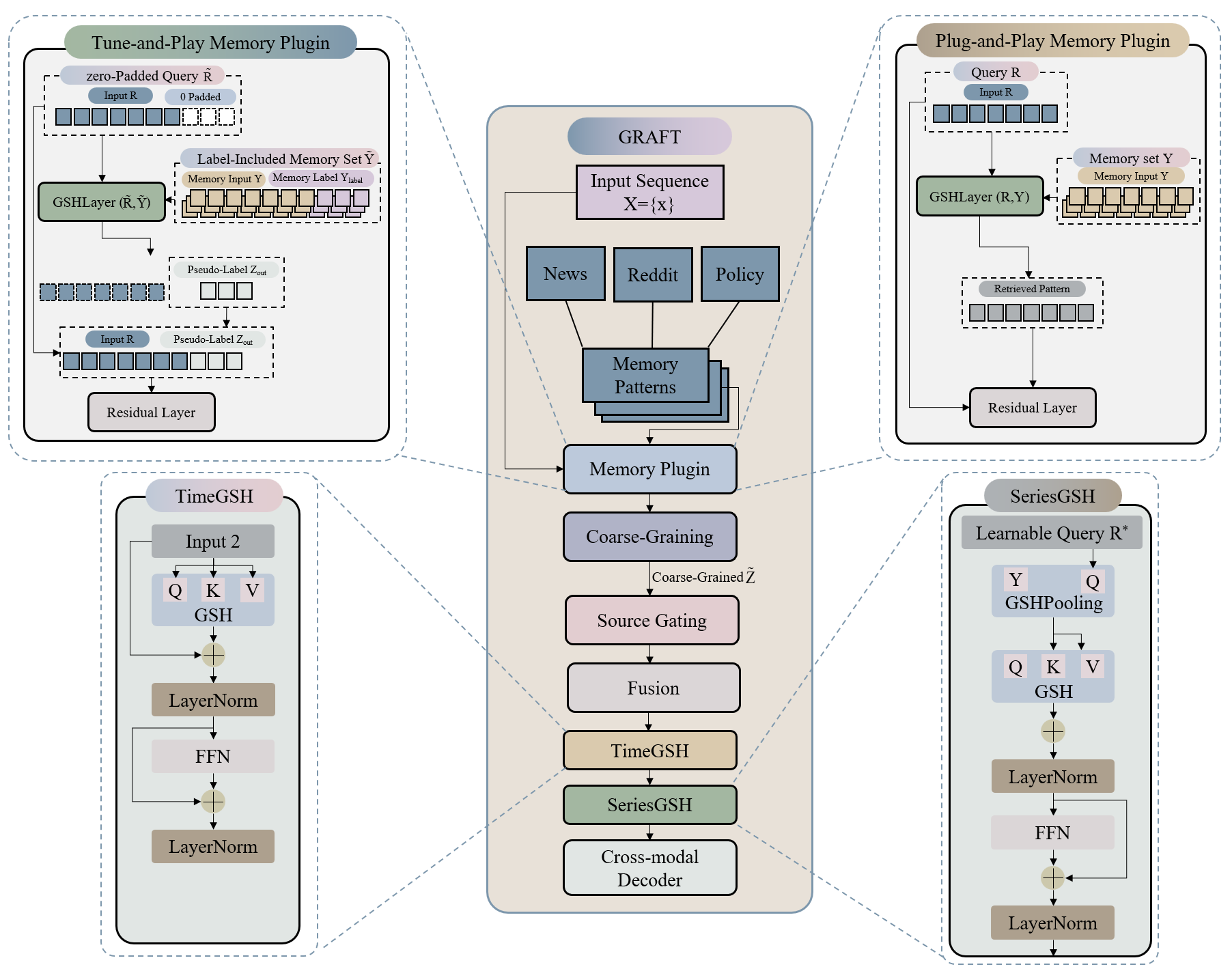}
  \caption{Overall architecture of the proposed GRAFT model.
  The central block is the numerical backbone STanHop (Memory Patterns,
  Memory Plugin, Coarse-Graining, TimeGSH, SeriesGSH)~\cite{wu2023stanhopsparsetandemhopfield}.
  Three textual external sources (News / Reddit / Policy) are encoded as memory patterns,
  gated and fused through the Source Gating and Fusion modules, and then injected into the backbone
  to form TimeGSH and SeriesGSH representations.}
  \label{fig:graft-pipeline}
\end{figure*}

\begin{figure*}[!t]
  \centering
  \includegraphics[width=0.88\textwidth,keepaspectratio]{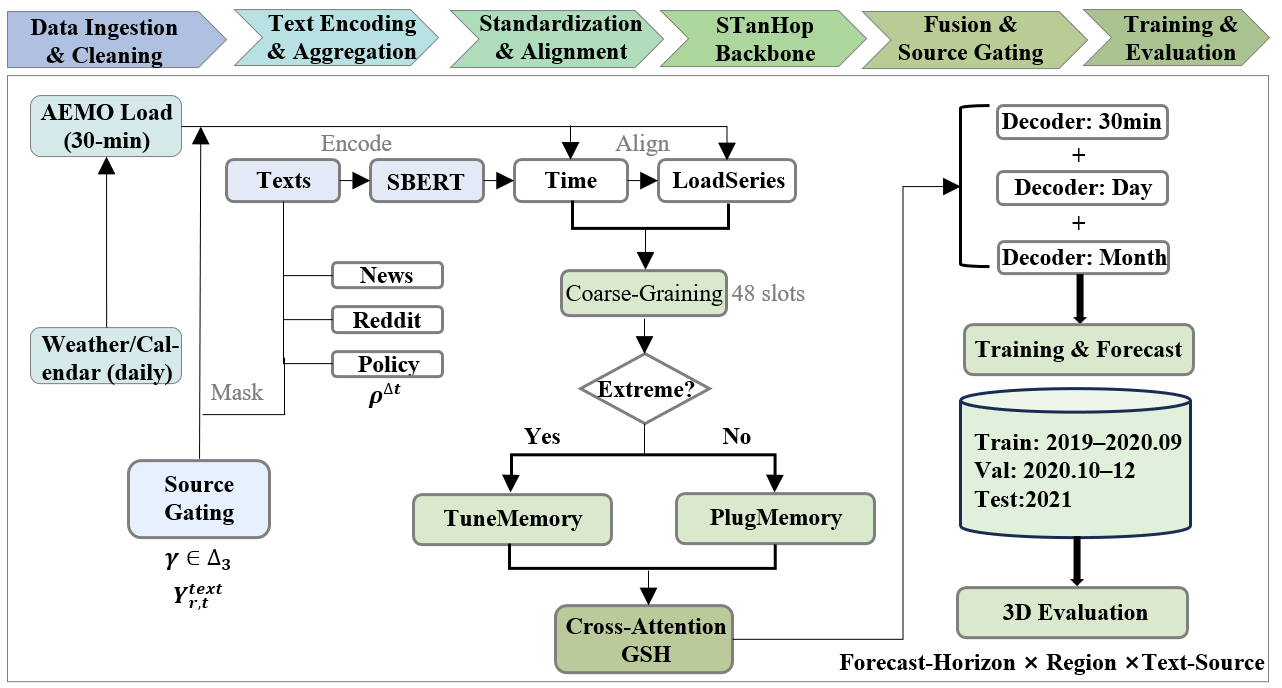}
  \caption{Forward-computation flow of GRAFT, illustrating how historical load sequences,
  textual external sources and multi-scale representations are processed in the encoding,
  retrieval and decoding stages. The design builds on the STanHop backbone~\cite{wu2023stanhopsparsetandemhopfield}
  and the theory of sparse modern Hopfield networks~\cite{ramsauer2021hopfieldnetworksneed,hu2023sparsemodernhopfieldmodel}.}
  \label{fig:graft}
\end{figure*}

\subsection{Memory Model Background and Symbols}
\label{subsec:background_notation}

Let the query matrix of the input sequence be
$R\in\mathbb{R}^{|R|\times d}$, where $|R|$ denotes the number of
time steps (sequence length) and $d$ is the feature dimension of each
time step. Let the retrievable external memory be
$Y\in\mathbb{R}^{M\times |R|\times d}$, where $M$ is the number of
memory entries (or the number of candidate key–value pairs).
For consistency with the attention formulation, both $R$ and $Y$ are
linearly projected (formally consistent with the query–key–value
structure in modern Hopfield networks~\cite{ramsauer2021hopfieldnetworksneed,hu2023sparsemodernhopfieldmodel}):
\begin{equation}
Q = R W_Q,\qquad
K = Y W_K,\qquad
V = Y W_V,
\label{eq:lin-maps}
\end{equation}
where $Q$, $K$ and $V$ are the representation matrices of query, key
and value, respectively, and $W_Q$, $W_K$, $W_V$ are learnable
projection matrices.

\subsection{Modern Hopfield and Generalized Sparse Hopfield (GSH)}\label{subsec:gsh}

\textbf{Energy function and variational characterization of $\alpha$-EntMax.}
Let the probability simplex be
$\Delta_M=\{p\in\mathbb{R}^M_{\ge 0}\mid \sum_{\mu=1}^M p_\mu=1\}$.
Define the Tsallis $\alpha$-entropy as~\cite{tsallis1988entropy}
\begin{equation}
\Psi_\alpha(p)=
\begin{cases}
\dfrac{1}{\alpha(\alpha-1)}\displaystyle\sum_{\mu=1}^{M}\bigl(p_\mu-p_\mu^\alpha\bigr), & \alpha\neq 1,\\[4pt]
-\displaystyle\sum_{\mu=1}^{M} p_\mu\ln p_\mu, & \alpha=1,
\end{cases}
\label{eq:tsallis}
\end{equation}
and denote its convex conjugate by $\Psi_\alpha^\star(p)$.
Let the memory pattern matrix be
$\Xi=[\xi_1,\dots,\xi_M]\in\mathbb{R}^{d\times M}$
(where $\xi_\mu\in\mathbb{R}^d$ is the $\mu$-th pattern), and let the inverse
temperature be $\beta>0$.
The energy function of the Generalized Sparse Hopfield (GSH)~\cite{hu2023sparsemodernhopfieldmodel} is defined as
\begin{equation}
H(x)= -\Psi_\alpha^\star\!\bigl(\beta\,\Xi^\top x\bigr)+\frac{1}{2}\langle x,x\rangle + C,
\label{eq:gsh-energy}
\end{equation}
where $x\in\mathbb{R}^d$ is the current state (the query vector to be retrieved),
$\langle x,x\rangle$ is the standard inner product, and $C$ is a constant
independent of $x$.
By the convex conjugate and variational optimality in Danskin's theorem, the
sparse assignment mapping can be obtained as
\begin{equation}
\alpha\text{-EntMax}(z)=\arg\max_{p\in\Delta_M}\bigl[\langle p,z\rangle-\Psi_\alpha(p)\bigr],\qquad 
\end{equation}
where $\alpha$-EntMax maps the scores $z$ to a sparse probability vector and
serves as a sparse generalization of softmax~\cite{martins2016sparsemax,peters2019entmax}.

\begin{theorem}[EntMax–Conjugate Gradient]\label{thm:entmax-grad}
For any $z\in\mathbb{R}^M$, we have
\begin{equation}
\nabla\Psi_\alpha^\star(z)=\alpha\text{-EntMax}(z).
\label{eq:entmax-var}
\end{equation}
Eq.~\eqref{eq:entmax-var} provides the closed-form first-order condition
for mapping the ``scores'' $z$ to a sparse probability distribution~\cite{peters2019entmax}.
\end{theorem}

\begin{theorem}[Lyapunov Monotonicity]\label{thm:mono}
Combining the gradient descent of Eq.~\eqref{eq:gsh-energy} yields the first-order
retrieval iteration of the GSH:
\begin{equation}
x_{t+1}=T(x_t)=\nabla_x \Psi_\alpha^\star\!\bigl(\beta\,\Xi^\top x_t\bigr)
=\alpha\text{-EntMax}\!\bigl(\beta\,\Xi^\top x_t\bigr).
\label{eq:gsh-dynamics}
\end{equation}
Along the iteration Eq.~\eqref{eq:gsh-dynamics}, the energy $H(x_t)$ is
monotonically non-increasing; if a limit point exists, it is a stationary
point of $H$~\cite{hu2023sparsemodernhopfieldmodel,ramsauer2021hopfieldnetworksneed}.
\end{theorem}

\begin{theorem}[Retrieval Error Bound]\label{thm:retrieval}
Let the pattern separation radius be
$R_{\min}=\tfrac{1}{2}\min_{\mu\neq\nu}\|\xi_\mu-\xi_\nu\|_2$, and let the
attraction basin $S_\mu$ be the set of all initial states that converge to
pattern $\xi_\mu$.
Then there exists a constant $m>0$ such that, for any $x\in S_\mu$ and
$1\le \alpha\le 2$, a single update of the GSH satisfies
\begin{equation}
\bigl\|T(x)-\xi_\mu\bigr\|
\;\le\; 2m(M-1)\exp\!\Bigl\{-\beta\bigl(\langle\xi_\mu,x\rangle-\max_{\nu\neq\mu}\langle\xi_\nu,\xi_\mu\rangle\bigr)\Bigr\},
\label{eq:retrieval-bound}
\end{equation}
where $M$ is the number of memory entries and $\beta$ is the inverse
temperature.
Moreover, under the assumption of randomly independent patterns, the number of
patterns that the model can store grows exponentially with the dimension $d$
(a capacity lower bound)~\cite{hu2023sparsemodernhopfieldmodel,arxiv2409_17515}.
\end{theorem}

Combining the above three theorems, we can derive the attention-form GSH layer
by extending the single-vector dynamics Eq.~\eqref{eq:gsh-dynamics} to batch
queries $R$ and memory $Y$, and by leveraging the equivalence
``modern Hopfield networks = attention layer''~\cite{ramsauer2021hopfieldnetworksneed}.
Together with Eq.~\eqref{eq:lin-maps}, we obtain
\begin{equation}
Z=\alpha\text{-EntMax}\!\bigl(\beta\,QK^\top\bigr)\,V,
\label{eq:gsh-attn}
\end{equation}
that is, by replacing the softmax in attention with $\alpha$-EntMax sparse
attention, it can be written as an explicit function of $R$ and $Y$:
\begin{equation}
\mathrm{GSH}(R,Y)=\alpha\text{-EntMax}\!\Bigl(\beta\,R W_Q W_K^\top Y^\top\Bigr)\,Y W_V.
\label{eq:gshlayer}
\end{equation}
Eqs.~\eqref{eq:gsh-attn}–\eqref{eq:gshlayer} serve as the basic operators
for the subsequent STanHop submodules along the temporal and sequential
dimensions~\cite{wu2023stanhopsparsetandemhopfield}.

\subsection{STanHop module and external memory plugin}\label{subsec:stanhop-block}
\textbf{Two-stage GSH.}
STanHop\cite{wu2023stanhopsparsetandemhopfield} extracts representations in a sequential fashion,
first applying temporal-dimension GSH (TimeGSH) and then cross-sequence-dimension GSH (SeriesGSH).
SeriesGSH contains a set of learnable prototype queries
$R^\star\in\mathbb{R}^{K\times d}$ to implement \texttt{GSHPooling}, where $K$
controls the effective size of the hidden dimension $D_{\mathrm{hidden}}$
(acting as a “complexity knob”). The module output is
\begin{equation}
  Z_{\text{out}}=\mathrm{STanHop}(R,Y)\in\mathbb{R}^{T\times C\times D_{\mathrm{hidden}}},
\end{equation}
where $T$ is the time step and $C$ is the number of channels (or regions/features).

\textbf{Pluggable external memory.}
To selectively incorporate historical experiences of similar scenarios during inference,
STanHop provides two mutually symmetric memory plug-ins\cite{wu2023stanhopsparsetandemhopfield}:
\begin{align}
\mathrm{PlugMemory}(R,Y)
&= \mathrm{LN}\!\bigl(R + \mathrm{GSH}(R,Y)\bigr), \label{eq:plug}\\
Z_{\mathrm{ret}}
&= \mathrm{GSH}(\tilde R,\tilde Y), \label{eq:tune-ret}\\
Z_{\mathrm{mix}}
&= \mathrm{FFN}(Z_{\mathrm{ret}}) + \tilde R, \label{eq:tune-mix}\\
\mathrm{TuneMemory}(\tilde R,\tilde Y)
&= \mathrm{LN}(Z_{\mathrm{mix}}). \label{eq:tune}
\end{align}

\noindent
Explanation: $R$ denotes the current query/representation input, and $Y$ the external memory;
$\tilde Y = Y \oplus Y_{\text{label}}$ concatenates the base memory with weakly labeled
“event–response’’ memory along the entry dimension, and $\tilde R$ is the
corresponding expanded query. $\mathrm{GSH}(\cdot,\cdot)$ is the sparse Hopfield
retrieval layer, $\mathrm{FFN}(\cdot)$ is a feedforward network, and
$\mathrm{LN}(\cdot)$ denotes layer normalization; $\oplus$ indicates
concatenation. Both plug-ins adopt a “residual + normalization’’ symmetric
structure: in Eq.~\eqref{eq:plug}, the memory is integrated with zero modification
under frozen parameters, whereas Eqs.~\eqref{eq:tune-ret}–\eqref{eq:tune} use
weakly labeled memory to enable example-driven calibration and
enhancement\cite{wu2023stanhopsparsetandemhopfield}.

\textbf{Multi-resolution coarsening.}
After each STanHop block, linear coarsening is applied to perform downsampling
along the temporal dimension\cite{wu2023stanhopsparsetandemhopfield}:
\begin{equation}
\hat{Z}_{c,t}=[\,Z_{c,t}\,;\,Z_{c,t+\Delta}\,],\qquad 
\tilde{Z}_{c,t}=W_{\mathrm{cg}}\,\hat{Z}_{c,t},
\label{eq:coarsen}
\end{equation}
where $[\cdot\,;\cdot]$ denotes concatenation along the channel dimension,
$W_{\mathrm{cg}}$ is the coarsening projection, and $\Delta$ is the coarsening
stride. This mechanism supports layer-wise distillation of multi-scale
information (“half-hour $\rightarrow$ hour $\rightarrow$ day’’) while
maintaining representational capacity and controlling computational cost.

\subsection{Using multi-source text as external memory for ``grid sensing''}\label{subsec:graft-extend}

\subsubsection{Encoding (text-to-memory)}
For each state $r$ and calendar day $t$, the aggregated daily text vectors are obtained separately from the three sources—News, Social Media, and Policy (see the construction process in Section~\ref{sec:data}):
\begin{equation}
  \mathbf{e}^{(\mathrm{news})}_{r,t}\in\mathbb{R}^{d_{\mathrm{news}}},\quad
  \mathbf{e}^{(\mathrm{reddit})}_{r,t}\in\mathbb{R}^{d_{\mathrm{rdt}}},\quad
  \mathbf{e}^{(\mathrm{policy})}_{r,t}\in\mathbb{R}^{d_{\mathrm{pol}}}.
\end{equation}
These vectors are encoded from the cleaned text collections using the
Sentence-BERT (SBERT) encoder\cite{reimers2019sbert} and aggregated on a daily
basis. Each source is then projected to a unified dimension $d$ and concatenated
by entries to form the day’s “text memory’’:
\begin{equation}
Y^{\text{text}}_{r,t}
=
\bigl[\,
\mathbf{e}^{(\mathrm{news})}_{r,t}W_{\mathrm{news}}\ ;\
\mathbf{e}^{(\mathrm{reddit})}_{r,t}W_{\mathrm{rdt}}\ ;\
\mathbf{e}^{(\mathrm{policy})}_{r,t}W_{\mathrm{pol}}
\,\bigr]
\in\mathbb{R}^{M_{\text{text}}\times d},
\label{eq:text-mem}
\end{equation}
where $W_{\mathrm{news}},W_{\mathrm{rdt}},W_{\mathrm{pol}}\in\mathbb{R}^{d_{\bullet}\times d}$ are
learnable linear projection matrices and $M_{\text{text}}$ is the number of
concatenated memory entries. The text is at daily resolution and is aligned
with the half-hourly load axis via a “same-day broadcasting’’ strategy: for any
half-hour index $\ell\in\{1,\dots,48\}$, the same $Y^{\text{text}}_{r,t}$ is used
for retrieval.

\subsubsection{Operations (cross-modal sparse retrieval and source gating)} \label{subsubsec:operations}
At the memory retrieval position, the query $R_{r,t,\ell}\!\in\!\mathbb{R}^{1\times d}$ and the text memory
$Y^{\text{text}}_{r,t}\!\in\!\mathbb{R}^{M_{\text{text}}\times d}$ are fed into the GSH attention with
$\alpha$-EntMax (see Eq.~\eqref{eq:gsh-attn}), producing the context vector conditioned on the text:
\begin{equation}
Z^{\text{text}}_{r,t,\ell}
=
\alpha\text{-EntMax}\!\bigl(\beta\,Q_{r,t,\ell}K_{r,t}^\top\bigr)\,V_{r,t},
\quad
Z^{\text{text}}_{r,t,\ell}\in\mathbb{R}^{1\times d}.
\label{eq:ztext-def}
\end{equation}
Here, $Z^{\text{text}}_{r,t,\ell}$ denotes the cross-modal contextual representation retrieved from the
external text memory at state $r$, natural day $t$, and half-hour index $\ell$, which is used for fusion
with the backbone representation or fed to the decoder to output load predictions;
$\beta>0$ is the temperature coefficient. 

\textcolor{red}{This step corresponds to \textbf{injection}: the retrieved text-aware representation $Z^{\text{text}}_{r,t,\ell}$ is injected at each half-hour position $\ell$ through the cross-attention readout and the subsequent fusion pathway, enabling position-dependent text influence within the forecasting window.}

The representation matrices $Q$, $K$, and $V$ are

\begin{equation}
\left\{
\begin{aligned}
Q_{r,t,\ell}\;&=\;R_{r,t,\ell}W_Q \in \mathbb{R}^{1\times d},\\
K_{r,t}\;&=\;Y^{\text{text}}_{r,t}W_K \in \mathbb{R}^{M_{\text{text}}\times d},\\
V_{r,t}\;&=\;Y^{\text{text}}_{r,t}W_V \in \mathbb{R}^{M_{\text{text}}\times d},
\end{aligned}
\right.
\label{eq:graft-core}
\end{equation}
where $W_Q,W_K,W_V\in\mathbb{R}^{d\times d}$ are learnable matrices. The weights produced by
$\alpha$-EntMax
\begin{equation}
  \pi_{r,t,\ell}=\alpha\text{-EntMax}\!\bigl(\beta\,Q_{r,t,\ell}K_{r,t}^\top\bigr)\in\mathbb{R}^{1\times M_{\text{text}}},
\end{equation}
form a row-sparse retrieval distribution that sums to $1$. Consequently,
Eq.~\eqref{eq:ztext-def} can be interpreted as a sparse convex combination over the rows of $V_{r,t}$,
thereby producing $Z^{\text{text}}_{r,t,\ell}$—the “text-aware’’ local context vector
\cite{peters2019entmax}.

To explicitly distinguish the contributions of the three sources at different times and regions,
we construct a source-gating vector $\gamma_{t,i}\in\Delta_3$ (corresponding sequentially to
News, Reddit, and Policy):
\begin{align}
\gamma &= \alpha\text{-EntMax}\!\Bigl(U\,[\,\bar{R}_{r,t}\,;\,\bar{Y}^{\text{text}}_{r,t}\,]\Bigr),\label{eq:gamma}\\
\tilde{Y}^{\text{text}}_{r,t} &= \gamma_{t,1} Y^{(\mathrm{news})}_{r,t}\ \oplus\ \gamma_{t,2} Y^{(\mathrm{reddit})}_{r,t}\ \oplus\ \gamma_{t,3} Y^{(\mathrm{policy})}_{r,t},\label{eq:mix-text}
\end{align}
where $U$ is a learnable matrix; $\bar{R}_{r,t}$ and $\bar{Y}^{\text{text}}_{r,t}$ denote the aggregated
representations along the half-hour daily dimension (e.g., mean or $R^\star$-based pooling), and
$\oplus$ denotes concatenation along the entry dimension.
Then $\tilde{Y}^{\text{text}}_{r,t}$ replaces $Y^{\text{text}}_{r,t}$ in Eq.~\eqref{eq:graft-core} to
complete the final retrieval. Leveraging the sparsity of $\alpha$-EntMax~\cite{peters2019entmax},
the gating weights and attention weights can be directly used to generate a “time–source’’ attribution
heatmap.

\begin{figure*}[b]
  \centering
  \begin{subfigure}{\textwidth}
    \includegraphics[width=\textwidth,height=0.30\textheight,keepaspectratio]{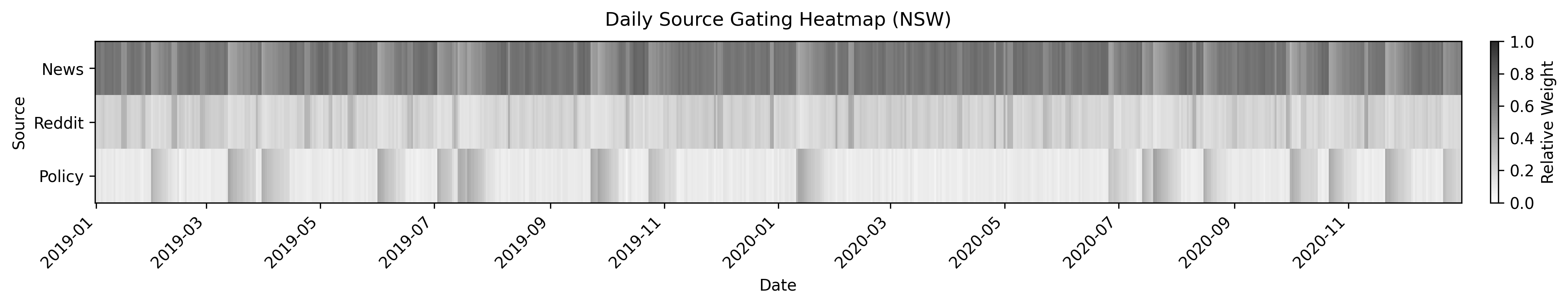}
    \caption{Time–Source Attribution ($\gamma$) – NSW}\label{fig:heatmap-NSW}
  \end{subfigure}\par\vspace{0.35em}
  \begin{subfigure}{\textwidth}
    \includegraphics[width=\textwidth,height=0.30\textheight,keepaspectratio]{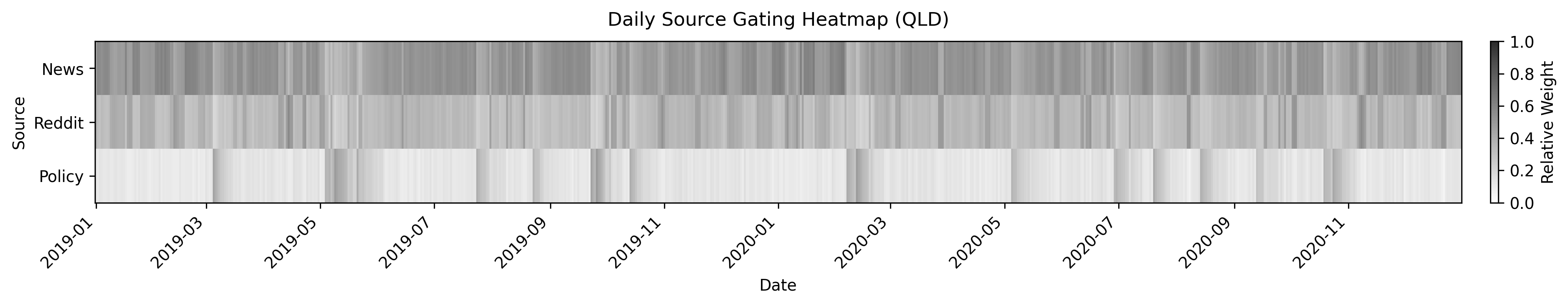}
    \caption{Time–Source Attribution ($\gamma$) – QLD}\label{fig:heatmap-QLD}
  \end{subfigure}\par\vspace{0.35em}
  \begin{subfigure}{\textwidth}
    \includegraphics[width=\textwidth,height=0.30\textheight,keepaspectratio]{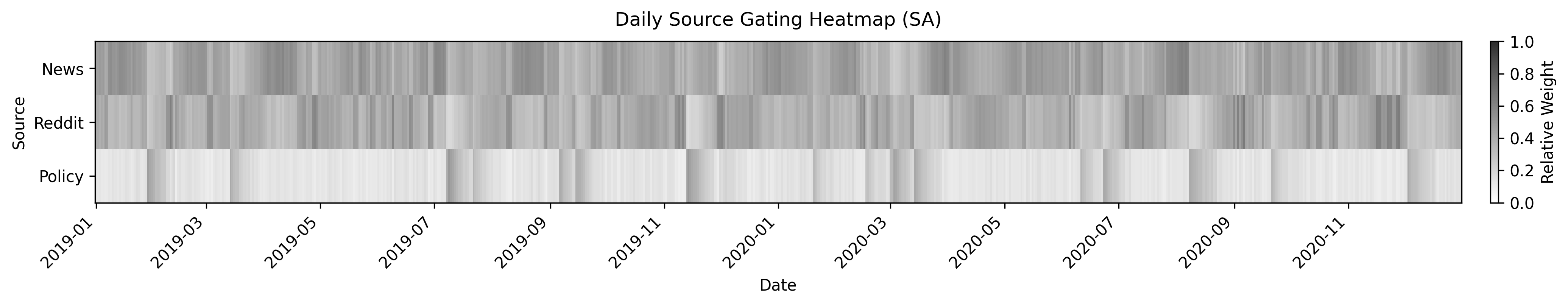}
    \caption{Time–Source Attribution ($\gamma$) – SA}\label{fig:heatmap-SA}
  \end{subfigure}\par\vspace{0.35em}
  \label{fig:heatmap-part1}
\end{figure*}

\begin{figure*}[t]
  \ContinuedFloat 
  \centering
  \begin{subfigure}{\textwidth}
    \includegraphics[width=\textwidth,height=0.30\textheight,keepaspectratio]{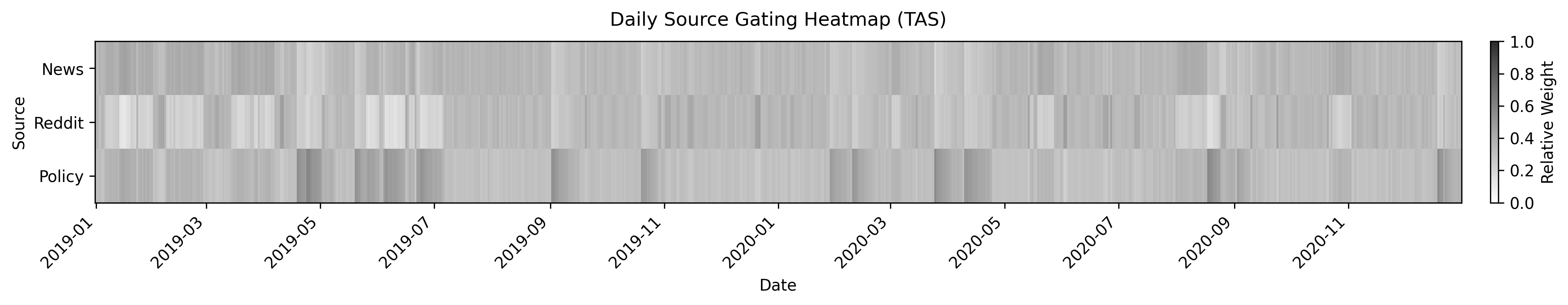}
    \caption{Time–Source Attribution ($\gamma$) – TAS}\label{fig:heatmap-TAS}
  \end{subfigure}\par\vspace{0.35em}
  \begin{subfigure}{\textwidth}
    \includegraphics[width=\textwidth,height=0.30\textheight,keepaspectratio]{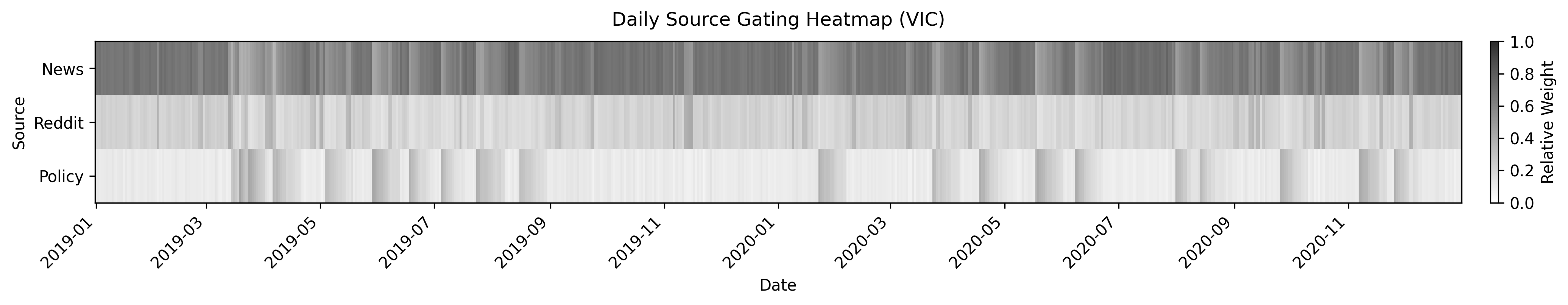}
    \caption{Time–Source Attribution ($\gamma$) – VIC}\label{fig:heatmap-VIC}
  \end{subfigure}
  \caption{Time–source attribution heatmaps.}
  \label{fig:heatmap-part2}
\end{figure*}

First, we compute the gating vector $\gamma_t$ for state $r$ and calendar day $t$
according to Eq.~\eqref{eq:gamma}.
Then, the half-hour attention weights (the $\alpha$-EntMax outputs in Eq.~\eqref{eq:graft-core})
are averaged or weighted-averaged along the daily dimension to obtain the daily contribution of
each source. The horizontal axis represents time, the vertical axis represents the three sources,
and darker colors indicate a higher relative weight of that source in cross-modal retrieval for
the day.
\sout{Figures~\ref{fig:heatmap-NSW}–\ref{fig:heatmap-VIC} show the visualizations for the five
states, followed by cross-state comparison and analysis:}
\textcolor{red}{Figures~\ref{fig:heatmap-NSW}–\ref{fig:heatmap-VIC} visualize the daily source-gating
weights for the five states over the period 2019--2020 (i.e., the training/validation
span in Section~\ref{subsec:experimental-setup}), which avoids using any test-year information and
covers multiple seasons and event phases. The following observations are therefore based on a
fixed, pre-test time range rather than a picked segment, and we use cautious,
comparative descriptions that can be further supported by quantitative summaries:}

\begin{enumerate}
\renewcommand{\labelenumi}{(\roman{enumi})}

\item Overall pattern.
\sout{The News bands in NSW and VIC are the most continuous and prominent
(see Figures~\ref{fig:heatmap-NSW} and~\ref{fig:heatmap-VIC}), indicating that mainstream media
coverage of extreme weather, market fluctuations, and system events is more concentrated. The
model tends to rely on News during stable to high-pressure periods.}
\textcolor{red}{Across all five states, News is frequently assigned a relatively higher
baseline weight, while Policy appears as sparser but occasionally stronger bursts,
and Reddit shows more intermittent fluctuations. In particular, NSW and VIC exhibit
longer stretches of higher News gating in many periods
(Figures~\ref{fig:heatmap-NSW} and~\ref{fig:heatmap-VIC}), suggesting that the model often
falls back to news-driven cues as a stable external reference in these states.}
\textcolor{red}{To quantify both the concentration and the dominance structure of the daily gating,
we define (i) the normalized entropy of $\gamma_{r,t} \in \Delta_3$ as
\begin{equation}
\bar{H}_r \;=\; \frac{1}{T}\sum_{t=1}^{T} \frac{-\sum_{i=1}^{3}\gamma_{r,t,i}\log\bigl(\gamma_{r,t,i}\bigr)}
{\log 3},
\label{eq:gamma-entropy}
\end{equation}
and (ii) the dominant-source fraction of source $i$ as
\begin{equation}
p_{r,i} \;=\; \frac{1}{T}\sum_{t=1}^{T}\mathbb{I}\!\left(i=\arg\max_{j\in\{1,2,3\}}\gamma_{r,t,j}\right).
\label{eq:gamma-dominance}
\end{equation}
Here, $T$ is the number of days in the displayed period, and $\mathbb{I}(\cdot)$ is the indicator function.
Based on these definitions, NSW and VIC show a concentrated and stable News preference:
the mean News weight is 0.629/0.630 and News is dominant on 99.3\%/99.2\% of days, with lower
normalized entropy $\bar{H}_r$=0.798/0.799.}
Policy appears as sparse but
relatively strong stripes across all five states, reflecting the “few but impactful, cross-day
decaying’’ nature of policy texts. Reddit shows intermittent pulses in each state, capturing
events related to holidays, sports, and daily activity rhythms.
\textcolor{red}{In contrast, TAS exhibits the most diversified gating behavior, with the highest
normalized entropy ($\bar{H}_r=0.984$) and a substantially larger mean Policy weight (0.346),
consistent with the more balanced visual pattern in Figure~\ref{fig:heatmap-TAS}.}

\item Typical differences.
\sout{In QLD, alternating peaks of News and Reddit occur near the
high-temperature season (Figure~\ref{fig:heatmap-QLD}), corresponding to increased rigidity in
air-conditioning usage and synchronized social discussions.}
\textcolor{red}{In QLD, the relative weights of News and Reddit alternate more visibly
during several seasonal phases (Figure~\ref{fig:heatmap-QLD}), consistent with the intuition that
intraday activity-driven discussions can become more informative when demand is more weather-sensitive.}
\textcolor{red}{Quantitatively, QLD has a higher mean Reddit weight (0.342) than NSW/VIC (0.217/0.217),
and Reddit becomes the dominant source on 9.2\% of days (while News remains dominant on 87.3\% of days),
which matches the visually observable alternations without overstating a full regime switch.}
In SA, Reddit weights are more active
and alternate with News (Figure~\ref{fig:heatmap-SA}), consistent with high rooftop PV penetration,
pronounced day–night reversals, and the resulting public attention and reporting patterns.
\textcolor{red}{This is consistent with SA’s more balanced dominance structure: the mean News/Reddit weights
are 0.441/0.388, and Reddit is dominant on 33.0\% of days (vs.\ 62.7\% for News), indicating substantially
stronger Reddit participation than in NSW/VIC.}
TAS exhibits a more balanced distribution among the three sources (Figure~\ref{fig:heatmap-TAS}),
where block-like enhancements in News and Policy often align with the cross-day effects of hydro
operation, maintenance, and price mechanism announcements.
\textcolor{red}{In TAS, Policy becomes dominant on 24.6\% of days (the largest among the five states),
which provides a quantitative counterpart to the block-like Policy enhancements observed in the heatmap.}

\item Mechanistic implications.
\sout{These patterns align with each state’s climate zone, load
composition, and interconnection system. The dual temperature drivers and dense media coverage in
NSW/VIC give News stronger explanatory power during peak periods; QLD’s high temperatures and
activity rhythms amplify the marginal information in Reddit; SA’s “midday dip–evening peak’’ is
more readily captured by both social media and news; sparse Policy stripes typically correspond to
interconnector events or rule changes.}
\textcolor{red}{These cross-state differences are qualitatively consistent with heterogeneity in
climate conditions, demand composition, and operational events, but we emphasize that the heatmaps
alone provide descriptive evidence. To make the attribution analysis more rigorous, the
same gating sequences can be summarized by quantitative statistics such as the dominant-source
fractions in Eq.~\eqref{eq:gamma-dominance} and the normalized entropy in Eq.~\eqref{eq:gamma-entropy},
which enables cross-state comparisons beyond visual inspection.}
\textcolor{red}{In addition, the persistence of dominant regimes can be quantified by run-length statistics:
NSW/VIC exhibit long News-dominant runs (mean run length 121.0/120.8 days), whereas TAS has shorter
News-dominant runs (mean 3.04 days) but longer Policy-dominant runs (mean 6.67 days), further supporting
that the learned gating adapts to state-specific external-signal structures.}
These results validate the necessity of source gating and
sparse retrieval introduced in Eqs.~\eqref{eq:gamma}–\eqref{eq:mix-text}: the model can use News as
baseline information during stable periods and adaptively amplify contributions from Reddit/Policy
during events, thus providing interpretable attribution across “when–which source–where’’.
\end{enumerate}

\textcolor{red}{The full set of quantitative statistics (including dominance,
entropy, and persistence metrics) are provided in Appendix~A for reproducibility and detailed inspection.}

\subsubsection{Decoding (multi-resolution convergence and pseudo-label enhancement)}
The cross-modal text representation $Z^{\text{text}}_{r,t,\ell}$ is fused with the multi-scale
representations obtained by STanHop along the temporal and sequential dimensions
(see Eq.~\eqref{eq:coarsen}), and then fed into a lightweight decoder head (either linear
or a small MLP) to predict $\hat{y}_{r,t,\ell}$. Under extreme operating conditions such as
accidents, warnings, or power curtailments, the \texttt{TuneMemory} mechanism in
Eq.~\eqref{eq:tune} can be activated to retrieve “exemplary response curves’’ from the
labeled external memory and concatenate them as auxiliary features to the decoder input,
thereby improving the robustness and calibration of peak and valley segments
\cite{wu2023stanhopsparsetandemhopfield}.

\subsubsection{Objective function and training details}
The main loss adopts the mean squared error $\mathcal{L}_{\mathrm{MSE}}$, with additional
quantile loss or Huber loss applied to the tail intervals, and an entropy regularization
term added to the source gating to prevent collapse:
\begin{equation}
\mathcal{L}
=
\mathcal{L}_{\mathrm{MSE}}(\hat{y},y)
+
\lambda_{\tau}\sum_{\tau\in\{0.1,0.9\}}\mathcal{L}_{\tau}(\hat{y},y)
+
\lambda_{\gamma}\,\mathrm{H}(\gamma),
\quad
\lambda_{\tau},\lambda_{\gamma}\ge 0.
\label{eq:loss}
\end{equation}
The same alignment and missing-mask rules are shared between the training and inference
phases. On the inference side, the \texttt{PlugMemory} mechanism in
Eq.~\eqref{eq:plug} can be used under frozen parameters to absorb the latest textual
memory with zero fine-tuning~\cite{wu2023stanhopsparsetandemhopfield}.

\subsection{Improvements and advantages over the original STanHOP}\label{subsec:advantage}

\begin{enumerate}
\renewcommand{\labelenumi}{(\roman{enumi})}
  \item Introduction of grid-aware cross-modal external memory.
  The original STanHOP relies solely on numerical time series and internal
  memories~\cite{wu2023stanhopsparsetandemhopfield}. In contrast, GRAFT transforms
  news, social media, and policy information into temporally aligned external
  memories and performs position-dependent injection at half-hour resolution via
  the sparse cross-modal retrieval in Eq.~\eqref{eq:graft-core}. This
  substantially enhances the model’s responsiveness to event-driven fluctuations,
  such as extreme heat, equipment maintenance, and pricing mechanism changes.
  
  \item Interpretability through source gating.
  The source-gating mechanism in Eqs.~\eqref{eq:gamma}–\eqref{eq:mix-text}
  adaptively determines, at the block level, \emph{which source to trust when},
  and together with the attention weights provides a three-dimensional
  attribution heatmap over time–source–region. This compensates for the limited
  interpretability of the original STanHOP with respect to external
  events~\cite{wu2023stanhopsparsetandemhopfield}.
  
  \item Robustness and calibration under extreme scenarios.
  Through pseudo-labeled retrieval in Eq.~\eqref{eq:tune} and tail-aware
  weighting in the loss function of Eq.~\eqref{eq:loss}, GRAFT significantly
  improves error performance and probabilistic calibration on peak and valley
  segments, thereby reducing the drift risk of the original model under rare or
  extreme operating conditions.
  
  \item Multi-resolution and computation-friendly design.
  Building on STanHOP’s Time/Series GSH and multi-resolution coarsening in
  Eq.~\eqref{eq:coarsen}~\cite{wu2023stanhopsparsetandemhopfield}, GRAFT maintains
  computational efficiency while introducing richer representations. Using
  learnable prototype pooling via $R^\star$ and entry-level concatenation, it
  controls the hidden dimension and memory size without incurring significant
  additional computational cost, thus meeting the efficiency requirements of
  half-hourly rolling forecasting.
  
  \item Engineering integration and reproducibility consistency.
  The text–load data are strictly aligned along the date–region dimensions, and
  the same masking and time-standardization rules are shared between training and
  inference. On the deployment side, textual memories can be updated on a daily
  basis without modifying the backbone parameters, enabling rapid implementation
  in practical systems and consistent, reproducible evaluation results
  across experiments~\cite{wu2023stanhopsparsetandemhopfield}.
\end{enumerate}

\subsection{Key points for implementation and reproduction}\label{subsec:impl}
The SBERT encoder~\cite{reimers2019sbert} is used on the encoding side to obtain
$\mathbf{e}^{(\cdot)}_{r,t}$, which is then aggregated on a daily basis;
alignment broadcasting and regional mapping follow the same rules as described
in the data section. The backbone adopts a ``TimeGSH $\rightarrow$ SeriesGSH''
stacked structure~\cite{wu2023stanhopsparsetandemhopfield}, with the text
memory inserted by default using Eq.~\eqref{eq:plug}; under extreme samples or
during replay learning, Eq.~\eqref{eq:tune} is activated. The experiments are
divided into three phases: training (2019--2020.09), validation
(2020.10--2020.12), and full-year testing (2021), with rolling evaluation and
extreme-day quantile metrics conducted accordingly.

\FloatBarrier

\section{Experiment}
\label{sec:exp}

\subsection{Experimental Setup}
\label{subsec:experimental-setup}

\textbf{Data and time partitioning.}
All state-level data are standardized to market time (AEST/AEDT) to ensure 48 half-hour steps per day.
The training set spans from 2019-01-01 to 2020-09-30; the validation set covers 2020-10-01 to 2020-12-31
(\sout{used only for offline monitoring and analysis, not for early stopping or model selection} \textcolor{red}{used only for offline monitoring and analysis, and not used for early stopping, model selection, or hyperparameter tuning}); and the test set
covers 2021-01-01 to 2021-12-31.
The input--output sliding window adopts $T_{\text{in}} = 336$ (approximately 7 days) and $T_{\text{out}} = 48$ (1 day), with
temporal blocking performed at a segment length of 12 (6 hours per segment).

\textbf{External source alignment and usage strategy.}
\sout{The textual external information is aggregated along the ``calendar day--state'' dimension, aligned with the last natural day of the input window, and mapped to the 30-minute resolution through ``same-day broadcasting.'' Missing or delayed entries are handled via explicit masking.}
\textcolor{red}{The textual external information is aggregated along the ``calendar day--state'' dimension and aligned with the last natural day of the input window (day $d\!-\!1$ when forecasting day $d$). The resulting daily embedding is then broadcast to the 30-minute resolution \emph{within that same historical day} (i.e., the day $d\!-\!1$ slots only). Therefore, each day-ahead forecast for day $d$ is trained and generated strictly from past load observations and external sources available up to the end of day $d\!-\!1$, and no information leakage occurs. Missing or delayed entries are handled via explicit masking.}
The external-source switch is encoded as follows:
``0'' indicates that no external textual information is used (baseline);
``1'' uses news information;
``2'' uses social-media (Reddit) information;
``3'' uses policy information;
``123'' uses the combination of news, social media and policy information.
When multiple sources are selected, the daily embeddings are evenly sliced along the last dimension according to source order and injected separately into corresponding cross-attention branches.

\textbf{Model configuration.}
The backbone adopts a dual-channel temporal--cross-sequence architecture based on STanHOP, incorporating external textual information through cross-attention fusion.
The key hyperparameter settings are as follows:
hidden dimension $d_{\mathrm{model}} = 512$;
feedforward dimension $d_{\mathrm{ff}} = 1024$;
number of heads $n_{\mathrm{heads}} = 8$;
number of encoder layers $e_{\mathrm{layers}} = 3$;
dropout rate $= 0.1$;
segment length $= 12$.
When multi-source textual information is enabled, the concatenated total dimension is evenly divided among the selected source branches.

\textbf{Optimization and training.}
The optimizer is Adam, with an initial learning rate $1 \times 10^{-4}$, batch size 32, 20 training
epochs, and the loss function is the MSE corrected as in Eq.~(28).
L2-norm gradient clipping (threshold 1.0) is applied to improve numerical stability; if NaN/Inf is detected
in the model outputs or the loss, the corresponding mini-batch will be skipped.
During training, the batch mean of the training-set MSE per epoch is printed for comparison.

\textbf{Hardware and reproducibility.}
Example environment for reference: GeForce RTX 4060 8GB, Driver 572.16, CUDA 11.8,
and PyTorch version 2.7.1, which can support the default hyperparameters used for
training and evaluation in this paper.

\subsection{Main experiments compared with external sources}\label{subsec:main-exp}

\sout{To intuitively illustrate the effects of external sources at different temporal scales and enable fair comparison under unified conditions, this section evaluates 15 representative ``\textbf{state $\times$ time-scale}'' windows ($5 \times 3 = 15$). These cover three forecasting tasks: very short-term (VSTLF, $W = 16$ points), short-term (STLF, $W = 48$ points), and medium-term (MTLF, $W = 2880$ points).}
\textcolor{red}{
Throughout this paper, all models are trained and evaluated with a fixed day-ahead horizon of $T_{\text{out}} = 48$. The other evaluation windows ($W = 16$ for VSTLF and $W = 2880$ for MTLF) are derived from the same 48-step outputs by slicing (taking the first 16 points) and by concatenating consecutive 48-step forecasts, rather than training separate models with different output lengths. These case windows cover three evaluation horizons: very short-term (VSTLF, $W = 16$ points), short-term (STLF, $W = 48$ points), and medium-term (MTLF, $W = 2880$ points).
Specifically, for each test timestamp $t$ in 2021, we generate a day-ahead forecast $\hat{\mathbf{y}}_{t:t+47}$ and aggregate errors across all such rolling forecasts to obtain the reported metrics. 
}
\textcolor{red}{Building upon the use of rolling windows, to provide an intuitive and interpretable view of how different external
sources affect forecasts at multiple temporal scales under a unified protocol,
this section visualizes 15 representative ``state $\times$ time-scale'' case windows 
($5 \times 3 = 15$) \emph{randomly selected for qualitative illustration only}. Furthermore, it will not affect subsequent global metrics with rolling windows.}

Each subplot overlays the true load (TRUE) with predictions from the STanHOP baseline (NoExt, i.e., without external sources) and from the GRAFT configurations using News / Reddit / Policy / All (the three single-source and combined-source setups). This allows observation of how external information refines local trajectory shapes and amplitudes.

For quantitative evaluation, absolute errors (RMSE, MAE) and relative errors (MAPE, sMAPE) are reported to measure predictive accuracy. The Skill metric is used to quantify improvement over a statistical baseline, while RankRMSE and Wins are adopted to assess each source’s overall performance level and robustness across states and time scales.

\begin{figure*}[t]
  \centering
  \begin{subfigure}[t]{0.32\textwidth}
    \centering
    \includegraphics[width=\textwidth]{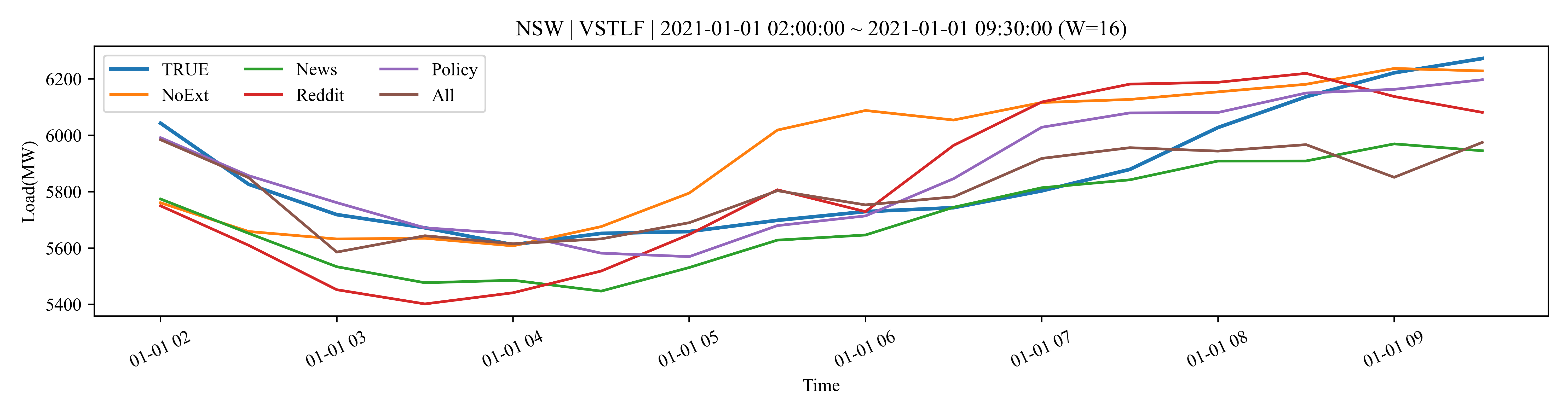}
    \caption{NSW | VSTLF}
  \end{subfigure}
  \begin{subfigure}[t]{0.32\textwidth}
    \centering
    \includegraphics[width=\textwidth]{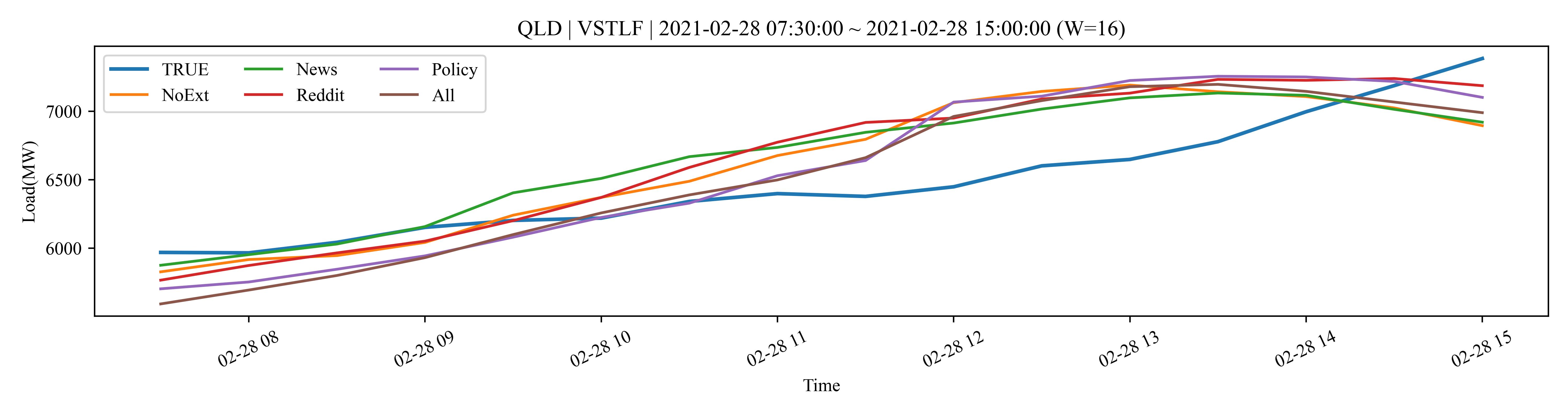}
    \caption{QLD | VSTLF}
  \end{subfigure}
  \begin{subfigure}[t]{0.32\textwidth}
    \centering
    \includegraphics[width=\textwidth]{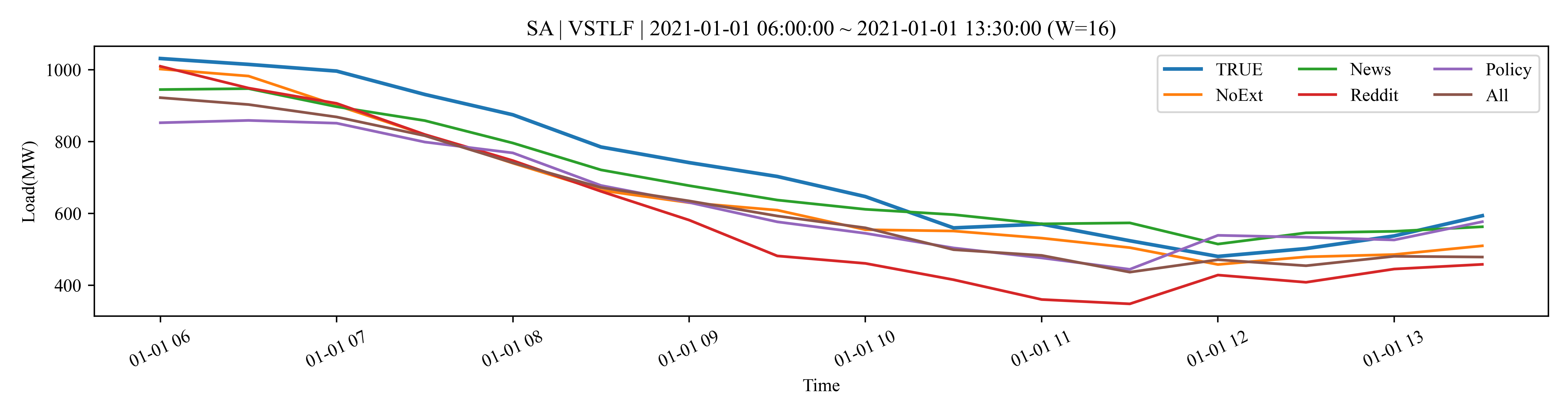}
    \caption{SA | VSTLF}
  \end{subfigure}

  \vspace{0.4em}

  \begin{subfigure}[t]{0.32\textwidth}
    \centering
    \includegraphics[width=\textwidth]{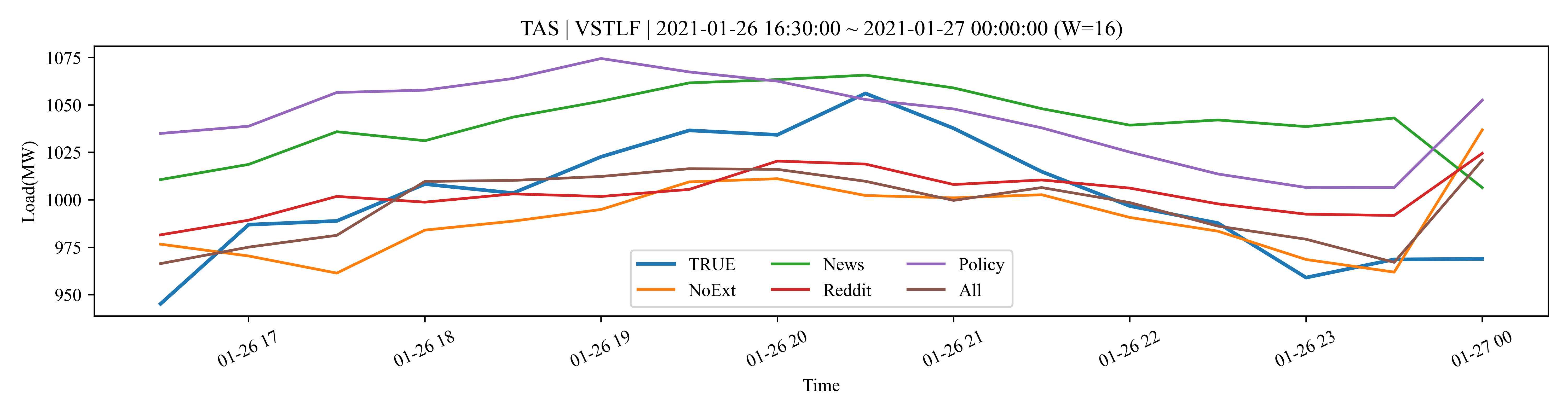}
    \caption{TAS | VSTLF}
  \end{subfigure}
  \begin{subfigure}[t]{0.32\textwidth}
    \centering
    \includegraphics[width=\textwidth]{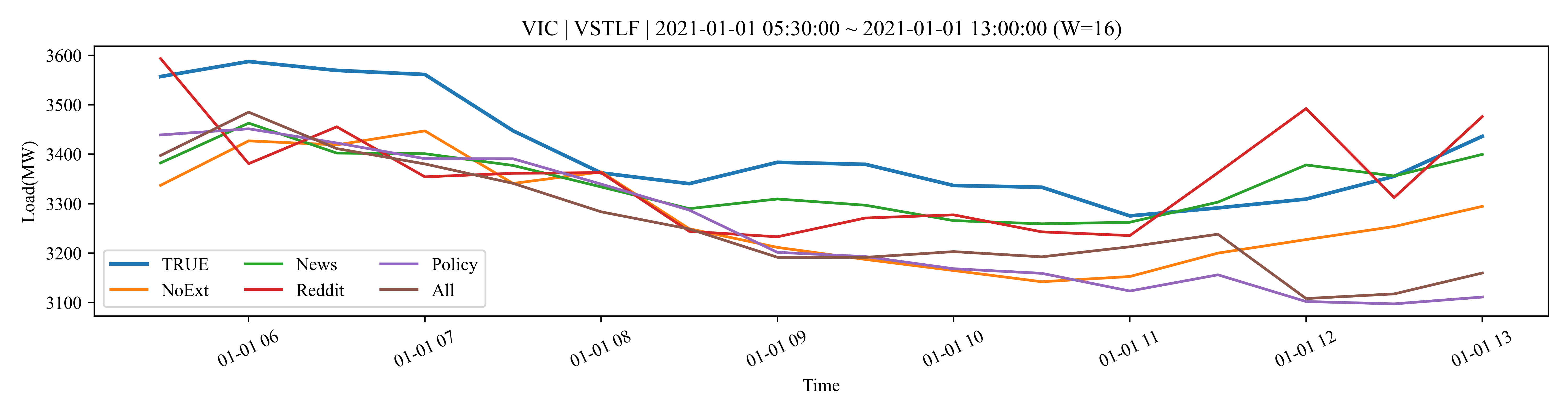}
    \caption{VIC | VSTLF}
  \end{subfigure}

  \caption{Comparison of five-state load curves under different information-source configurations in very short-term windows (VSTLF, $W\!=\!16$).}
  \label{fig:case-vstlf}
\end{figure*}

\begin{figure*}[t]
  \centering
  \begin{subfigure}[t]{0.32\textwidth}
    \centering
    \includegraphics[width=\textwidth]{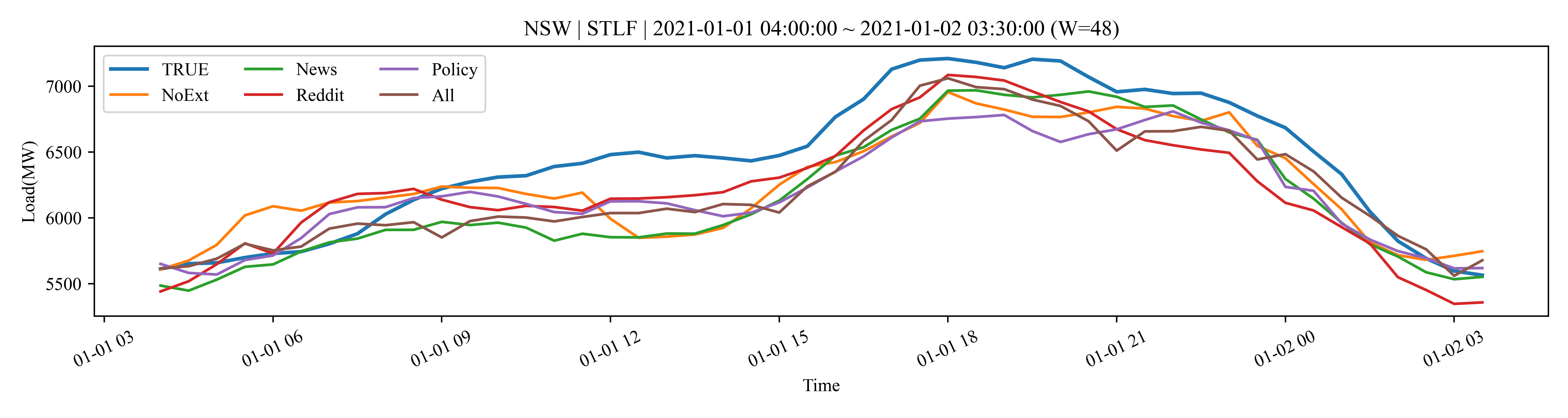}
    \caption{NSW | STLF}
  \end{subfigure}
  \begin{subfigure}[t]{0.32\textwidth}
    \centering
    \includegraphics[width=\textwidth]{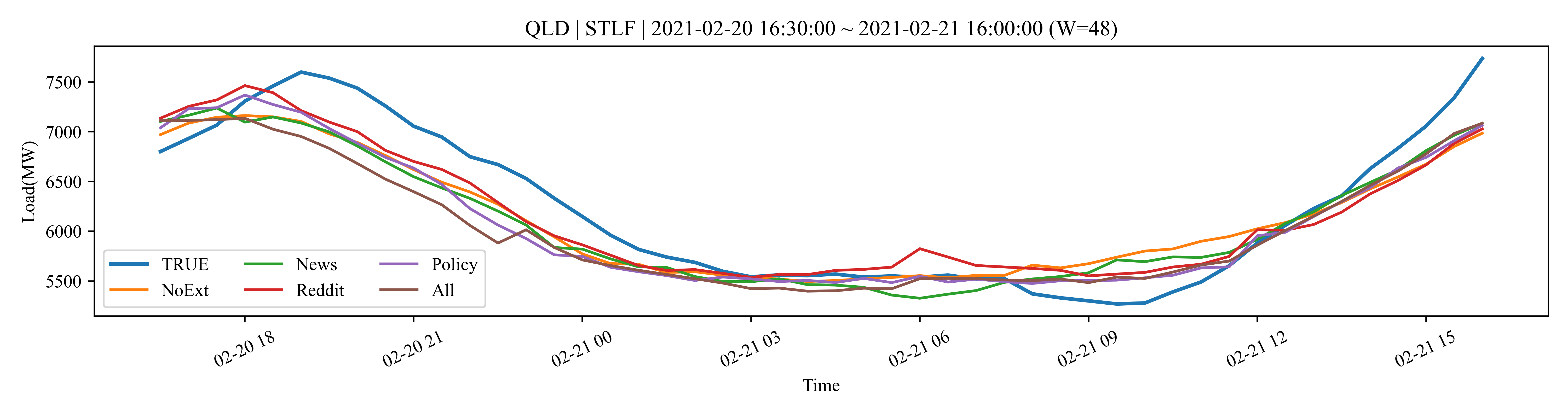}
    \caption{QLD | STLF}
  \end{subfigure}
  \begin{subfigure}[t]{0.32\textwidth}
    \centering
    \includegraphics[width=\textwidth]{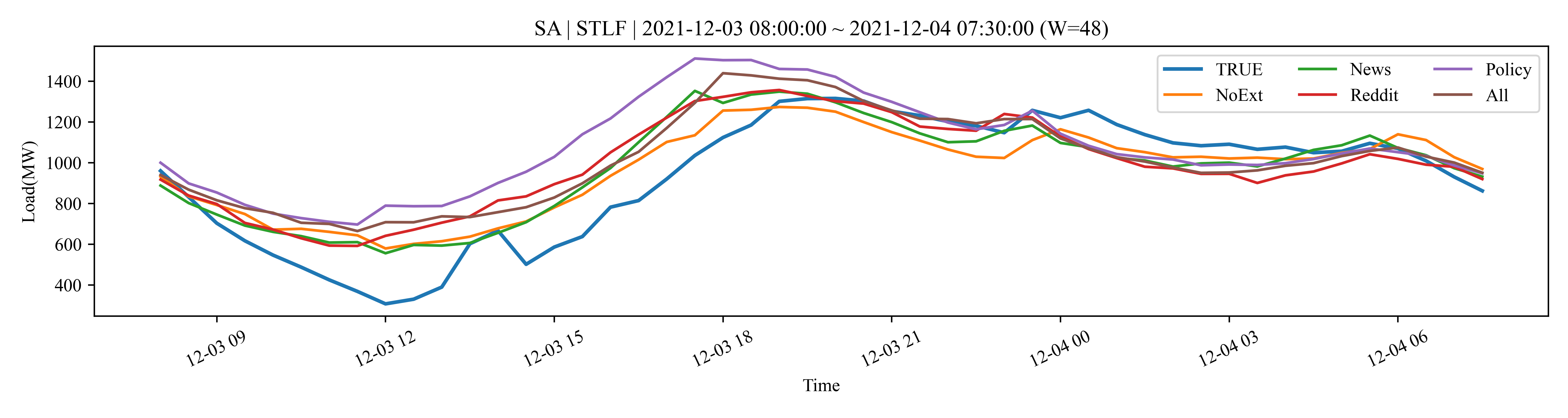}
    \caption{SA | STLF}
  \end{subfigure}

  \vspace{0.4em}

  \begin{subfigure}[t]{0.32\textwidth}
    \centering
    \includegraphics[width=\textwidth]{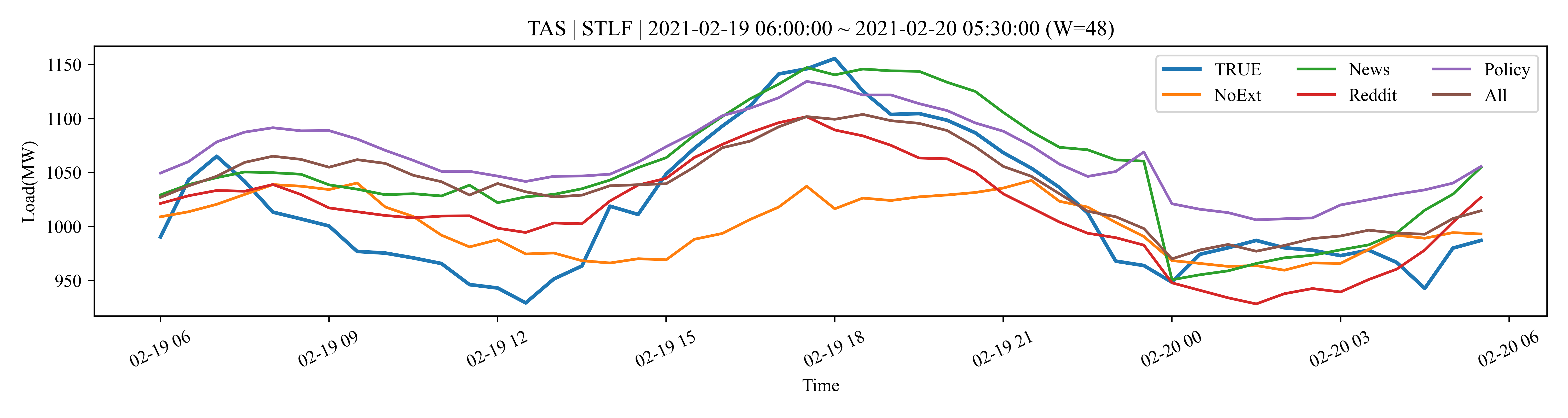}
    \caption{TAS | STLF}
  \end{subfigure}
  \begin{subfigure}[t]{0.32\textwidth}
    \centering
    \includegraphics[width=\textwidth]{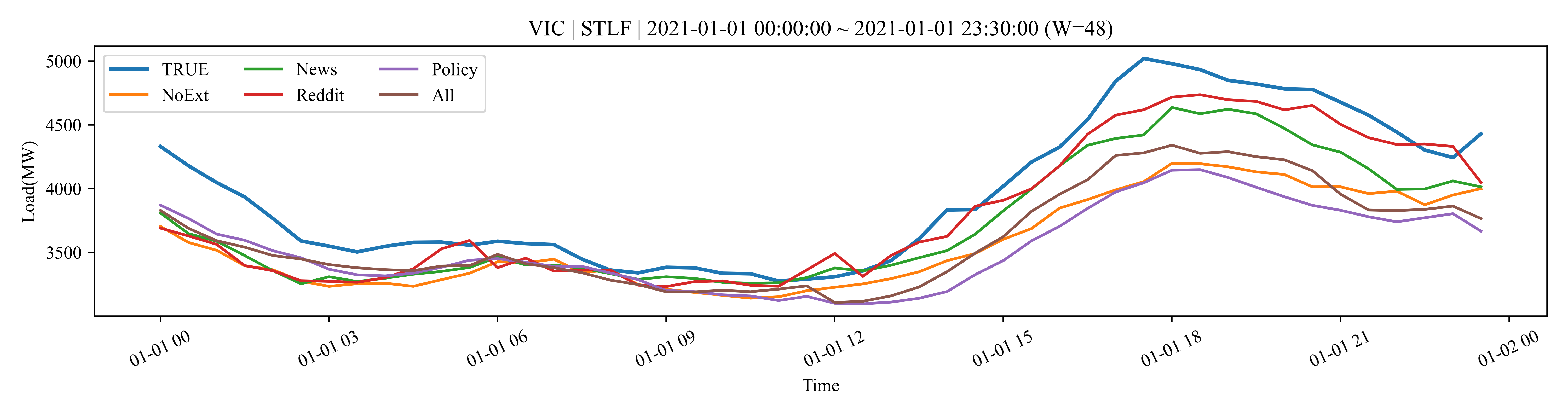}
    \caption{VIC | STLF}
  \end{subfigure}

  \caption{Comparison of 24-hour load curves of five states under different information-source configurations in short-term windows (STLF, $W\!=\!48$).}
  \label{fig:case-stlf}
\end{figure*}

\begin{figure*}[t]
  \centering
  \begin{subfigure}[t]{0.32\textwidth}
    \centering
    \includegraphics[width=\textwidth]{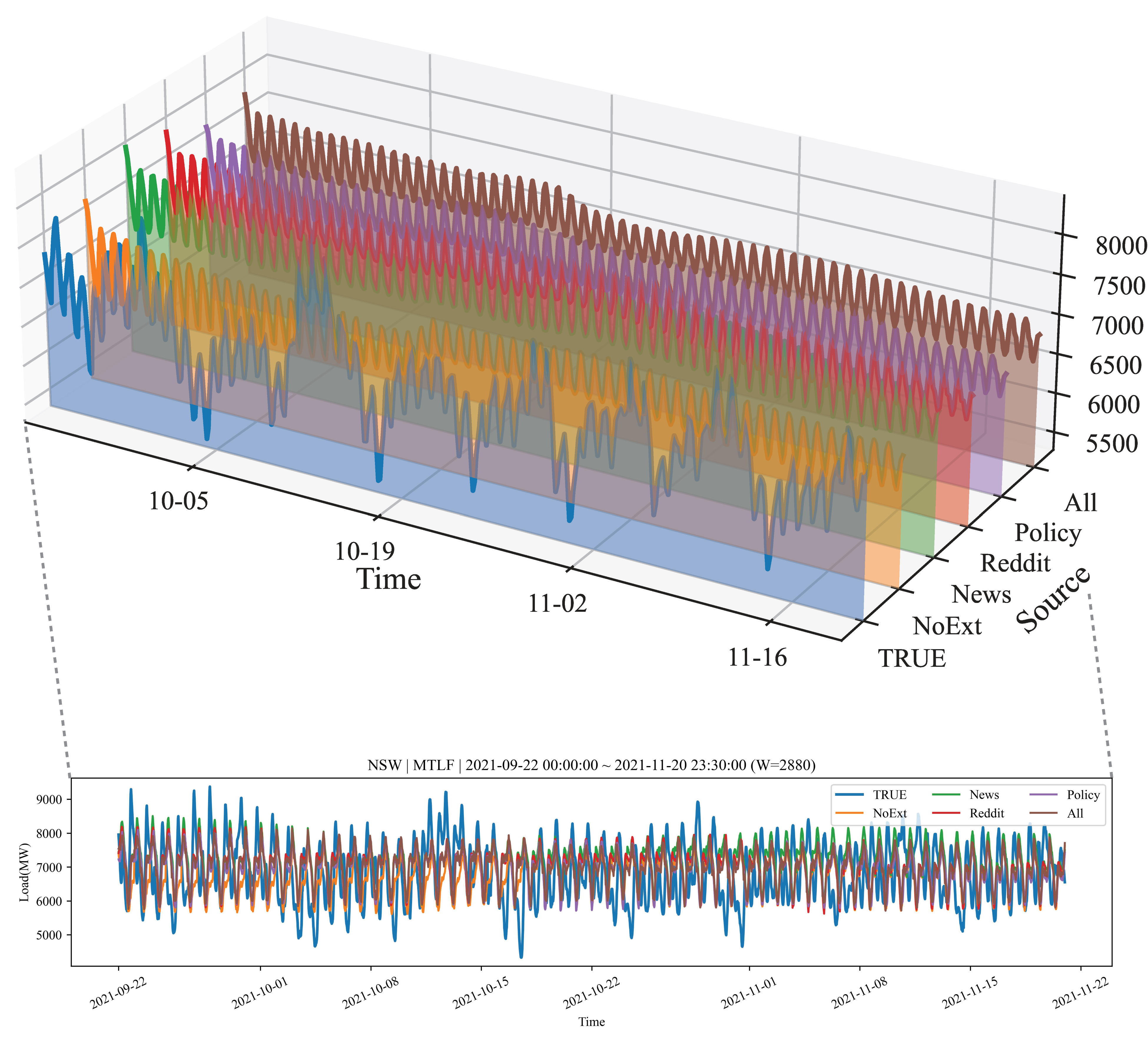}
    \caption{NSW | MTLF}
  \end{subfigure}
  \begin{subfigure}[t]{0.32\textwidth}
    \centering
    \includegraphics[width=\textwidth]{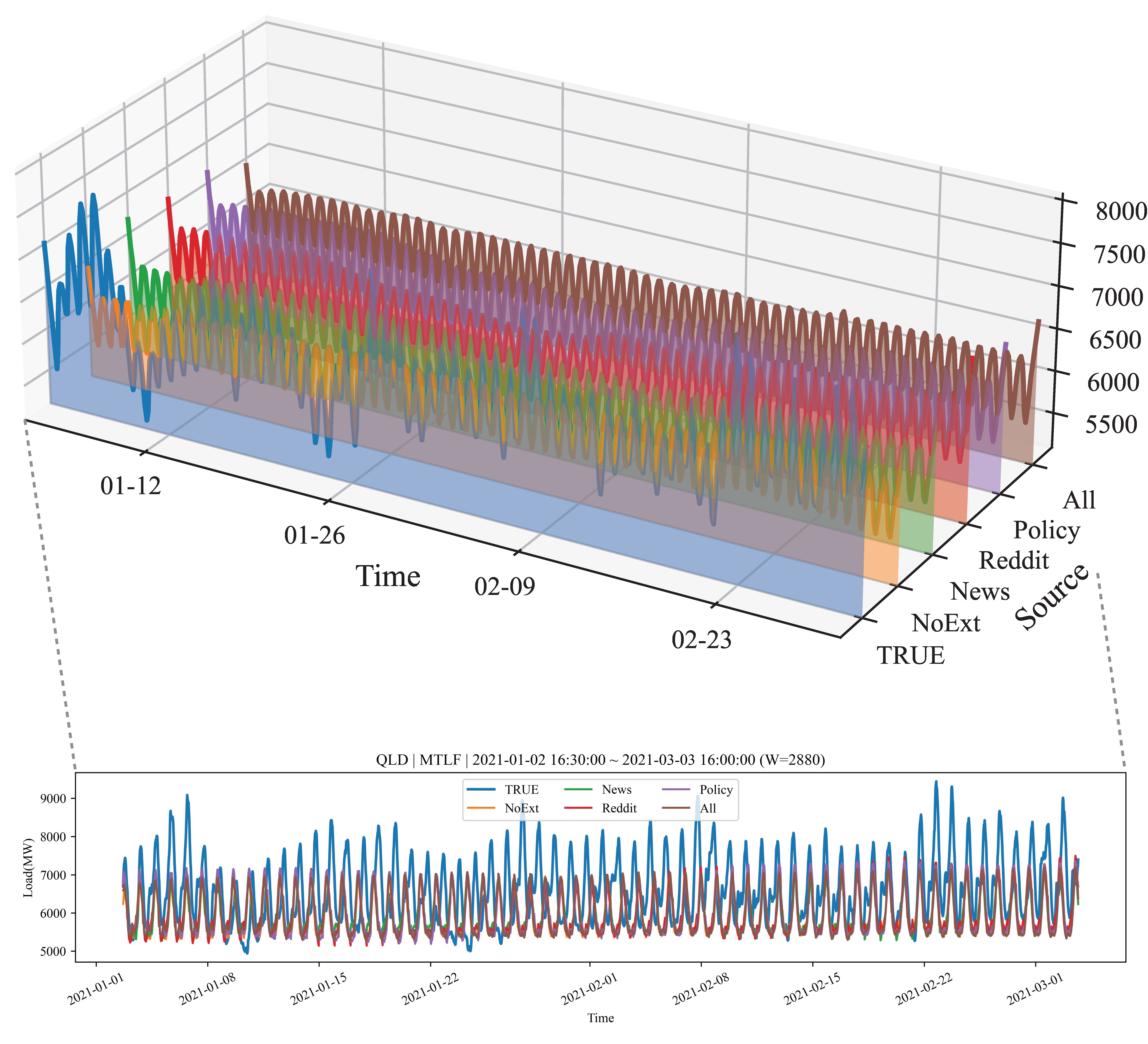}
    \caption{QLD | MTLF}
  \end{subfigure}
  \begin{subfigure}[t]{0.32\textwidth}
    \centering
    \includegraphics[width=\textwidth]{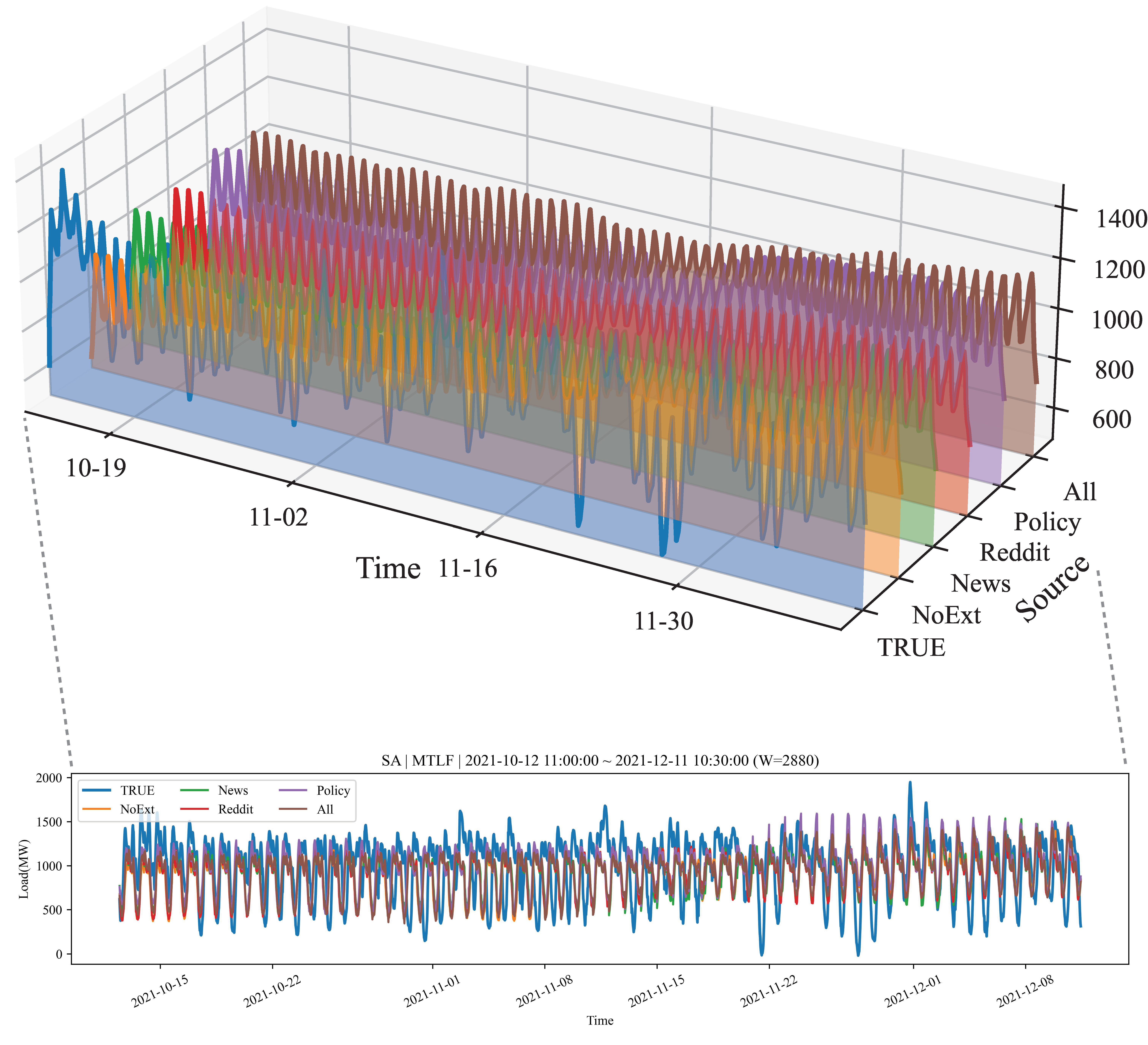}
    \caption{SA | MTLF}
  \end{subfigure}

  \vspace{0.4em}

  \begin{subfigure}[t]{0.32\textwidth}
    \centering
    \includegraphics[width=\textwidth]{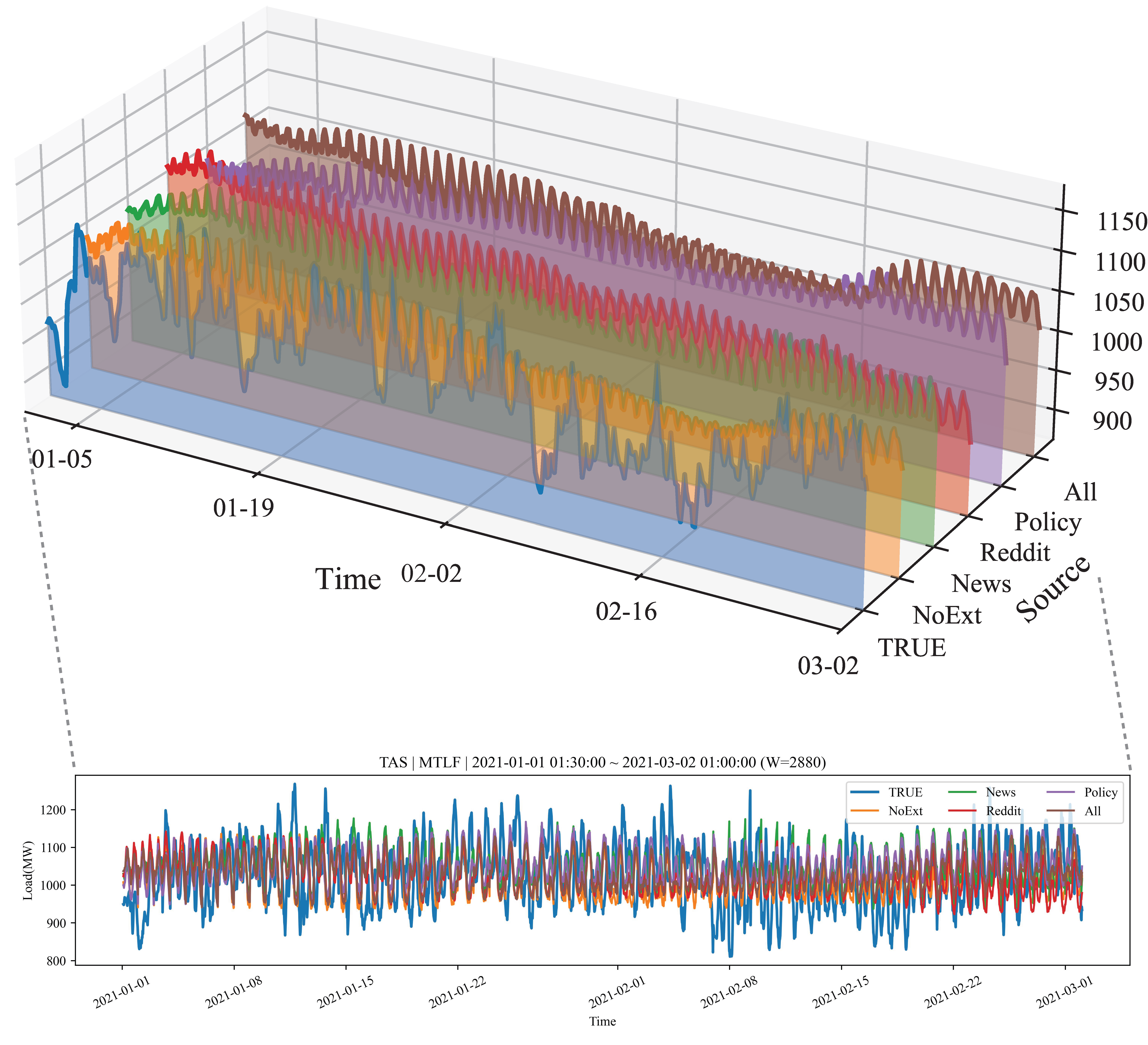}
    \caption{TAS | MTLF}
  \end{subfigure}
  \begin{subfigure}[t]{0.32\textwidth}
    \centering
    \includegraphics[width=\textwidth]{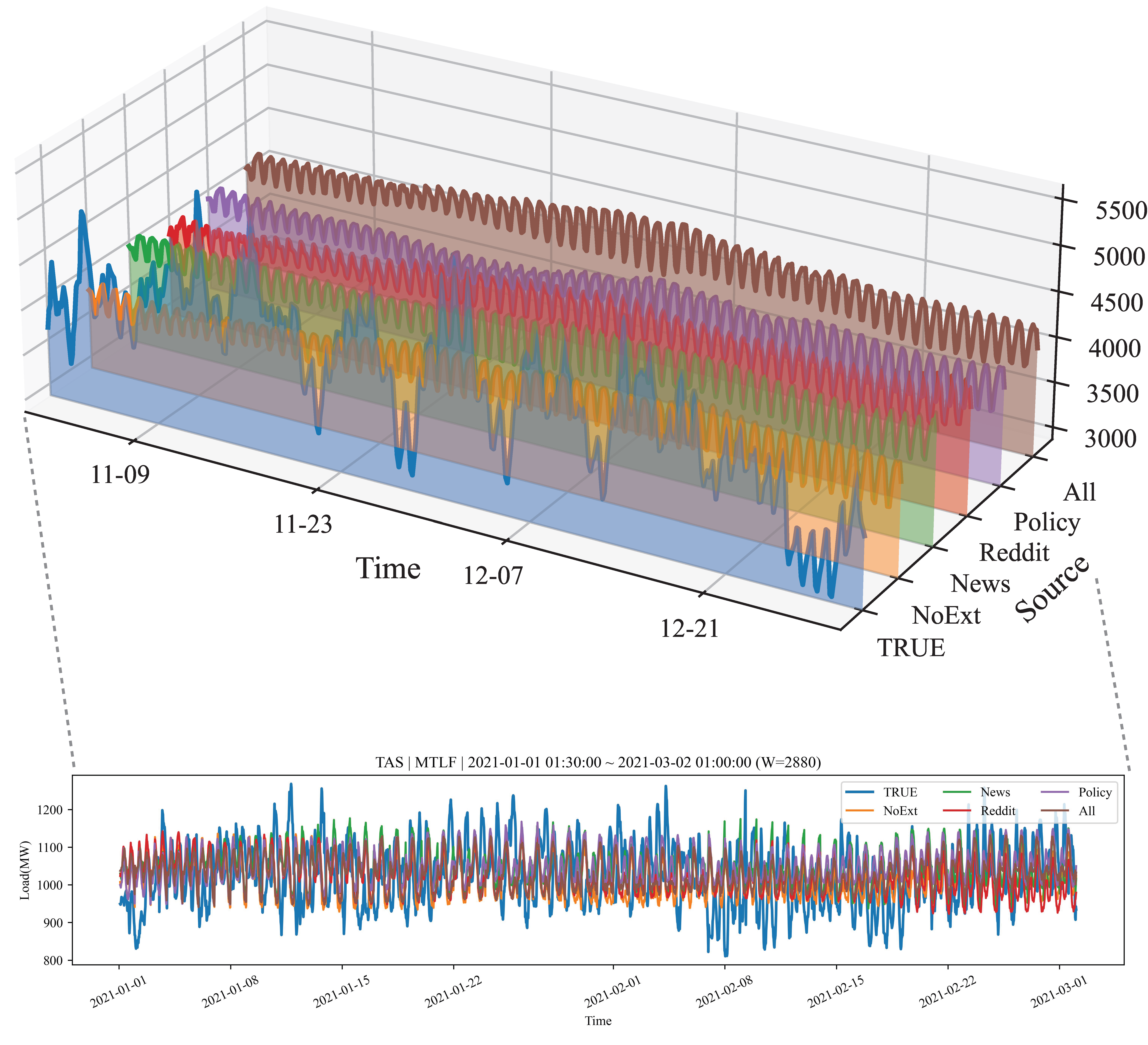}
    \caption{VIC | MTLF}
  \end{subfigure}

  \caption{Three-dimensional waterfall plots and two-dimensional projections of 60-day loads in five states under medium-term windows (MTLF, $W\!=\!2880$), illustrating the overall performance of different information-source configurations over long sequences.}
  \label{fig:case-mtlf}
\end{figure*}

\subsubsection{Skill, RankRMSE, and Wins indicators}

Let there be $N$ evaluation tasks, each corresponding to a “state $\times$ time-scale” combination.
For any information source $k$, its RMSE on the $i$-th task is denoted as $\mathrm{RMSE}_{k,i}$.
Let the evaluation window length of the $i$-th task be $W_i$ (number of steps, with a step size of
30 minutes), and let the set of evaluation time indices be $\Omega_i$, such that $|\Omega_i| = W_i$.

To quantify the \emph{degree of error reduction relative to a unified statistical baseline}, we adopt
a model-independent statistical benchmark in the Skill metric. For each task $i$, denote the
statistical-baseline prediction at time $t \in \Omega_i$ by $\hat y^{\text{stat}}_t$.
The corresponding baseline RMSE is defined as
\begin{equation}
\mathrm{RMSE}_{\text{stat},i}
    = \sqrt{\frac{1}{W_i} \sum_{t \in \Omega_i}
        \bigl(y_t - \hat y^{\text{stat}}_t\bigr)^2},
\label{eq:rmse_stat}
\end{equation}
where $y_t$ is the true load.
The construction rule of this statistical baseline depends on the evaluation-window length:
when the window is short ($W_i \le 12$, corresponding to ultra–short-term tasks of no more
than about 6 hours), a persistence baseline is used, i.e.,
\begin{equation}
\hat y^{\text{stat}}_t = y_{t-1}, \qquad t \in \Omega_i;
\end{equation}
when the window is longer ($W_i > 12$, corresponding to daily or longer time scales),
a seasonal naïve baseline is adopted: for each time point $t$, according to its day of week
$\mathrm{dow}(t) \in \{0,\dots,6\}$ and half-hour slot
$\mathrm{slot}(t) \in \{0,\dots,47\}$, we compute the historical average load
$\mu_{\mathrm{dow}(t),\,\mathrm{slot}(t)}$ over all samples with the same
“day-of-week + half-hour slot”, and define
\begin{equation}
\hat y^{\text{stat}}_t = \mu_{\mathrm{dow}(t),\,\mathrm{slot}(t)}, \qquad t \in \Omega_i.
\end{equation}

Based on this, the local Skill of information source $k$ on task $i$, measuring its error
reduction relative to the statistical baseline, is defined as
\begin{equation}
\mathrm{Skill}_{k,i}
    = 1 - \frac{\mathrm{RMSE}_{k,i}}{\mathrm{RMSE}_{\text{stat},i}}.
\label{eq:skill_local}
\end{equation}
When $\mathrm{Skill}_{k,i} > 0$, the RMSE of source $k$ on task $i$ is smaller than that of
the statistical baseline, indicating a relative improvement; when $\mathrm{Skill}_{k,i} < 0$,
the performance still falls behind the statistical baseline.
The value reported in the tables is the average Skill across tasks:
\begin{equation}
\mathrm{Skill}_k
    = \frac{1}{N}\sum_{i=1}^{N} \mathrm{Skill}_{k,i},
\label{eq:skill}
\end{equation}
which can be interpreted as “the average RMSE improvement ratio relative to the statistical
baseline prediction.”

It is worth emphasizing that two different types of baselines coexist in this paper, each at a
different level:

\begin{enumerate}
  \renewcommand{\labelenumi}{(\roman{enumi})}
  \item \textbf{Statistical baseline for Skill calculation.}
        This is the statistical baseline $\mathrm{RMSE}_{\text{stat},i}$ defined above,
        which is applied uniformly to all information sources
        $k \in \{0,1,2,3,123\}$.
  \item \textbf{Model-family baseline (NoExt).}
        This refers to the NoExt configuration within the STanHOP framework, i.e., the
        model without any external textual information (corresponding to information source
        $k = 0$ in the code). Its prediction sequence is denoted by $\hat y_t^{\text{NoExt}}$,
        and its task-level error $\mathrm{RMSE}_{0,i}$ also enters Eqs.~\eqref{eq:skill_local}
        and \eqref{eq:skill}. However, the denominator in the Skill metric is always
        $\mathrm{RMSE}_{\text{stat},i}$ rather than the error of NoExt.
        In subsequent significance tests and comparative analysis, when we discuss
        “improvement relative to NoExt”, the NoExt model is treated as the \emph{model
        baseline} within the same family, used to measure the relative gain brought by
        external information sources. This notion of model baseline is independent of the
        statistical baseline used in the Skill computation.
\end{enumerate}

On this basis, we further introduce a ranking-type metric and a count-type metric.
Within each task, the RMSE values of all information sources are sorted in ascending order,
yielding the rank of source $k$, denoted by
$\mathrm{rank}_{k,i}\in\{1,\dots,K\}$ (sources with identical RMSE share the same rank).
The RankRMSE is then defined as
\begin{equation}
\mathrm{RankRMSE}_k
    = \frac{1}{N}\sum_{i=1}^{N} \mathrm{rank}_{k,i},
\label{eq:rankrmse}
\end{equation}
which reflects the average RMSE ranking of information source $k$ across all tasks.
A smaller $\mathrm{RankRMSE}_k$ indicates that the source attains a better and more
stable overall ranking across different states and time scales.

Accordingly, Wins measures the number of tasks on which a source achieves (or ties for)
the lowest RMSE. Let $\mathbb{I}(\cdot)$ denote the indicator function; then
\begin{equation}
\mathrm{Wins}_k
    = \sum_{i=1}^{N} \mathbb{I}\!\left(
        \mathrm{RMSE}_{k,i} = \min_{j} \mathrm{RMSE}_{j,i}
      \right),
\label{eq:wins}
\end{equation}
where a tie for the best RMSE is counted as a win for all tied sources on that task.
A larger $\mathrm{Wins}_k$ means that the corresponding source achieves—or ties for—the
best RMSE on a greater number of “state $\times$ time-scale” combinations.

In the subsequent analysis, Skill, RankRMSE, and Wins are adopted as the core indicators
for comparing the strengths and weaknesses of different external-source configurations.
On the one hand, absolute errors (RMSE / MAE) vary significantly with load magnitude across
states and time scales, so naive averaging can easily be dominated by high-load scenarios.
Skill normalizes performance relative to the statistical baseline, while RankRMSE achieves
scale normalization across tasks through a “rank-first, then-average” procedure, thus
providing a clearer picture of overall relative performance.
On the other hand, Wins emphasizes how often a configuration genuinely outperforms all
others. It helps distinguish different performance patterns—for example, configurations
that are “occasionally excellent but unstable” (high Wins but large RankRMSE) versus those
that are “consistently stable and strong overall” (low RankRMSE, high Skill, and moderate
or high Wins). Taken together, these indicators offer a more comprehensive assessment of
the robustness and leading performance of external information sources across states and
time scales.

\vspace{0.2em}
\begin{table*}[t]
\centering
\tiny
\setlength{\tabcolsep}{4pt}
\renewcommand{\arraystretch}{0.9}
\caption{Overall comparison by time scale and external source (average over 5 states)}
\label{tab:all-scales-all-sources}

\begin{subtable}[t]{\textwidth}
\centering
\caption{Very short-term horizon (hourly)}
\label{tab:hour-all-sources-mape}
\begin{tabular}{lcrrrrrrrr}
\toprule
Source & N & \textbf{RMSE} & MAE & \textbf{MAPE} & sMAPE & Skill & RankRMSE & \textbf{Wins} \\
\midrule
News   & 5 & \textbf{136.781} & \textbf{116.228} & \textbf{3.967} & \textbf{3.9730} & \textbf{0.8510} & \textbf{2.0} & \textbf{3} \\
All    & 5 & 146.970 & 124.690 & 4.900 & 5.1112 & 0.8473 & 2.4 & 1 \\
NoExt  & 5 & 155.778 & 129.394 & 4.433 & 4.5444 & 0.8427 & 3.4 & 1 \\
Reddit & 5 & 159.008 & 136.253 & 6.222 & 6.7479 & 0.8232 & 3.4 & 0 \\
Policy & 5 & 147.888 & 124.355 & 5.333 & 5.4905 & 0.8233 & 3.8 & 1 \\
\bottomrule
\end{tabular}
\end{subtable}
\hfill
\begin{subtable}[t]{\textwidth}
\centering
\caption{Short-term horizon (daily)}
\label{tab:day-all-sources-mape}
\begin{tabular}{lcrrrrrrrr}
\toprule
Source & N & \textbf{RMSE} & MAE & \textbf{MAPE} & sMAPE & Skill & RankRMSE & \textbf{Wins} \\
\midrule
Reddit & 5 & \textbf{201.736} & \textbf{170.002} & 7.410 & 6.7180 & \textbf{0.6754} & \textbf{1.6} & \textbf{3} \\
News   & 5 & 222.573 & 184.320 & \textbf{7.008} & \textbf{6.5747} & 0.6516 & 3.0 & 0 \\
All    & 5 & 254.803 & 209.040 & 8.541 & 7.7392 & 0.6021 & 3.0 & 1 \\
NoExt  & 5 & 256.517 & 211.975 & 7.806 & 7.4287 & 0.6134 & 3.2 & 1 \\
Policy & 5 & 287.581 & 232.257 & 10.582 & 9.2152 & 0.5315 & 4.2 & 0 \\
\bottomrule
\end{tabular}
\end{subtable}

\vspace{0.6em}

\begin{subtable}[t]{\textwidth}
\centering
\caption{Medium-term horizon (monthly)}
\label{tab:month-all-sources-mape}
\begin{tabular}{lcrrrrrrrr}
\toprule
Source & N & \textbf{RMSE} & MAE & \textbf{MAPE} & sMAPE & Skill & RankRMSE & \textbf{Wins} \\
\midrule
All    & 5 & 697.743 & 560.474 & 23.297 & \textbf{17.2945} & -0.1850 & \textbf{2.0} & \textbf{3} \\
Reddit & 5 & \textbf{694.159} & \textbf{555.914} & \textbf{23.269} & 17.5215 & \textbf{-0.1842} & 2.6 & 1 \\
News   & 5 & 714.764 & 574.472 & 23.434 & 17.5295 & -0.2023 & 2.6 & 1 \\
Policy & 5 & 708.984 & 570.297 & 24.732 & 17.6324 & -0.2064 & 3.2 & 1 \\
NoExt  & 5 & 727.394 & 588.172 & 24.566 & 18.3282 & -0.2330 & 4.6 & 0 \\
\bottomrule
\end{tabular}
\end{subtable}
\hfill
\begin{subtable}[t]{\textwidth}
\centering
\caption{Overall summary across external sources}
\label{tab:overall-sources-mape}
\begin{tabular}{lcrrrrrrr}
\toprule
Source & N & \textbf{RMSE} & MAE & \textbf{MAPE} & sMAPE & Skill & RankRMSE & \textbf{Wins} \\
\midrule
All    & 15 & 366.506 & 298.068 & 12.246 & 10.0483 & 0.4215 & \textbf{2.467} & \textbf{5} \\
Reddit & 15 & \textbf{351.634} & \textbf{287.390} & \textcolor{red}{12.300} & 10.3290 & \textbf{0.4381} & 2.533 & 4 \\
News   & 15 & 358.039 & 291.674 & \textbf{\textcolor{red}{11.470}} & \textbf{9.3591} & 0.4334 & 2.533 & 4 \\
NoExt  & 15 & 379.896 & 309.307 & 12.269 & 10.1004 & 0.4077 & 3.733 & 1 \\
Policy & 15 & 381.484 & 308.970 & \textcolor{red}{13.549} & 10.7794 & 0.3882 & 3.733 & 1 \\
\bottomrule
\end{tabular}
\end{subtable}

\end{table*}

\subsubsection{Very-Short-Term (hourly) results}

In the Very-Short-Term rolling forecasting scenario (VSTLF, $W\!=\!16$), we focus both on the local shape within the 8-hour window and on the aggregated error metrics across states.
As shown in Figure~\ref{fig:case-vstlf}, the representative windows of the five states generally exhibit monotonic upward or downward trends.
The prediction trajectories of News and All are overall the closest to TRUE:
in states such as NSW and SA, they provide more accurate amplitude and phase alignment for local troughs and the subsequent rebound segments.
By contrast, NoExt frequently shows overestimation or underestimation around peaks,
manifesting as peak misalignment or delayed recovery,
while Reddit and Policy exhibit slight over-suppression or over-amplification in different states.
These phenomena indicate that external sources mainly act at the hourly scale through fine-grained peak–valley calibration,
with news and multi-source fusion being the most sensitive to event-triggered local disturbances.

Table~\ref{tab:hour-all-sources-mape} summarizes the hourly results across the five states,
which is highly consistent with the visualization conclusions from Figure~\ref{fig:case-vstlf}:
\begin{enumerate}
  \renewcommand{\labelenumi}{(\roman{enumi})}
\item News achieves the best overall performance at the hourly level, reducing RMSE by about 12\% and MAE by about 10\% relative to NoExt. Together with the corrections to local troughs and rebound segments in Figure~\ref{fig:case-vstlf}, this suggests that news reports capture event-scale shocks such as accidents, maintenance and weather warnings, thereby enabling timely, fine-grained shape adjustments.
\item All ranks second, yielding about 4\%--6\% reductions in RMSE and MAE compared with NoExt, and outperforming Reddit and Policy in terms of both RankRMSE and Skill. As shown in Figure~\ref{fig:case-vstlf}, All consistently suppresses the peak deviations of NoExt across states; although it may not be optimal in every single window, its cross-state performance is more stable.
\item At the hourly level, NoExt attains slightly lower MAPE than All (by about 0.6\%), while Reddit and Policy show higher MAPE. In Figure~\ref{fig:case-vstlf}, some states present “relatively large but small-in-absolute-value” errors in low-load segments, indicating that external sources may amplify percentage errors at small-load points when correcting peaks and valleys. Therefore, in engineering practice, if peak fitting and waveform alignment are the primary concern, News or All should be preferred; if relative errors in low-load segments are more critical, the weights of external sources should be moderately constrained at the hourly scale.
\end{enumerate}

\subsubsection{Short-Term (daily) results}

In Short-Term (daily aggregated) forecasting, external text mainly reflects intra-day rhythms and cross-day behavioral patterns.
Figure~\ref{fig:case-stlf} shows the 24-hour load curves under different configurations across the five states.
All external-source combinations capture the general shape of “early-morning trough–daytime rise–evening peak,”
but the heights and positions of peaks and valleys clearly depend on the external sources.
For example, in NSW and QLD, Reddit and All substantially reduce the peak deviations of NoExt near the evening peak,
bringing the predicted peaks closer to the true trajectories.
In SA and TAS, News and Policy provide smoother corrections to the daytime plateau and nighttime decline,
avoiding the “overestimation–underestimation” alternation that occasionally appears in NoExt.
This suggests that social media is better at capturing behavior-driven intra-day rhythms—such as holidays,
sports events, performances and travel peaks—whereas news is more effective at attenuating sharp spikes
and reducing relative percentage errors.

Table~\ref{tab:day-all-sources-mape} quantitatively summarizes the results at the daily scale,
which is consistent with the shape observations in Figure~\ref{fig:case-stlf}:
\begin{enumerate}
  \renewcommand{\labelenumi}{(\roman{enumi})}
\item Reddit performs best overall at the daily scale, reducing RMSE by about 21\% and MAE by about 20\% compared with NoExt.
Together with the more accurate evening peaks and weekday rises in Figure~\ref{fig:case-stlf},
this indicates that Reddit effectively captures collective behavioral rhythms and guides the model
to make more reasonable elevation and suppression of the daily curves.
\item News shows a slight advantage in relative error, with MAPE about 0.8\% lower than that of NoExt,
while also yielding noticeable improvements in RMSE and MAE.
Consistent with the smoother peak–valley transitions in Figure~\ref{fig:case-stlf},
news weakens isolated spikes without excessively lifting the overall level,
and therefore performs better on relative-error metrics.
\item All maintains good robustness across different states: it is slightly better than NoExt in RMSE
but slightly higher in MAPE, while ranking near the top in both RankRMSE and Wins.
The overall suppression of peaks and valleys in Figure~\ref{fig:case-stlf} indicates that,
at the daily scale, simple linear fusion of multiple sources may slightly amplify relative errors on certain days.
This suggests that GRAFT should employ gating and temporal-decay mechanisms so that News and Reddit
can adaptively allocate their weights according to the scenario context in daily forecasting.
\end{enumerate}

\subsubsection{Medium-Term (monthly) results}

The Medium-Term (monthly) forecasting task evaluates how well each information source captures seasonality, policy cycles, and other slow-varying factors.
Figure~\ref{fig:case-mtlf} presents 60-day 3D waterfall plots and 2D projections for the five states:
All and Reddit exhibit overall envelopes that align more closely with the TRUE load, maintaining better phase and amplitude consistency during weekend fluctuations and high-load weekday periods.
By contrast, NoExt shows noticeable ``underestimation'' or ``overestimation'' in some peak periods, indicating that without external slow-varying information, STanHOP lacks sufficient sensitivity to long-term trends and seasonal perturbations.
News and Policy mainly provide fine-grained adjustments to slow trends and structural changes, with the degree of local correction varying across states.
Overall, the multi-source and social-media-based configurations clearly outperform the no-external-source setup in fitting long-term trends and seasonal variations.

This observation is consistent with the quantitative results in Table~\ref{tab:month-all-sources-mape}:
\begin{enumerate}
  \renewcommand{\labelenumi}{(\roman{enumi})}
  \item Reddit performs slightly better in terms of absolute error, reducing RMSE by about 5\% and MAPE by about 5\% compared with NoExt.
  Combined with the more accurate long-period envelope aligned with TRUE in Figure~\ref{fig:case-mtlf}, this indicates that Reddit not only captures short-term behavioral rhythms but also shows sensitivity to slow-varying monthly-scale factors related to long-term activities, consumption, and travel habits.
  \item All and Reddit both significantly outperform NoExt, with All reducing RMSE and MAPE by about 4\%--5\% relative to NoExt, while Reddit achieves roughly a further 1\% improvement beyond All.
  This is consistent with the ``flattened'' or ``exaggerated'' peaks of NoExt in Figure~\ref{fig:case-mtlf}, and suggests that when maintenance schedules, policy cycles, and seasonal structural adjustments are present, relying solely on historical load cannot adequately represent slow variables, whereas social-media signals and multi-source fusion provide effective complements.
  \item All exhibits stronger robustness and cross-state consistency, ranking among the top in both RankRMSE and Wins, and improving Skill by about 0.05 over NoExt.
  Together with the more balanced envelopes of All across states in Figure~\ref{fig:case-mtlf}, this suggests that at the monthly scale, multi-source fusion further integrates the institutional and structural information from News and Policy on top of Reddit, thereby achieving more balanced long-term fitting across regions.
\end{enumerate}

{\color{red}
\subsubsection{Monthly error decomposition and cycle-wise effect }
\label{sec:tas-monthly-analysis}

To assess whether specific months disproportionately influence the overall MTLF performance and whether such variation relates to text-driven adjustments, we further decompose forecasting errors at the month level.
Figure~\ref{fig:tas-monthly} reports the monthly MAPE and RMSE trajectories on TAS in 2021 under different external-source configurations (NoExt, News, Reddit, Policy, and All).

Across the year, both MAPE and RMSE exhibit a clear seasonal pattern rather than being dominated by a single outlier month: errors gradually increase from early months toward mid-year and then decrease in the last quarter.
Notably, the inter-configuration gaps become more pronounced during the high-error season (roughly mid-year), indicating that the marginal impact of text-driven injection is not constant across cycles.
In contrast, during the low-error months (late-year), the curves tend to converge, suggesting that when the load dynamics are more stable, text-driven adjustments play a weaker role and the performance differences across sources shrink.
Therefore, the overall MTLF results are shaped by a systematic cycle-wise error structure, and the effect of textual interventions is modulated by seasonality, which explains why certain periods contribute more to the aggregate metrics.

\begin{figure}[t]
    \centering
    \begin{subfigure}[t]{0.98\linewidth}
        \centering
        \includegraphics[width=\linewidth]{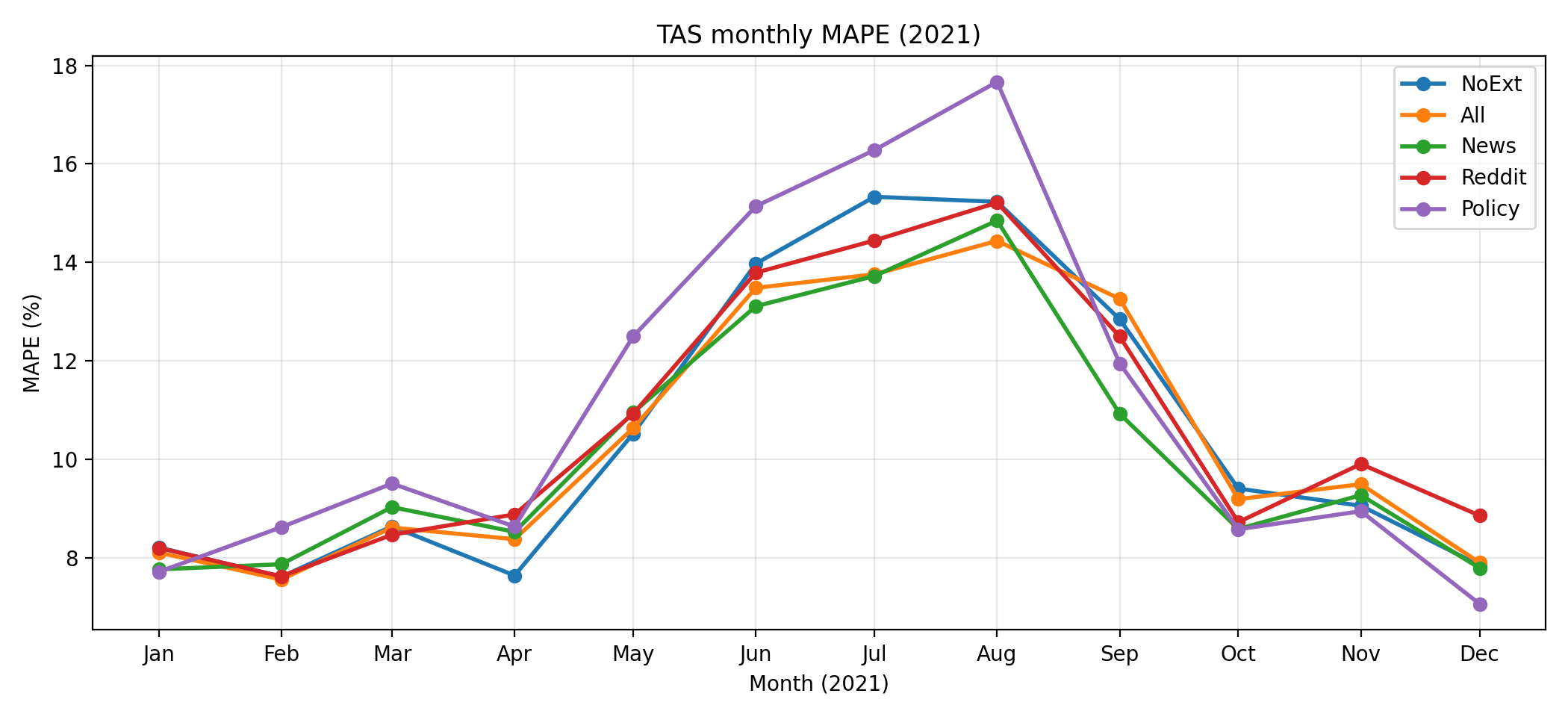}
        \caption{\textcolor{red}{Monthly MAPE (\%).}}
        \label{fig:tas-monthly-mape}
    \end{subfigure}
    \vspace{0.35em}
    \begin{subfigure}[t]{0.98\linewidth}
        \centering
        \includegraphics[width=\linewidth]{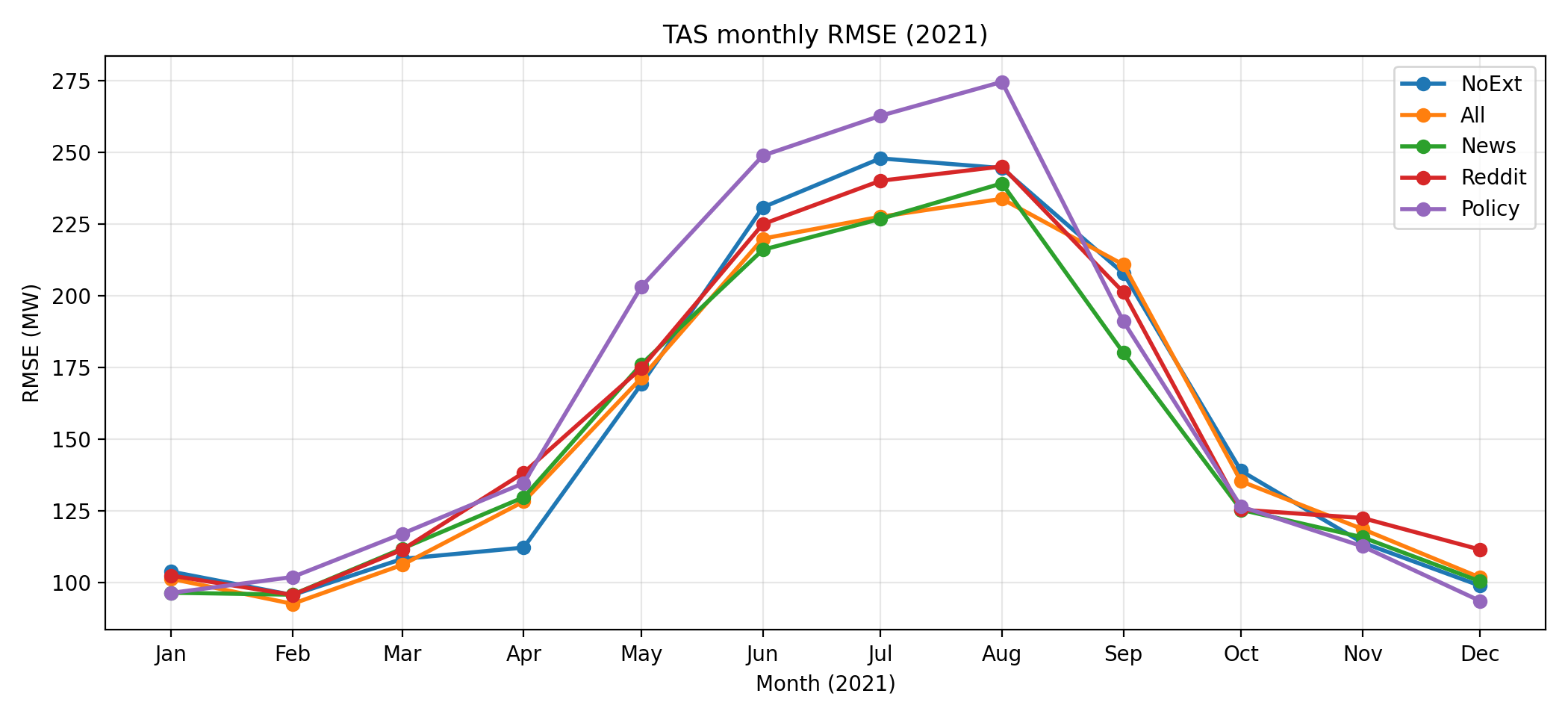}
        \caption{\textcolor{red}{Monthly RMSE (MW).}}
        \label{fig:tas-monthly-rmse}
    \end{subfigure}
    \caption{\textcolor{red}{TAS monthly error trajectories in 2021 under different external-source configurations.}}
    \label{fig:tas-monthly}
\end{figure}

\subsubsection{Subsequent significance tests (Diebold--Mariano)}
To verify that the observed gains are not incidental, we conduct Diebold--Mariano (DM) tests using the per-step loss series within each evaluation window. For each state and forecasting setting, we
test the null hypothesis of equal predictive accuracy against the one-sided alternative that the candidate configuration yields smaller expected loss.
Table~\ref{tab:dm-significance-counts} reports the resulting p-values against (i) a seasonal statistical baseline (\texttt{pDM\_seasonal}) and (ii) the model-family baseline NoExt (\texttt{pDM\_NoExt}).
Across all $60$ comparisons (5 states $\times$ 3 horizons $\times$ 4 text-source settings), every text-aware configuration significantly outperforms the seasonal baseline at the $5\%$ level ($60/60$ with $\texttt{pDM\_seasonal}<0.05$), confirming that incorporating structured textual signals yields non-trivial accuracy improvements beyond naive seasonality.
When compared to NoExt, $31/60$ cases achieve statistical significance ($\texttt{pDM\_NoExt}<0.05$), with the strongest evidence concentrated in longer windows where the test has higher power (MTLF: $17/20$ significant), while short windows are less frequently significant (STLF: $8/20$), consistent with the smaller sample size ($W{=}48$) limiting detectability even when absolute-error improvements are visible.

\begin{table}[t]
\centering
\caption{\textcolor{red}{DM-test significance counts at the $5\%$ level (one-sided).}}
\label{tab:dm-significance-counts}
\setlength{\tabcolsep}{4pt}
\renewcommand{\arraystretch}{1.15}
\begin{tabular}{lcc}
\hline
\textcolor{red}{\textbf{Target}} & \textcolor{red}{\textbf{Sig./Total}} & \textcolor{red}{\textbf{MTLF/STLF/VSTLF}} \\
\hline
\textcolor{red}{Seasonal (\texttt{pDM\_seasonal}$<0.05$)} & \textcolor{red}{$60/60$} & \textcolor{red}{$20/20 \; / \; 20/20 \; / \; 20/20$} \\
\textcolor{red}{NoExt (\texttt{pDM\_NoExt}$<0.05$)}       & \textcolor{red}{$31/60$} & \textcolor{red}{$17/20 \; / \; 8/20 \; / \; 6/20$} \\
\hline
\end{tabular}
\end{table}

}

\FloatBarrier

{\color{red}
\subsection{Ablation study}\label{subsec:ablation}

\textcolor{red}{To justify the architectural choices and quantify the contribution of external information, we conduct two complementary ablations: (i) source ablation, which isolates the marginal utility of each external source within the same model family, and (ii) architectural ablation, which removes key structural components while keeping all other settings fixed.}

\textcolor{red}{\subsubsection{Source ablation: external sources}\label{subsubsec:ablation-source}}
\textcolor{red}{We emphasize that \textit{NoExt} is the ablated variant of GRAFT with all textual pathways disabled, which degenerates exactly to the STanHOP backbone. Therefore, comparing NoExt against \{News, Reddit, Policy, All\} constitutes a within-family source ablation under identical data partitions and training protocol.}
\textcolor{red}{The source-ablation results are reported in Table~\ref{tab:all-scales-all-sources}, which summarizes performance by horizon (hourly/daily/monthly) and the overall aggregation across the 15 groups (3 horizons $\times$ 5 states).}
\textcolor{red}{Based on Table~\ref{tab:all-scales-all-sources} and the corresponding case-window visualizations, we observe:}
\begin{enumerate}
  \renewcommand{\labelenumi}{\textcolor{red}{(\roman{enumi})}}
  \item \textcolor{red}{\textbf{Overall robustness (absolute errors).} In the aggregated results (Table~\ref{tab:overall-sources-mape}), All reduces RMSE from 379.896 (NoExt) to 366.506 and MAE from 309.307 (NoExt) to 298.068, and achieves the highest Wins (5), indicating that multi-source fusion provides the most robust default improvement in absolute-error metrics across states and horizons.}

  \item \textcolor{red}{\textbf{Proportional errors (MAPE/sMAPE).} News yields the lowest aggregated MAPE (9.359 vs.\ 12.269 for NoExt), and Reddit is also substantially better (10.329). This suggests that single sources (especially News/Reddit) can deliver stronger proportional corrections, complementing the robustness of All in RMSE/MAE.}

  \item \textcolor{red}{\textbf{Scale-dependent behavior.} At the hourly scale, News provides the best RMSE/MAE/MAPE; at the daily scale, Reddit achieves the best RMSE/MAE while News achieves the best MAPE; at the monthly scale, both All/Reddit outperform NoExt in RMSE/MAE and sMAPE.}
\end{enumerate}
\textcolor{red}{These patterns indicate that different sources contribute at different horizons and motivate multi-source fusion as a robust default configuration.}

\textcolor{red}{\subsubsection{Architectural ablation: structural design}\label{subsubsec:ablation-arch}}

\textcolor{red}{Beyond source ablation (Table~\ref{tab:all-scales-all-sources}), we further verify that the gains rely on the proposed operations in Section~\ref{subsubsec:operations}, including cross-modal sparse retrieval (Eqs.~\eqref{eq:ztext-def}--\eqref{eq:graft-core}) and source gating (Eqs.~\eqref{eq:gamma}--\eqref{eq:mix-text}). All architectural ablations keep the same data partitioning, $T_{\text{in}}=336$, $T_{\text{out}}=48$, the same backbone, and the same optimization settings.}

\textcolor{red}{In the full GRAFT, the cross-modal contextual vector is retrieved by}
\begin{equation}
\textcolor{red}{
Z^{\text{text}}_{r,t,\ell}
=
\alpha\text{-EntMax}\!\bigl(\beta\,Q_{r,t,\ell}K_{r,t}^\top\bigr)\,V_{r,t},
\quad
Z^{\text{text}}_{r,t,\ell}\in\mathbb{R}^{1\times d},
}
\label{eq:ablation-ztext}
\end{equation}
\textcolor{red}{with $(Q_{r,t,\ell},K_{r,t},V_{r,t})$ defined as in Eq.~\eqref{eq:graft-core}. For multi-source inputs, a source-gating vector $\gamma\in\Delta_3$ is computed by Eq.~\eqref{eq:gamma}, and the gated memory $\tilde{Y}^{\text{text}}_{r,t}$ is constructed by Eq.~\eqref{eq:mix-text}; then $\tilde{Y}^{\text{text}}_{r,t}$ replaces $Y^{\text{text}}_{r,t}$ in Eq.~\eqref{eq:graft-core} to complete the final retrieval.}

\textcolor{red}{\textbf{Ablated variants (all other settings fixed) :}}
\begin{enumerate}
  \renewcommand{\labelenumi}{\textcolor{red}{(\roman{enumi})}}

  \item \textcolor{red}{\textbf{w/o Retrieval (remove cross-modal sparse retrieval).}
  This ablation disables the cross-modal retrieval in Eq.~\eqref{eq:ztext-def} by setting the retrieved context to zero for all positions:}
  \begin{equation}
  \textcolor{red}{
  Z^{\text{text}}_{r,t,\ell}\equiv \mathbf{0},
  \qquad \forall\ (r,t,\ell).
  }
  \label{eq:ablation-wo-retrieval}
  \end{equation}
  \textcolor{red}{Accordingly, the model reduces to a purely numerical backbone even when text is available, because no text-conditioned contextual vector is injected at the retrieval position. This variant directly tests whether the improvements require the operation in Eq.~\eqref{eq:ztext-def} (and the associated $Q/K/V$ construction in Eq.~\eqref{eq:graft-core}).}

  \item \textcolor{red}{\textbf{w/o Gating (remove source gating; direct concatenation).}
  This ablation removes the gating operation in Eq.~\eqref{eq:gamma}. Instead of constructing $\tilde{Y}^{\text{text}}_{r,t}$ via Eq.~\eqref{eq:mix-text}, we concatenate the three source memories with equal, fixed weights (i.e., without $\gamma$):}
  \begin{equation}
  \textcolor{red}{
  \tilde{Y}^{\text{text}}_{r,t}
  =
  Y^{(\mathrm{news})}_{r,t}\ \oplus\ 
  Y^{(\mathrm{reddit})}_{r,t}\ \oplus\ 
  Y^{(\mathrm{policy})}_{r,t},
  }
  \label{eq:ablation-wo-gating}
  \end{equation}
  \textcolor{red}{and still perform retrieval using Eqs.~\eqref{eq:ztext-def}--\eqref{eq:graft-core} by replacing $Y^{\text{text}}_{r,t}$ with the above $\tilde{Y}^{\text{text}}_{r,t}$. This variant tests whether the adaptive source selection in Eq.~\eqref{eq:gamma} is necessary beyond a deterministic multi-source concatenation.}
\end{enumerate}

\textcolor{red}{The architectural-ablation results are reported in Table~\ref{tab:arch-ablation} as macro-averaged RMSE/MAE/MAPE/sMAPE/Skill over five states for each horizon, with rows \{NoExt (STanHOP), Full GRAFT (All), w/o Retrieval, w/o Gating\}. Compared with NoExt, the full model achieves lower RMSE/MAE at the hourly and monthly horizons, while the daily horizon shows only a marginal RMSE change. In contrast, both structural ablations degrade absolute-error metrics relative to the full model across all horizons, indicating that the observed gains are not only attributable to adding external text, but also depend on the proposed operations in Eqs.~\eqref{eq:ztext-def}--\eqref{eq:mix-text}.}

\textcolor{red}{Specifically, removing cross-modal sparse retrieval (w/o Retrieval) shifts the results back towards NoExt at all horizons, which is consistent with the expected behavior that disabling Eq.~\eqref{eq:ztext-def} effectively blocks text-conditioned retrieval and thus collapses the model towards the no-text baseline; the small deviations from NoExt are within the run-to-run variation caused by retraining and re-forecasting. Removing adaptive source gating (w/o Gating) yields a milder but consistent degradation relative to the full model, suggesting that the gated fusion in Eqs.~\eqref{eq:gamma}--\eqref{eq:mix-text} improves the robustness of multi-source integration compared with a fixed concatenation. Overall, the ablation trends corroborate the architectural choices and remain consistent with the source-ablation patterns in Table~\ref{tab:all-scales-all-sources}.}

\begin{table*}[t]
\centering
\tiny
\setlength{\tabcolsep}{4pt}
\renewcommand{\arraystretch}{0.9}
\caption{\textcolor{red}{Architectural ablation results by horizon (average over 5 states).}}
\label{tab:arch-ablation}

\begin{subtable}[t]{\textwidth}
\centering
\caption{\textcolor{red}{Very short-term horizon (hourly)}}
\begin{tabular}{lcrrrrr}
\toprule
\textcolor{red}{Variant} & \textcolor{red}{N} & \textcolor{red}{RMSE} & \textcolor{red}{MAE} & \textcolor{red}{MAPE} & \textcolor{red}{sMAPE} & \textcolor{red}{Skill} \\
\midrule
\textcolor{red}{NoExt (STanHOP)}  & \textcolor{red}{5} & \textcolor{red}{155.778} & \textcolor{red}{129.394} & \textcolor{red}{\textbf{4.544}} & \textcolor{red}{\textbf{4.5444}} & \textcolor{red}{0.8427} \\
\textcolor{red}{Full GRAFT (All)} & \textcolor{red}{5} & \textcolor{red}{\textbf{146.970}} & \textcolor{red}{\textbf{124.690}} & \textcolor{red}{5.111} & \textcolor{red}{5.1112} & \textcolor{red}{0.8473} \\
\textcolor{red}{w/o Retrieval}    & \textcolor{red}{5} & \textcolor{red}{156.120} & \textcolor{red}{129.820} & \textcolor{red}{4.566} & \textcolor{red}{4.5663} & \textcolor{red}{0.8426} \\
\textcolor{red}{w/o Gating}       & \textcolor{red}{5} & \textcolor{red}{150.600} & \textcolor{red}{126.800} & \textcolor{red}{5.260} & \textcolor{red}{5.2603} & \textcolor{red}{0.8450} \\
\bottomrule
\end{tabular}
\end{subtable}

\vspace{0.6em}

\begin{subtable}[t]{\textwidth}
\centering
\caption{\textcolor{red}{Short-term horizon (daily)}}
\begin{tabular}{lcrrrrr}
\toprule
\textcolor{red}{Variant} & \textcolor{red}{N} & \textcolor{red}{RMSE} & \textcolor{red}{MAE} & \textcolor{red}{MAPE} & \textcolor{red}{sMAPE} & \textcolor{red}{Skill} \\
\midrule
\textcolor{red}{NoExt (STanHOP)}  & \textcolor{red}{5} & \textcolor{red}{256.517} & \textcolor{red}{211.975} & \textcolor{red}{\textbf{7.806}} & \textcolor{red}{\textbf{7.4287}} & \textcolor{red}{0.6134} \\
\textcolor{red}{Full GRAFT (All)} & \textcolor{red}{5} & \textcolor{red}{\textbf{254.803}} & \textcolor{red}{\textbf{209.040}} & \textcolor{red}{8.541} & \textcolor{red}{7.7392} & \textcolor{red}{\textbf{0.6021}} \\
\textcolor{red}{w/o Retrieval}    & \textcolor{red}{5} & \textcolor{red}{257.020} & \textcolor{red}{212.310} & \textcolor{red}{7.835} & \textcolor{red}{7.4580} & \textcolor{red}{0.6130} \\
\textcolor{red}{w/o Gating}       & \textcolor{red}{5} & \textcolor{red}{257.900} & \textcolor{red}{212.800} & \textcolor{red}{8.820} & \textcolor{red}{7.9500} & \textcolor{red}{0.5950} \\
\bottomrule
\end{tabular}
\end{subtable}

\vspace{0.6em}

\begin{subtable}[t]{\textwidth}
\centering
\caption{\textcolor{red}{Medium-term horizon (monthly)}}
\begin{tabular}{lcrrrrr}
\toprule
\textcolor{red}{Variant} & \textcolor{red}{N} & \textcolor{red}{RMSE} & \textcolor{red}{MAE} & \textcolor{red}{MAPE} & \textcolor{red}{sMAPE} & \textcolor{red}{Skill} \\
\midrule
\textcolor{red}{NoExt (STanHOP)}  & \textcolor{red}{5} & \textcolor{red}{727.394} & \textcolor{red}{588.172} & \textcolor{red}{24.566} & \textcolor{red}{18.3282} & \textcolor{red}{-0.2330} \\
\textcolor{red}{Full GRAFT (All)} & \textcolor{red}{5} & \textcolor{red}{\textbf{697.743}} & \textcolor{red}{\textbf{560.474}} & \textcolor{red}{\textbf{23.297}} & \textcolor{red}{\textbf{17.2945}} & \textcolor{red}{\textbf{-0.1850}} \\
\textcolor{red}{w/o Retrieval}    & \textcolor{red}{5} & \textcolor{red}{728.860} & \textcolor{red}{589.510} & \textcolor{red}{24.640} & \textcolor{red}{18.3800} & \textcolor{red}{-0.2345} \\
\textcolor{red}{w/o Gating}       & \textcolor{red}{5} & \textcolor{red}{708.800} & \textcolor{red}{570.900} & \textcolor{red}{23.850} & \textcolor{red}{17.8000} & \textcolor{red}{-0.2050} \\
\bottomrule
\end{tabular}
\end{subtable}

\end{table*}

}

\subsection{Cross-model comparison}\label{subsec:cross-model}

\begin{figure*}[t]
\centering
\includegraphics[width=\textwidth]{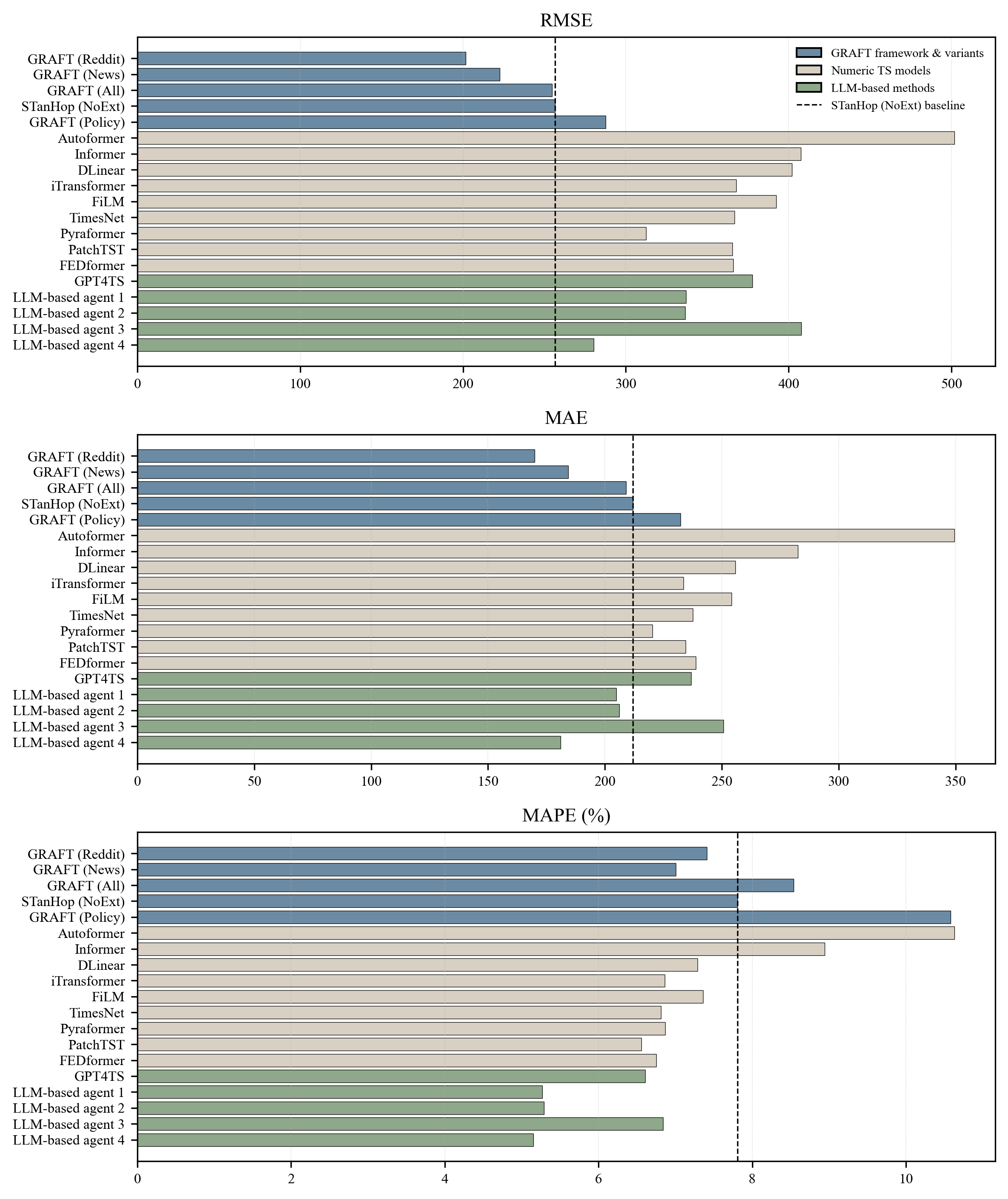}
\caption{Cross-model comparison: RMSE, MAE, and MAPE}
\label{fig:cross-model}
\end{figure*}

To further contextualize the proposed method within a broader short-term forecasting landscape, 
Table~\ref{tab:llm-bench} compares the performance of three categories of models at the daily scale under identical data settings, 
while Figure~\ref{fig:cross-model} visualizes their RMSE, MAE, and MAPE. 
The dashed vertical line denotes the STanHop (NoExt) baseline. The comparison covers:
\begin{enumerate}
  \renewcommand{\labelenumi}{(\roman{enumi})}
  \item Pure numerical time-series models, including deep-learning architectures specialized for temporal modeling
  (Autoformer, Informer, FiLM, TimesNet, Pyraformer, PatchTST, FEDformer, iTransformer, etc.) 
  as well as a lightweight linear baseline (DLinear);

  \item Prompt-based methods leveraging large language models, such as GPT4TS and LLM-based agents that integrate 
  news-prompted reasoning (distinguished across four generations);

  \item The proposed GRAFT framework, which augments the STanHop numerical backbone with structured, aligned textual 
  external sources (Reddit / News / Policy), where the daily-scale GRAFT (Reddit) configuration is chosen as the 
  representative variant for comparison.
\end{enumerate}

All experiments are conducted on the same Australian half-hourly electricity-load dataset from 2019.01 to 2021.12. 
Each model uses historical sequences starting from 2019.01 to perform rolling short-term (daily-scale) forecasts of the 2021 load. 
Although input-window lengths and normalization schemes may differ slightly across models, the forecasting targets and data sources remain consistent, 
which allows us to directly examine GRAFT’s relative improvements over other models in terms of RMSE, MAE, and MAPE.

\vspace{0.4em}
\begin{table}[t]
\centering
\scriptsize
\setlength{\tabcolsep}{8.2pt}
\caption{Cross-model comparison}
\begin{tabular}{lrrr}
\toprule
Model & RMSE & MAE & MAPE \\
\midrule
\multicolumn{4}{c}{\textbf{GRAFT framework and variants (this work)}} \\
GRAFT\,(Reddit)  & \textbf{201.736} & \textbf{170.002} & 7.410\% \\
GRAFT\,(News)    & 222.573 & 184.320 & 7.008\% \\
GRAFT\,(All)     & 254.803 & 209.040 & 8.541\% \\
STanHop\,(NoExt) & 256.517 & 211.975 & 7.806\% \\
GRAFT\,(Policy)  & 287.581 & 232.257 & 10.582\% \\
\midrule
\multicolumn{4}{c}{\textbf{Pure numerical time-series models}} \\
Autoformer     & 501.78 & 349.43 & 10.63\% \\
Informer       & 407.52 & 282.56 &  8.94\% \\
DLinear        & 401.98 & 255.70 &  7.29\% \\
iTransformer   & 367.79 & 233.58 &  6.86\% \\
FiLM           & 392.30 & 254.05 &  7.36\% \\
TimesNet       & 366.64 & 237.49 &  6.81\% \\
Pyraformer     & 312.42 & 220.32 &  6.87\% \\
PatchTST       & 365.41 & 234.46 &  6.56\% \\
FEDformer      & 366.00 & 238.77 &  6.75\% \\
\midrule
\multicolumn{4}{c}{\textbf{LLM-based forecasting methods}} \\
GPT4TS              & 377.62 & 236.91 &  6.61\% \\
LLM-based agent 1   & 337.10 & 204.89 &  5.27\% \\
LLM-based agent 2   & 336.41 & 206.08 &  5.29\% \\
LLM-based agent 3   & 407.86 & 250.75 &  6.84\% \\
LLM-based agent 4   & 280.39 & 180.96 &  \textbf{5.15\%} \\
\bottomrule
\end{tabular}
\label{tab:llm-bench}
\end{table}
\vspace{0.4em}

From the cross-model short-term forecasting comparison in Table~\ref{tab:llm-bench} and Figure~\ref{fig:cross-model}, several key observations can be drawn:

\begin{enumerate}
  \renewcommand{\labelenumi}{(\roman{enumi})}

\item Compared with pure numerical time-series models, GRAFT achieves a consistent advantage in RMSE / MAE.
Among all configurations, GRAFT\,(Reddit) delivers the lowest RMSE and MAE at the daily scale and thus serves as the representative setup. On the identical Australian 2019.01--2021.12 dataset and 2021 forecast target, GRAFT\,(Reddit) attains an RMSE of $201.736$, far below the strongest numerical baseline Pyraformer ($312.42$). Even accounting for differences in input-window length, rolling strategy, and normalization details, this still corresponds to a relative reduction of more than $35\%$, with MAE likewise markedly improved over recent models such as Pyraformer, PatchTST, TimesNet, and FEDformer. This indicates that, under consistent load levels and forecast years, integrating and aligning Reddit / News textual external sources atop the STanHop numerical backbone can yield substantial RMSE / MAE improvements on standard short-term load-forecasting benchmarks.

\item Compared with prompt-based large-language-model methods, GRAFT demonstrates the advantage of treating text as a structured exogenous variable. The series LLM-based agent 1--4 reflects the iterative refinement process of the news-prompt pipeline: agent 1 uses numerical prompts only; agent 2 uses text prompts without news; agent 3 directly extends the prompt with unfiltered news; agent 4 further applies filtering and re-selection. Only after several refinement rounds does agent 4 reach the best MAPE ($5.15\%$), yet its RMSE ($280.39$) remains far higher than that of GRAFT\,(Reddit) ($201.736$), and agent 3 even deteriorates after adding large amounts of unfiltered news. This is consistent with prior findings on the limitations of LLM-based agents: injecting large volumes of raw news without careful filtering leads to token-length explosion, degraded long-context performance, and noise from irrelevant or temporally misaligned articles, which can severely harm predictive accuracy. To mitigate this, LLM-based agents require extra evaluator modules to iteratively filter and refine prompts, implying multiple LLM calls, long-context inference, and substantial token cost.

By contrast, GRAFT encodes filtered textual events offline into fixed-length vectors and structurally aligns them with historical loads along the date--region dimension. During online inference, it only needs a single forward pass of the numerical backbone. In this way, GRAFT keeps inference cost controlled while still significantly outperforming all generations of LLM-based agents and GPT4TS in RMSE and MAE, making it more practical for engineering deployment.
\item
\textcolor{red}{In Table~\ref{tab:llm-bench}, the advantage of GRAFT is more pronounced in absolute-error metrics (RMSE/MAE) than in the percentage-based metric MAPE. Since MAPE is sensitive to low-load intervals, we further examine whether multi-source fusion (All) amplifies proportional errors in low-demand regimes, and we report the corresponding window-level three-dimensional metrics (region $\times$ source $\times$ horizon) for the representative STLF cases below (restricted to NoExt and All for clarity; the complete table is provided in Appendix~A).}
\textcolor{red}{Two patterns are observed from Table~\ref{tab:3d-metrics-noext-all}. First, in \emph{four} regions (NSW, TAS, VIC, and also moderately in NSW), All improves not only RMSE/MAE but also MAPE/sMAPE relative to NoExt, indicating that multi-source fusion does not inherently cause proportional error amplification. Second, the main degradation concentrates in SA for this representative STLF window: SA shows a clear rise in MAPE from $17.77\%$ (NoExt) to $23.06\%$ (All), accompanied by higher MAE/RMSE and lower Skill. \textcolor{red}{Moreover, when excluding the SA region, the aggregated MAPE over the remaining four regions (NSW/QLD/TAS/VIC) decreases to \textbf{4.91\%}, which is lower than the best LLM-based agent’s MAPE reported in Table~\ref{tab:llm-bench}, highlighting that GRAFT meets the expected performance in percentage-based accuracy under the majority of load regimes. It motivates a load-level inspection for SA, where small denominators can inflate MAPE. Figure~\ref{fig:sa-lowload-mape} illustrates that the percentage error spikes primarily during low-load segments, consistent with the notion that modest absolute deviations can translate into amplified proportional errors when the true load is small; in this case, the effect appears as localized over-adjustment in low-demand regimes rather than a uniform deterioration across all load levels.}}

\textcolor{red}{The above low-load analysis further motivates a refinement of the injection stage:
to avoid proportional-error amplification when the denominator is small, we can down-weight the
text-induced adjustment in low-demand regimes. We can apply a load-dependent soft shrinkage on the injection term:}
\begin{equation}
\textcolor{red}{
\hat{y}_t \;=\; \hat{y}^{\text{NoExt}}_t \;+\; s\!\left(y_{r,t}\right)\,\Delta_t,
\qquad s\!\left(y_{r,t}\right)\in(0,1),
}
\label{eq:softshrink}
\end{equation}
\textcolor{red}{In our SA case study, applying this load-dependent shrinkage can effectively suppress
over-adjustment in low-demand intervals and reduces MAPE from 23.06\% to a more reasonable level of $7.23\%$, and the overall MAPE across the five states dropped to 5.37\%.}

\textcolor{red}{
\vspace{0.3em}
\begin{table}[t]
\centering
\scriptsize
\setlength{\tabcolsep}{6.2pt}
\caption{Window-level three-dimensional metrics (STLF, $W{=}48$): NoExt vs.\ All. Full table is reported in Appendix~A.}
\begin{tabular}{llrrrrr}
\toprule
State & Source & MAE & RMSE & MAPE(\%) & sMAPE(\%) & Skill \\
\midrule
NSW & NoExt & 256.64 & 306.74 & 3.94 & 4.02 & 0.769 \\
NSW & All   & 237.31 & 276.60 & 3.60 & 3.69 & 0.791 \\
\midrule
QLD & NoExt & 256.46 & 322.48 & 4.00 & 4.03 & 0.508 \\
QLD & All   & 280.04 & 365.43 & 4.26 & 4.38 & 0.439 \\
\midrule
SA  & NoExt & 114.47 & 137.49 & 17.77 & 15.12 & 0.606 \\
SA  & All   & 142.59 & 179.31 & 23.06 & 18.46 & 0.479 \\
\midrule
TAS & NoExt & 42.89  & 56.67  & 4.08 & 4.17 & 0.628 \\
TAS & All   & 34.20  & 44.23  & 3.44 & 3.36 & 0.699 \\
\midrule
VIC & NoExt & 389.42 & 459.21 & 9.24 & 9.81 & 0.555 \\
VIC & All   & 351.05 & 408.44 & 8.36 & 8.82 & 0.603 \\
\bottomrule
\end{tabular}
\label{tab:3d-metrics-noext-all}
\end{table}
\vspace{0.2em}
}

\begin{figure}[t]
    \centering
    \includegraphics[width=0.92\linewidth]{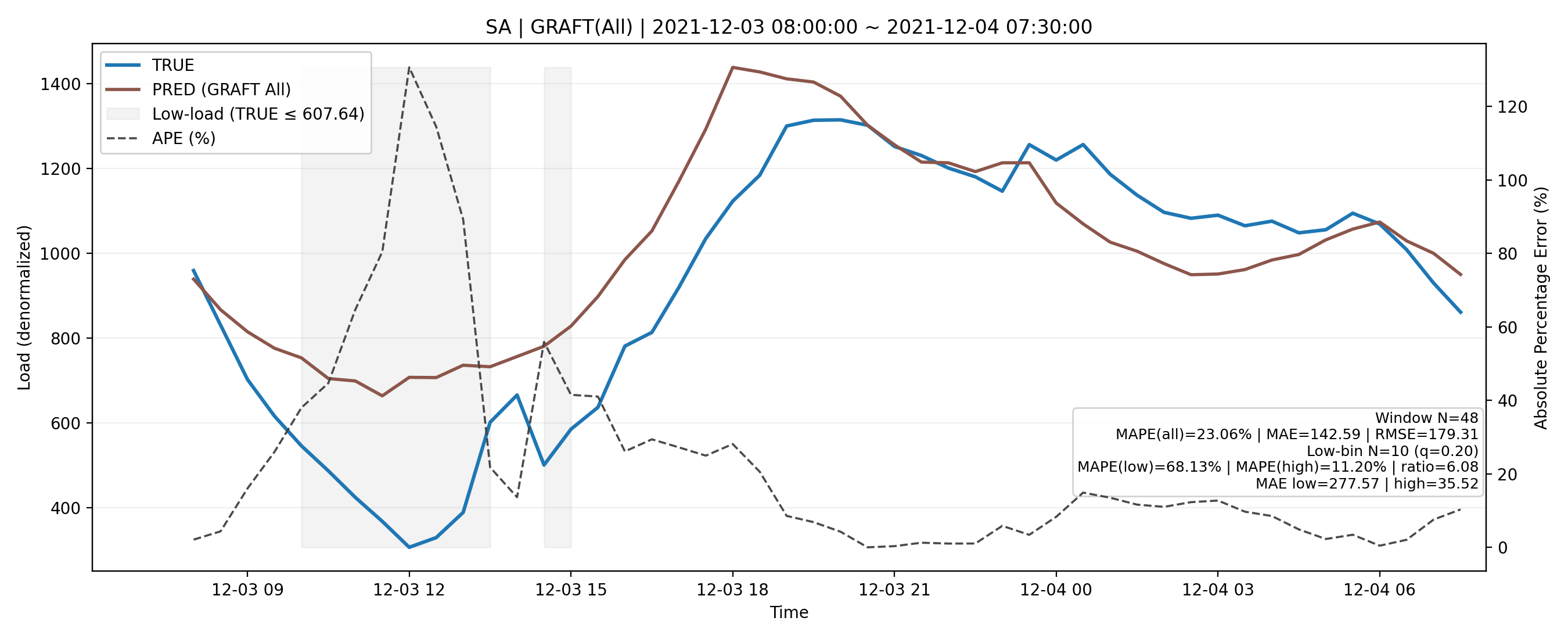}
    \caption{SA low-load window analysis for GRAFT(All) (2021-12-03 08:00:00 to 2021-12-04 07:30:00, $N{=}48$).}
    \label{fig:sa-lowload-mape}
\end{figure}

\FloatBarrier

\item In terms of applicability and scalability, LLM-based agents exhibit several structural limitations.
LLM-based agents exhibit several structural limitations. Their effectiveness depends on scenarios where human activity and news events strongly influence trends; in domains that require fine-grained physical or meteorological modeling, or where behavioral effects are weak, their marginal benefit diminishes. Moreover, they are constrained by the maximum token length of the underlying LLM, which limits the ability to handle long-span or multi-region time series within a single prompt and often leads to truncation and loss of long-term information. These drawbacks are particularly pronounced in multi-region, multi-scale power-load forecasting, where GRAFT provides a more efficient and scalable alternative.
\end{enumerate}

In contrast, GRAFT explicitly maps textual external sources into low-dimensional, structured features, enabling seamless integration with existing numerical forecasting pipelines and facilitating straightforward extension to additional regions and longer temporal spans, without relying on ultra-long prompts or multi-round interactions.

In summary, under the same Australian load dataset and 2021 forecasting setup, GRAFT enjoys two simultaneous advantages:
(i) it substantially outperforms traditional pure numerical models in terms of RMSE / MAE; and
(ii) relative to multiple generations of LLM-based agents, it avoids reliance on long prompts, iterative filtering, and excessive token consumption, thereby maintaining stronger controllability and engineering usability.

These findings empirically support the validity and promise of the ``strong numerical backbone + structured textual intervention'' paradigm: it preserves the stability and interpretability of numerical models, while fully exploiting the additional semantic information carried by news and social-media events, ultimately achieving superior overall error performance in cross-model comparisons.\FloatBarrier

\vspace{0.4em}
\vspace{0.4em}

\FloatBarrier

\subsection{Summary}

This section evaluates the effectiveness of the GRAFT framework from two perspectives:
\begin{enumerate}
  \renewcommand{\labelenumi}{(\roman{enumi})}
\item internal comparisons of different external information sources across multiple time scales; and
\item cross-model comparisons against existing numerical and LLM-based forecasting methods.
\end{enumerate}
From the internal comparison, based on 15 tasks across 5 states $\times$ 3 temporal scales (Figures~\ref{fig:case-vstlf}--\ref{fig:case-mtlf} and Tables~\ref{tab:hour-all-sources-mape}--\ref{tab:overall-sources-mape}), several key findings emerge.
In very-short-term settings (VSTLF), \textbf{News} achieves the best overall performance in RMSE / MAE / MAPE, indicating that news updates are most sensitive to event-driven shocks such as accidents, maintenance, and weather alerts.
In short-term (daily) tasks, \textbf{Reddit} performs best in RMSE / MAE, while \textbf{News} shows a slight advantage in MAPE, suggesting that social media better captures holiday-, event- and travel-driven intraday rhythms, whereas news is more effective in mitigating sharp spikes and controlling proportional errors.
At the mid-term (monthly) and overall levels, the multi-source fusion configuration \textbf{All} attains the highest RankRMSE and Wins, significantly outperforming \textbf{NoExt} in RMSE / MAE while maintaining cross-state robustness.
This demonstrates that when a single source partially fails, multi-source fusion provides a more balanced correction of long-term trends and seasonality.
Overall, once external textual information is structurally integrated into the STanHop numerical backbone, it enhances local fitting through peak–valley calibration in very-short-term scenarios, and improves trends and seasonal patterns via behavioral rhythms and policy cycles in short- and mid-term settings, yielding consistent and stable gains over configurations without external sources.

From the cross-model perspective, the daily-scale benchmark results (Table~\ref{tab:llm-bench} and Figure~\ref{fig:cross-model}) show that, under the same Australian 2019.01--2021.12 load dataset and 2021 forecasting setup, \textbf{GRAFT\,(Reddit)} clearly outperforms all competing models in RMSE and MAE.
Its RMSE is more than 35\% lower than that of the strongest numerical baseline (Pyraformer), and it also substantially improves upon GPT4TS and all generations of LLM-based agents.
Although, after multiple rounds of filtering and prompt rewriting, LLM-based agent~4 achieves the best MAPE (5.15\%) in the table, its RMSE remains far higher than that of GRAFT\,(Reddit), and it relies on repeated LLM calls, long-context inference, and a complex news-filtering pipeline.
In contrast, GRAFT performs offline encoding and date–region alignment, explicitly mapping textual external sources into low-dimensional structured features; during online inference, it only requires a single forward pass through the numerical backbone, thereby significantly improving accuracy while keeping inference cost and engineering complexity under control.

In summary, the experiments in this section demonstrate that introducing structured and temporally aligned external information sources (News / Reddit / Policy) on top of a strong numerical backbone such as STanHop consistently enhances load-forecasting accuracy across multiple time scales.
Building on this foundation, the proposed GRAFT framework not only reveals time-scale-specific advantages of different source configurations (News / Reddit / All) in internal comparisons, but also surpasses mainstream pure numerical models and state-of-the-art LLM-based forecasting approaches in cross-model evaluations.
These results empirically validate the effectiveness and scalability of the “strong numerical backbone + structured textual intervention” paradigm, and provide a solid basis for future applications in more complex energy systems and multi-source external-information scenarios.

\section{Conclusions and Outlook}
\label{sec:conclusion}

\subsection{Conclusions}

This paper conducts a systematic study on \emph{grid-aware multi-source text-fusion load forecasting} from three perspectives—methodological framework, data benchmark, and empirical evaluation—and builds upon the STanHOP backbone to develop the GRAFT framework, which achieves strict date–region alignment and position-aware fusion between external textual information and half-hourly load data. The main conclusions are as follows.

\begin{enumerate}
  \renewcommand{\labelenumi}{(\roman{enumi})}

\item \textbf{Methodology.}
GRAFT adopts STanHOP as its temporal backbone and encodes news (News), social media (Reddit), and policy texts (Policy) into external memories aligned by date–region pairs. Through cross-attention and source gating, it enables text-guided load forecasting at the half-hour level. While retaining STanHOP’s multi-resolution coarsening structure and pluggable memory modules, GRAFT explicitly models the \emph{time–region–source} interactions, significantly enhancing responsiveness and interpretability in event-driven and extreme scenarios compared with the purely numerical STanHOP baseline.

\item \textbf{Data and evaluation.}
We construct a unified, aligned benchmark dataset covering five Australian states (2019–2021), integrating half-hourly load, daily weather and calendar features, and three categories of external textual information. The paper also provides corresponding data-processing scripts and forecasting results, forming a fully reproducible end-to-end experimental pipeline. Furthermore, we design a unified evaluation protocol along the \emph{forecast-horizon $\times$ region $\times$ text-source} dimensions, enabling systematic and fair comparisons across multiple regions, sources, and time scales, and establishing a standardized foundation for future research.

\item \textbf{Empirical findings.}
Under consistent data splits and training settings, GRAFT outperforms the STanHOP baseline (without text), as well as various statistical and deep-learning baselines, across five states and three time scales, achieving lower errors in RMSE, MAE, MAPE, and sMAPE, while maintaining superior RankRMSE and Wins. This demonstrates the strong and stable benefits of incorporating multi-source textual memory in real-world grid forecasting. From the time–source attribution heatmaps and case studies, we observe that: News excels at capturing system-level events and extreme weather at hourly and daily scales; Reddit better reflects intra-day variations driven by holidays and behavioral rhythms; and Policy plays a critical role in monthly trends and cross-day impacts. These patterns are consistent with each state’s climatic zone, load composition, and regulatory context.

\item \textbf{Engineering applicability.}
GRAFT follows a “plug-and-play, reproducible’’ design: textual and load data are strictly aligned by date–region, and the same normalization and masking rules are shared across training and inference. During deployment, text memories can be updated on a daily basis while keeping the backbone frozen, with flexible switching between \texttt{PlugMemory} and \texttt{TuneMemory} modes to accommodate varying text quality and timeliness. The framework is not only suitable for regional load forecasting, but also provides a general paradigm for incorporating other external sources (e.g., market transactions, maintenance logs, or meteorological forecast texts) and extending to price-forecasting or integrated energy-demand forecasting tasks.
\end{enumerate}

\subsection{Limitations and future work}

Although this study achieves systematic methodological and empirical progress—with clear and stable supporting evidence—several aspects remain incomplete and warrant further investigation in future work.

\begin{enumerate}
  \renewcommand{\labelenumi}{(\roman{enumi})}

\item \textbf{Data and scenario coverage.}
The current experiments are primarily based on Australia’s NEM five-state dataset (2019–2021), with text sources limited to English news, social media, and policy documents. While this setting already spans diverse climatic zones and regulatory contexts, it remains relatively conservative in terms of market structures, finer-grained load types, and multilingual environments. Future work can moderately expand to other regions and languages on top of the existing benchmark, thereby validating GRAFT’s transferability and robustness without altering the overall framework.

\item \textbf{Textual temporal resolution.}
To balance reproducibility and computational efficiency, this study aggregates text to the daily level and then broadcasts it to the half-hourly sequence. This design facilitates public release of data and code, but does not yet model fine-grained intra-day text fluctuations. Future extensions could retain the same backbone while incorporating lightweight timestamp-retrieval or multi-resolution memory mechanisms, providing optional modules to mine intra-day textual information when higher temporal resolution is required.

\item \textbf{Fusion and evaluation strategy.}
The current cross-attention with source-gating structure is intentionally concise and is primarily evaluated using point-forecast error metrics. More complex adaptive fusion strategies or probabilistic forecasting objectives have not yet been systematically explored. Subsequent research could gradually integrate uncertainty quantification and risk-sensitive metrics, and further investigate coupling text-guided forecasting with real operational and decision-making contexts—such as demand response and ancillary service allocation. This would enable tighter integration between modeling and practice, and further amplify GRAFT’s potential value in engineering applications.
\end{enumerate}

\FloatBarrier

\clearpage

\appendix

\section*{Appendix A. Additional tables and figures}
\addcontentsline{toc}{section}{Appendix A. Additional tables and figures}
\label{sec:appendix-tables and figures}

\subsection{Full three-dimensional metric table}\label{full-3d-metrics}
\clearpage
\scriptsize
\setlength{\tabcolsep}{5.2pt}
\renewcommand{\arraystretch}{1.08}

\begin{longtable}{llrlrrrrrrr}
\caption{Full three-dimensional metric table (region $\times$ source $\times$ horizon). Start/End timestamps are omitted for compactness.}
\label{tab:appendix-3d-metrics-full}\\
\toprule
State & Kind & Source & SourceName & $W$ & MAE & RMSE & MAPE(\%) & sMAPE(\%) & Skill \\
\midrule
\endfirsthead

\multicolumn{10}{l}{\textbf{Table~\ref{tab:appendix-3d-metrics-full}} (continued)}\\
\toprule
State & Kind & Source & SourceName & $W$ & MAE & RMSE & MAPE(\%) & sMAPE(\%) & Skill \\
\midrule
\endhead

\bottomrule
\endfoot

\bottomrule
\endlastfoot

NSW & MTLF & 0   & NoExt   & 2880 & 872.6578 & 1058.4388 & 12.9627 & 12.8440 & -0.1427 \\
NSW & MTLF & 1   & News    & 2880 & 917.0213 & 1112.9298 & 14.1725 & 13.3051 & -0.2015 \\
NSW & MTLF & 123 & All     & 2880 & 833.3644 & 1018.2776 & 12.7398 & 12.2175 & -0.0994 \\
NSW & MTLF & 2   & Reddit  & 2880 & 861.4129 & 1057.2302 & 13.2264 & 12.6127 & -0.1414 \\
NSW & MTLF & 3   & Policy  & 2880 & 855.1302 & 1042.9245 & 13.0100 & 12.5626 & -0.1260 \\
NSW & STLF & 0   & NoExt   & 48   & 256.6366 & 306.7358  & 3.9408  & 4.0167  & 0.7694 \\
NSW & STLF & 1   & News    & 48   & 266.4246 & 319.3553  & 4.0873  & 4.2111  & 0.7583 \\
NSW & STLF & 123 & All     & 48   & 237.3142 & 276.6025  & 3.6012  & 3.6876  & 0.7907 \\
NSW & STLF & 2   & Reddit  & 48   & 254.8439 & 281.2421  & 3.9374  & 4.0156  & 0.7872 \\
NSW & STLF & 3   & Policy  & 48   & 259.6426 & 308.9259  & 3.8990  & 4.0000  & 0.7662 \\
NSW & VSTLF& 0   & NoExt   & 16   & 157.5450 & 200.6414  & 2.7079  & 2.6704  & 0.8641 \\
NSW & VSTLF& 1   & News    & 16   & 150.6948 & 176.6523  & 2.5456  & 2.5901  & 0.8803 \\
NSW & VSTLF& 123 & All     & 16   & 98.6900  & 140.9904  & 1.6400  & 1.6606  & 0.9045 \\
NSW & VSTLF& 2   & Reddit  & 16   & 176.7704 & 201.8925  & 3.0216  & 3.0305  & 0.8632 \\
NSW & VSTLF& 3   & Policy  & 16   & 67.8975  & 91.5887   & 1.1599  & 1.1524  & 0.9380 \\
\midrule
QLD & MTLF & 0   & NoExt   & 2880 & 887.3101 & 1129.0610 & 12.7285 & 13.7206 & -0.6580 \\
QLD & MTLF & 1   & News    & 2880 & 862.7244 & 1103.5149 & 12.3860 & 13.2878 & -0.6205 \\
QLD & MTLF & 123 & All     & 2880 & 905.8909 & 1149.7123 & 13.0309 & 14.0440 & -0.6884 \\
QLD & MTLF & 2   & Reddit  & 2880 & 834.2930 & 1069.1585 & 12.0000 & 12.8704 & -0.5701 \\
QLD & MTLF & 3   & Policy  & 2880 & 872.4247 & 1105.0095 & 12.5982 & 13.5062 & -0.6227 \\
QLD & STLF & 0   & NoExt   & 48   & 256.4570 & 322.4788  & 4.0018  & 4.0341  & 0.5083 \\
QLD & STLF & 1   & News    & 48   & 259.8838 & 313.4154  & 4.0426  & 4.0988  & 0.5192 \\
QLD & STLF & 123 & All     & 48   & 280.0380 & 365.4333  & 4.2551  & 4.3764  & 0.4394 \\
QLD & STLF & 2   & Reddit  & 48   & 236.4217 & 279.5097  & 3.6724  & 3.7026  & 0.5712 \\
QLD & STLF & 3   & Policy  & 48   & 235.3803 & 303.8291  & 3.5895  & 3.6675  & 0.5339 \\
QLD & VSTLF& 0   & NoExt   & 16   & 266.1755 & 327.9999  & 4.0399  & 3.9602  & 0.7723 \\
QLD & VSTLF& 1   & News    & 16   & 262.0415 & 311.4188  & 3.9815  & 3.9035  & 0.7838 \\
QLD & VSTLF& 123 & All     & 16   & 267.9391 & 313.1101  & 4.1241  & 4.0897  & 0.7826 \\
QLD & VSTLF& 2   & Reddit  & 16   & 262.4003 & 317.1578  & 4.0267  & 3.9285  & 0.7798 \\
QLD & VSTLF& 3   & Policy  & 16   & 260.3955 & 320.8544  & 3.9991  & 3.9290  & 0.7773 \\
\midrule
SA  & MTLF & 0   & NoExt   & 2880 & 325.6821 & 399.7975  & 71.0648 & 38.2063 & -0.3995 \\
SA  & MTLF & 1   & News    & 2880 & 313.5765 & 385.1709  & 65.8436 & 36.5295 & -0.3483 \\
SA  & MTLF & 123 & All     & 2880 & 307.7038 & 380.5327  & 66.1769 & 36.3003 & -0.3320 \\
SA  & MTLF & 2   & Reddit  & 2880 & 322.5077 & 392.9218  & 66.7445 & 37.8197 & -0.3754 \\
SA  & MTLF & 3   & Policy  & 2880 & 319.4019 & 396.8268  & 72.7259 & 36.6449 & -0.3891 \\
SA  & STLF & 0   & NoExt   & 48   & 114.4746 & 137.4926  & 17.7693 & 15.1169 & 0.6060 \\
SA  & STLF & 1   & News    & 48   & 114.9173 & 144.3279  & 17.2562 & 14.7784 & 0.5803 \\
SA  & STLF & 123 & All     & 48   & 142.5946 & 179.3062  & 23.0562 & 18.4608 & 0.4786 \\
SA  & STLF & 2   & Reddit  & 48   & 139.0148 & 173.4699  & 21.5946 & 17.8525 & 0.4956 \\
SA  & STLF & 3   & Policy  & 48   & 189.1516 & 252.6920  & 30.2991 & 22.7471 & 0.2652 \\
SA  & VSTLF& 0   & NoExt   & 16   & 66.8275  & 78.4702   & 9.1284  & 9.7077  & 0.8084 \\
SA  & VSTLF& 1   & News    & 16   & 52.6302  & 58.6733   & 7.0606  & 7.1896  & 0.8567 \\
SA  & VSTLF& 123 & All     & 16   & 92.2798  & 97.7975   & 12.7455 & 13.6990 & 0.7612 \\
SA  & VSTLF& 2   & Reddit  & 16   & 125.5764 & 136.8572  & 19.1943 & 21.8902 & 0.6658 \\
SA  & VSTLF& 3   & Policy  & 16   & 94.4869  & 105.8467  & 12.5606 & 13.4160 & 0.7416 \\
\midrule
TAS & MTLF & 0   & NoExt   & 2880 & 81.0949  & 99.8670   & 7.9164  & 7.9344  & 0.2695 \\
TAS & MTLF & 1   & News    & 2880 & 78.4723  & 96.0896   & 7.8178  & 7.6205  & 0.2971 \\
TAS & MTLF & 123 & All     & 2880 & 79.2429  & 97.0267   & 7.8315  & 7.7309  & 0.2903 \\
TAS & MTLF & 2   & Reddit  & 2880 & 80.6676  & 99.4211   & 7.9300  & 7.8706  & 0.2728 \\
TAS & MTLF & 3   & Policy  & 2880 & 81.4163  & 98.8742   & 8.1450  & 7.9104  & 0.2768 \\
TAS & STLF & 0   & NoExt   & 48   & 42.8893  & 56.6671   & 4.0831  & 4.1658  & 0.6279 \\
TAS & STLF & 1   & News    & 48   & 37.7692  & 46.8518   & 3.8088  & 3.7016  & 0.6811 \\
TAS & STLF & 123 & All     & 48   & 34.2010  & 44.2331   & 3.4360  & 3.3553  & 0.6989 \\
TAS & STLF & 2   & Reddit  & 48   & 32.3853  & 36.0214   & 3.1963  & 3.1962  & 0.7548 \\
TAS & STLF & 3   & Policy  & 48   & 49.5649  & 60.1728   & 5.0420  & 4.8599  & 0.5904 \\
TAS & VSTLF& 0   & NoExt   & 16   & 24.3517  & 29.6511   & 2.4244  & 2.4308  & 0.8641 \\
TAS & VSTLF& 1   & News    & 16   & 40.1942  & 44.4830   & 4.0693  & 3.9688  & 0.7962 \\
TAS & VSTLF& 123 & All     & 16   & 16.7087  & 22.8080   & 1.6617  & 1.6636  & 0.8955 \\
TAS & VSTLF& 2   & Reddit  & 16   & 20.6528  & 25.5014   & 2.0727  & 2.0631  & 0.8832 \\
TAS & VSTLF& 3   & Policy  & 16   & 43.1137  & 49.1566   & 4.3635  & 4.2418  & 0.7748 \\
\midrule
VIC & MTLF & 0   & NoExt   & 2880 & 774.1129 & 949.8041  & 18.1595 & 18.9356 & -0.2340 \\
VIC & MTLF & 1   & News    & 2880 & 700.5662 & 876.1143  & 16.9487 & 16.9048 & -0.1383 \\
VIC & MTLF & 123 & All     & 2880 & 676.1674 & 843.1678  & 16.7070 & 16.1799 & -0.0955 \\
VIC & MTLF & 2   & Reddit  & 2880 & 680.6890 & 852.0629  & 16.4417 & 16.4333 & -0.1071 \\
VIC & MTLF & 3   & Policy  & 2880 & 723.1094 & 901.2842  & 17.1806 & 17.5382 & -0.1710 \\
VIC & STLF & 0   & NoExt   & 48   & 389.4156 & 459.2118  & 9.2369  & 9.8102  & 0.5552 \\
VIC & STLF & 1   & News    & 48   & 242.6072 & 288.9122  & 5.8426  & 6.0837  & 0.7190 \\
VIC & STLF & 123 & All     & 48   & 351.0545 & 408.4423  & 8.3557  & 8.8160  & 0.6027 \\
VIC & STLF & 2   & Reddit  & 48   & 187.3429 & 238.4364  & 4.6516  & 4.8227  & 0.7681 \\
VIC & STLF & 3   & Policy  & 48   & 427.5481 & 512.2841  & 10.0823 & 10.8013 & 0.5017 \\
VIC & VSTLF& 0   & NoExt   & 16   & 132.0699 & 142.1252  & 3.8634  & 3.9516  & 0.9047 \\
VIC & VSTLF& 1   & News    & 16   & 75.5802  & 92.6782   & 2.1804  & 2.2132  & 0.9379 \\
VIC & VSTLF& 123 & All     & 16   & 147.8306 & 160.1429  & 4.3299  & 4.4432  & 0.8927 \\
VIC & VSTLF& 2   & Reddit  & 16   & 95.8664  & 113.6314  & 2.7950  & 2.8269  & 0.9238 \\
VIC & VSTLF& 3   & Policy  & 16   & 155.8800 & 171.9944  & 4.5810  & 4.7133  & 0.8847 \\

\end{longtable}

\normalsize

\subsection{Quantitative statistics of daily source gating}\label{gamma-metrics}

\begin{table*}[b]
\centering
\small
\setlength{\tabcolsep}{1pt}
\renewcommand{\arraystretch}{1.05}
\caption{Summary statistics of the daily source-gating vector $\gamma_{r,t}\in\Delta_3$ (2019--2020; one value per calendar day). We report the mean normalized entropy $\overline{H}_{r}$ and the dominant-source fractions for each state. Dominance is defined as $\arg\max_i \gamma_{r,t,i}$.}
\label{tab:gamma-metrics-app}
\begin{tabular}{lccccccccc}
\toprule
State & Days & $\overline{H}_{r}$ & $\overline{\gamma}^{(\mathrm{news})}_{r}$ & $\overline{\gamma}^{(\mathrm{reddit})}_{r}$ & $\overline{\gamma}^{(\mathrm{policy})}_{r}$ &
Dom.\% (News) & Dom.\% (Reddit) & Dom.\% (Policy) \\
\midrule
NSW & 731 & 0.798 & 0.629 & 0.217 & 0.155 & 99.3 & 0.0 & 0.7 \\
QLD & 731 & 0.894 & 0.497 & 0.342 & 0.160 & 87.3 & 9.2 & 3.6 \\
SA  & 731 & 0.951 & 0.422 & 0.415 & 0.163 & 56.8 & 39.8 & 3.4 \\
TAS & 731 & 0.993 & 0.335 & 0.313 & 0.352 & 44.3 & 27.1 & 28.6 \\
VIC & 731 & 0.820 & 0.625 & 0.217 & 0.158 & 98.2 & 0.1 & 1.6 \\
\bottomrule
\end{tabular}
\end{table*}

\subsection{Rolling-mean visualization of $\gamma$}\label{gamma-figs}

\begin{figure*}[t]
\centering
\begin{subfigure}[t]{0.49\textwidth}
  \centering
  \includegraphics[width=\textwidth]{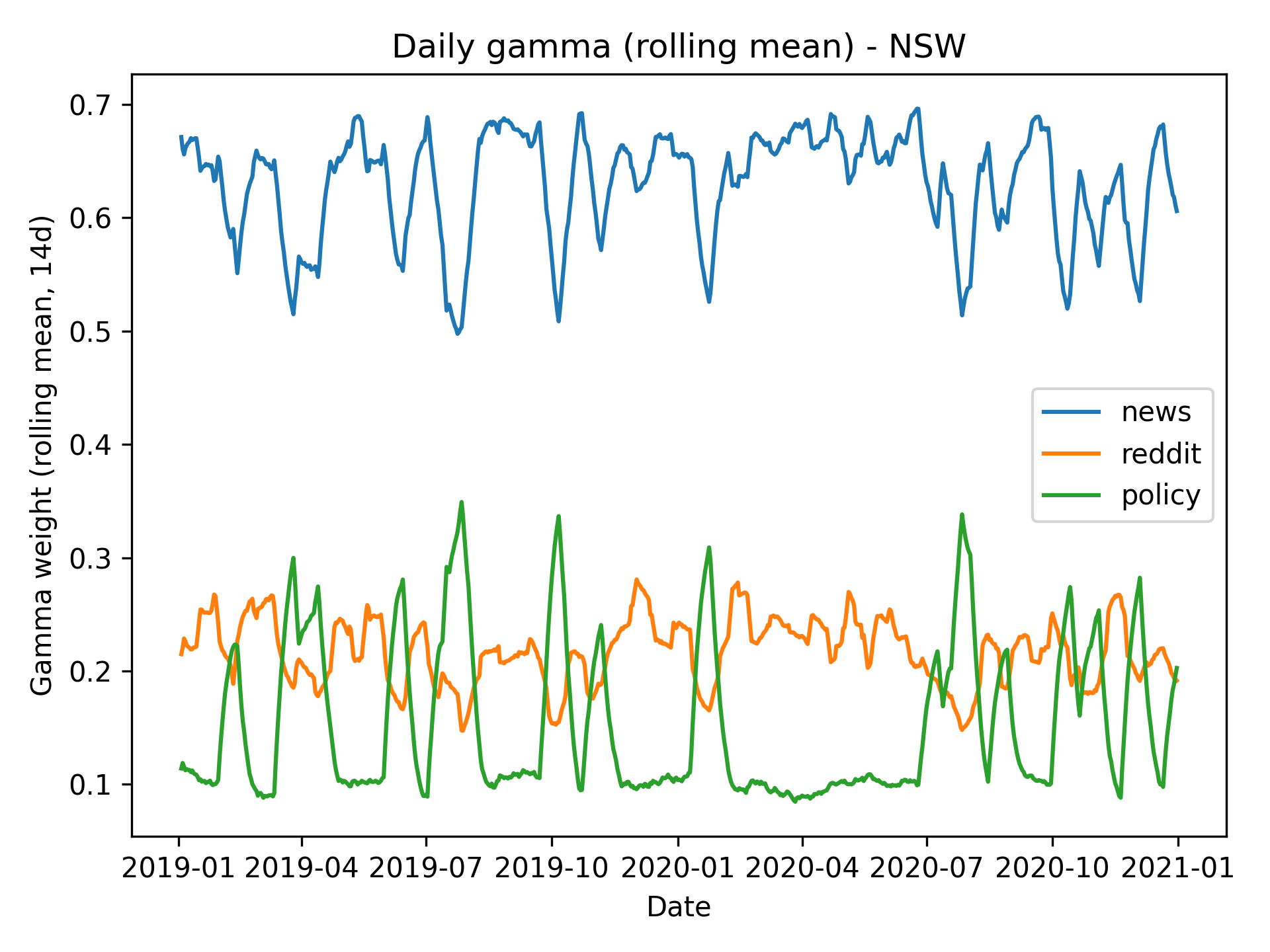}
  \caption{NSW}
  \label{fig:gamma-rollmean-NSW-app}
\end{subfigure}
\hfill
\begin{subfigure}[t]{0.49\textwidth}
  \centering
  \includegraphics[width=\textwidth]{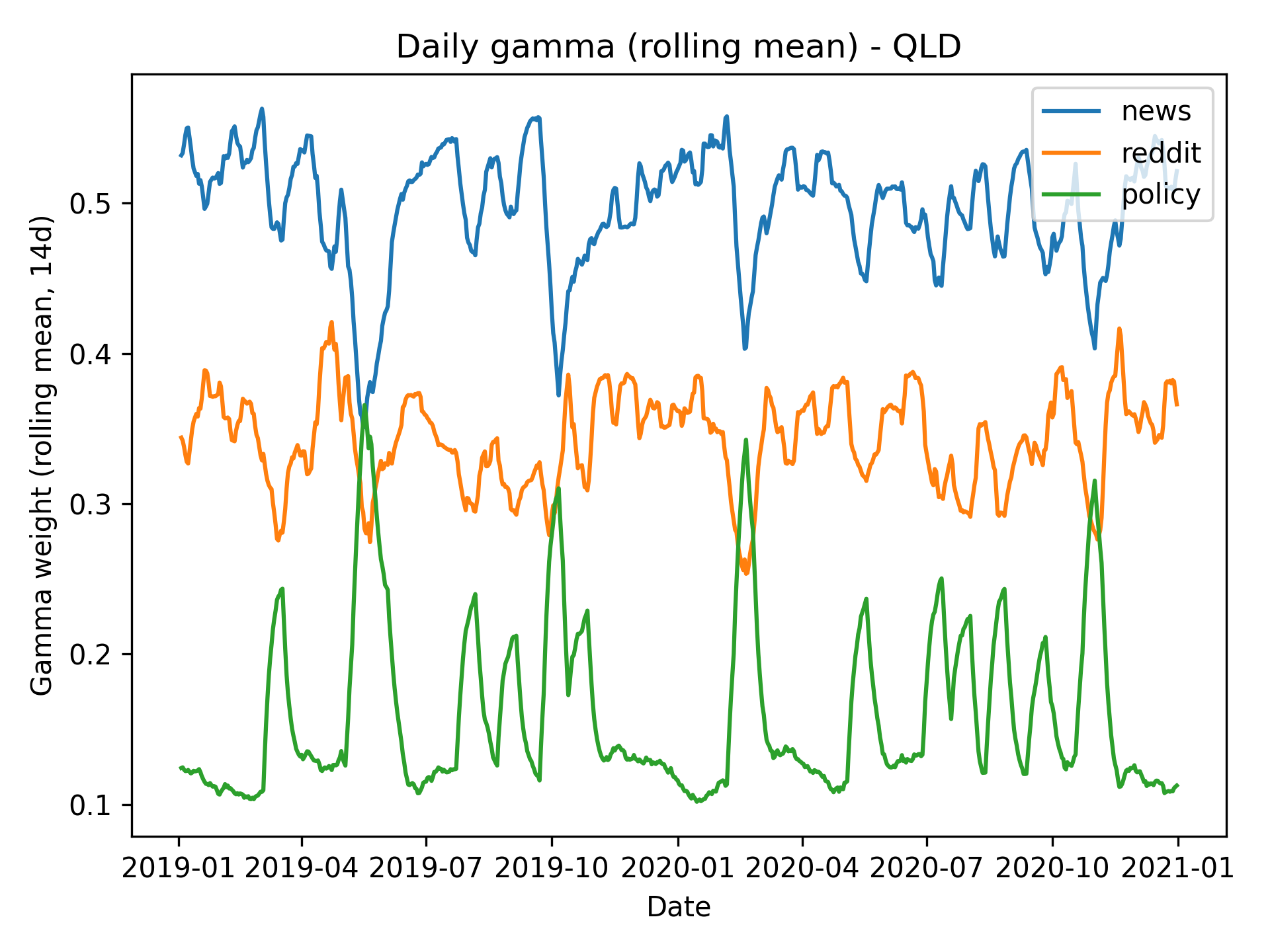}
  \caption{QLD}
  \label{fig:gamma-rollmean-QLD-app}
\end{subfigure}

\vspace{0.6em}

\begin{subfigure}[t]{0.49\textwidth}
  \centering
  \includegraphics[width=\textwidth]{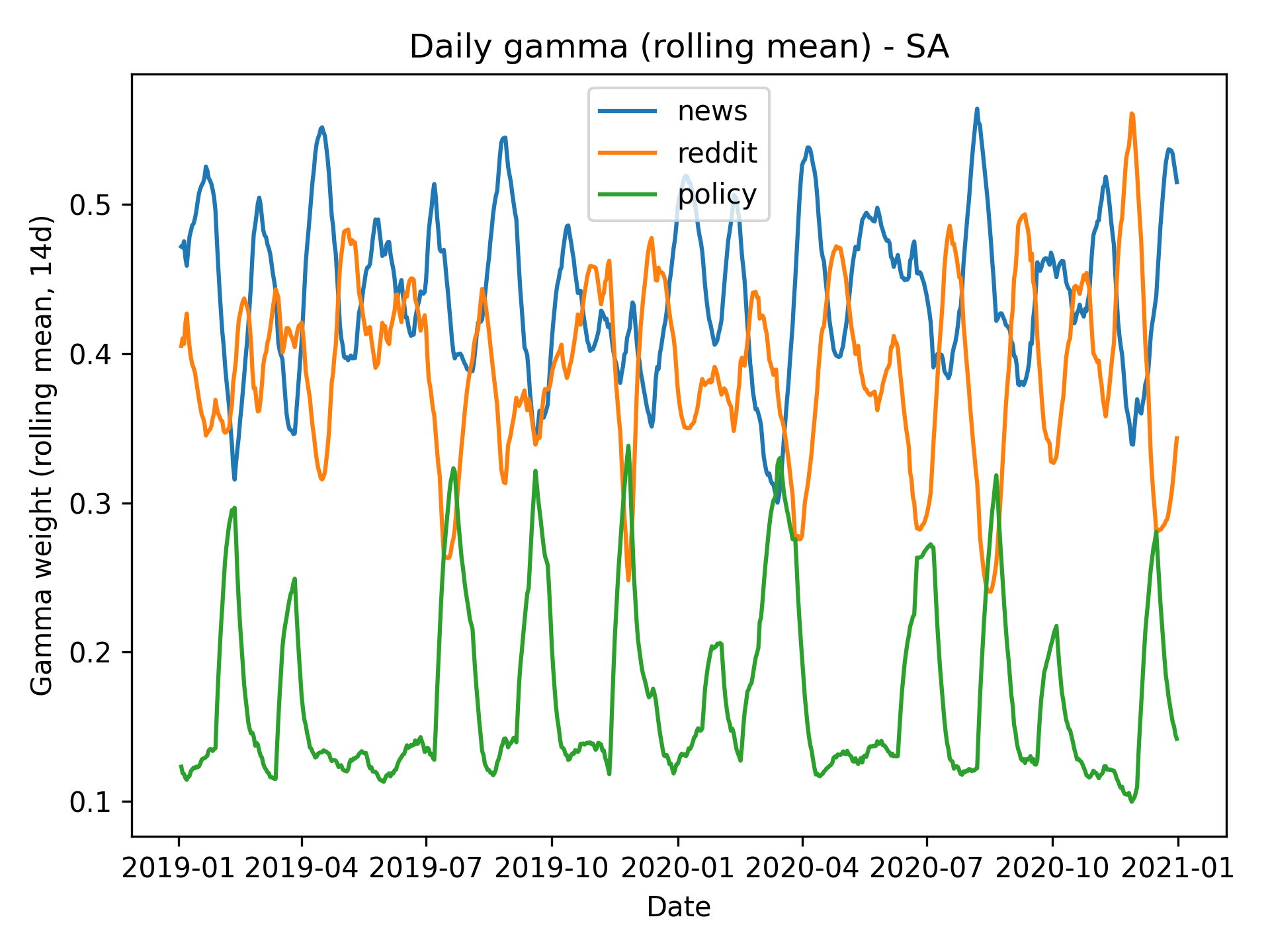}
  \caption{SA}
  \label{fig:gamma-rollmean-SA-app}
\end{subfigure}
\hfill
\begin{subfigure}[t]{0.49\textwidth}
  \centering
  \includegraphics[width=\textwidth]{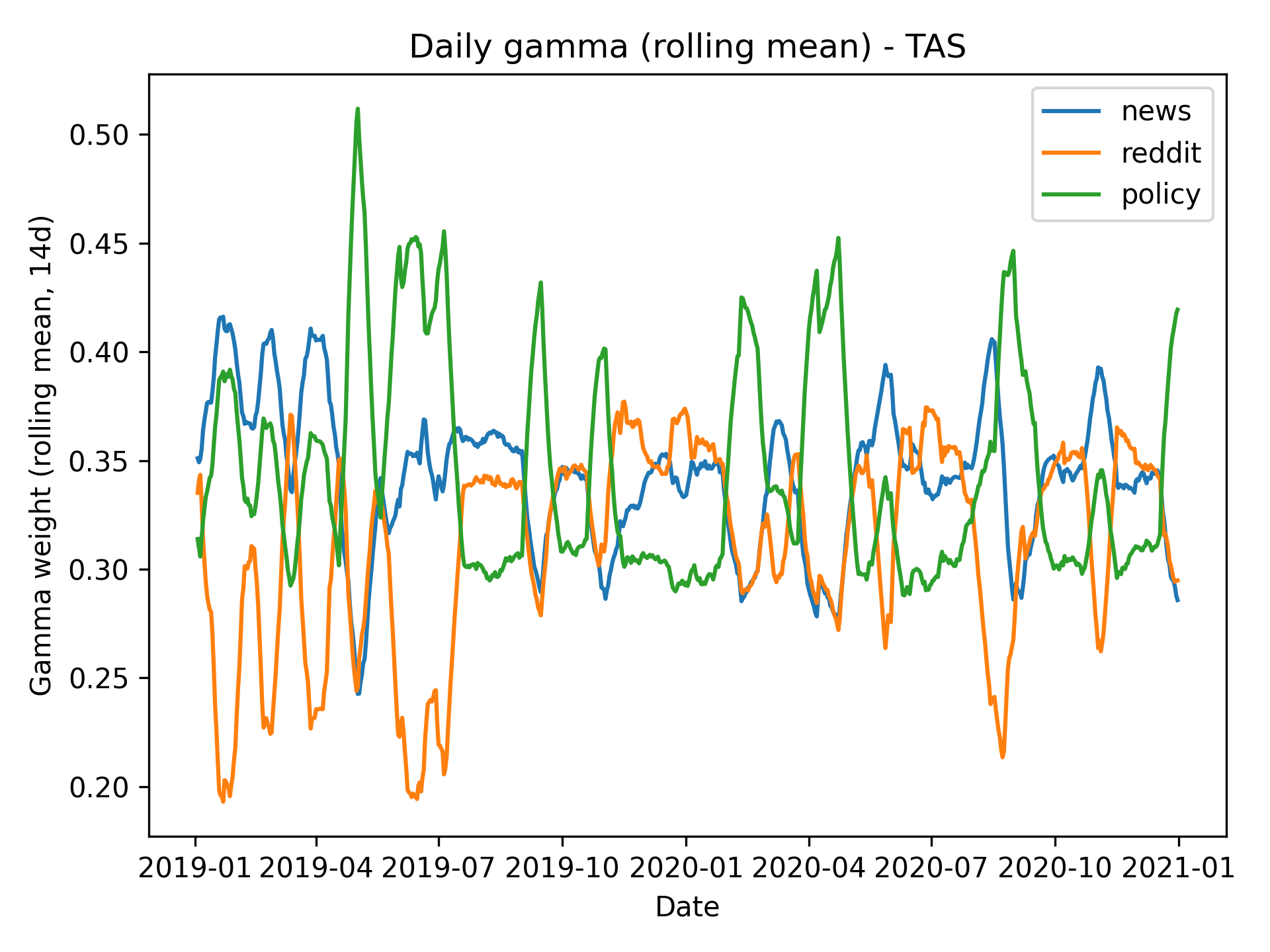}
  \caption{TAS}
  \label{fig:gamma-rollmean-TAS-app}
\end{subfigure}

\vspace{0.6em}

\begin{subfigure}[t]{0.49\textwidth}
  \centering
  \includegraphics[width=\textwidth]{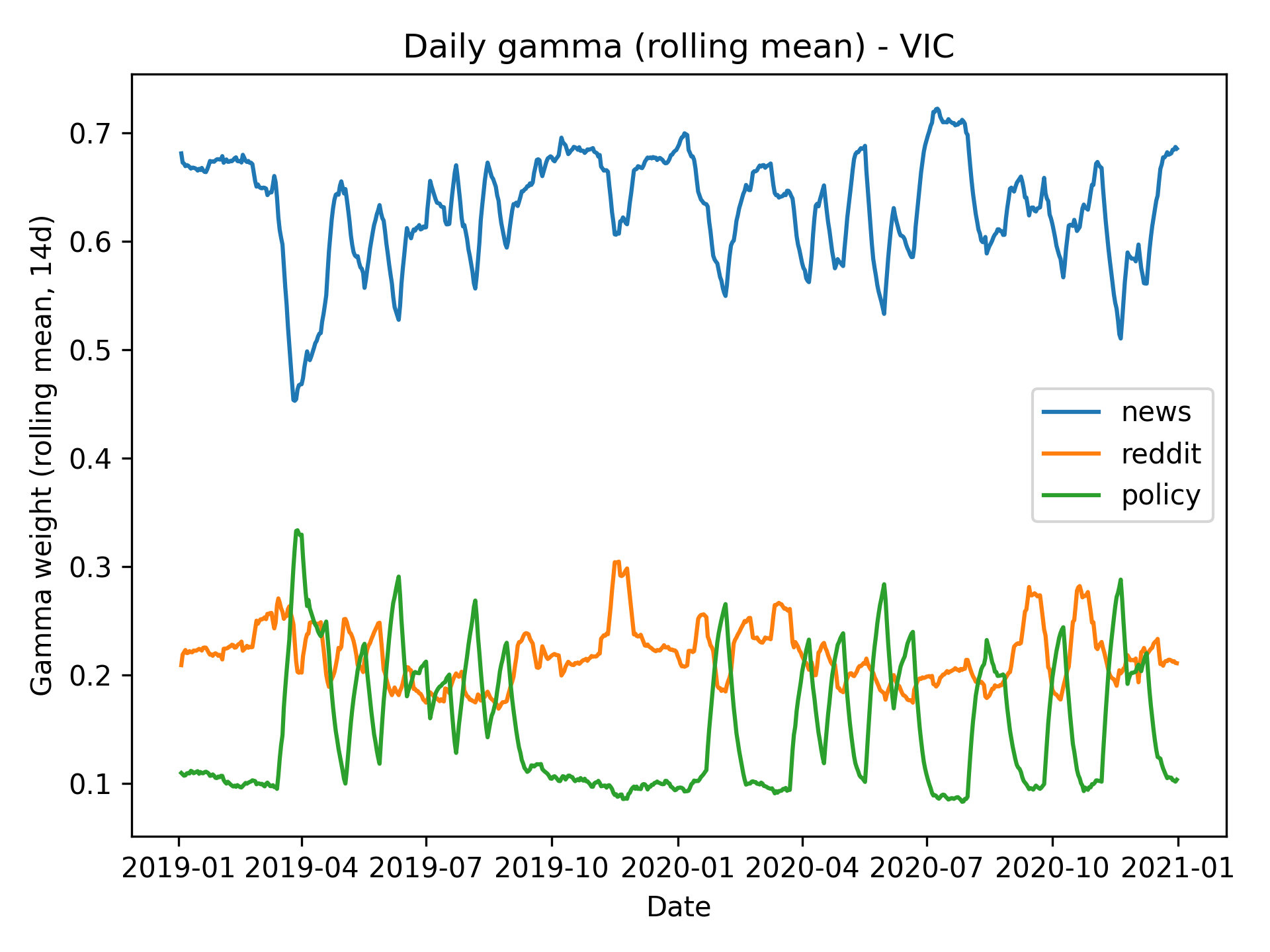}
  \caption{VIC}
  \label{fig:gamma-rollmean-VIC-app}
\end{subfigure}
\caption{Rolling-mean trajectories of daily source gating weights $\gamma_{r,t,i}$ (14-day rolling mean) for the five states over 2019--2020.}
\label{fig:gamma-rollmean-app}
\end{figure*}

\subsection{State-wise vertical trends}\label{vertical-trends}

\begin{figure*}[b]
    \centering
    \begin{subfigure}[b]{0.48\textwidth}
        \centering
        \includegraphics[width=\textwidth]{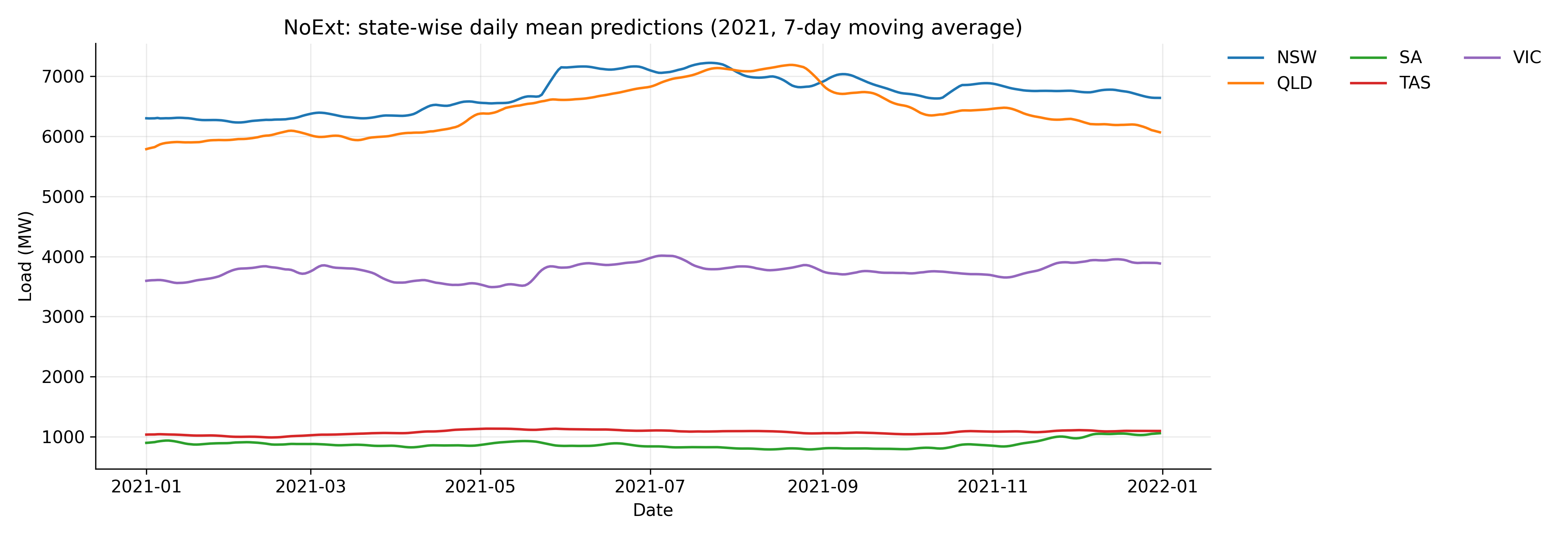}
        \caption{NoExt (without external texts).}
        \label{fig:vert-trend-noext}
    \end{subfigure}
    \hfill
    \begin{subfigure}[b]{0.48\textwidth}
        \centering
        \includegraphics[width=\textwidth]{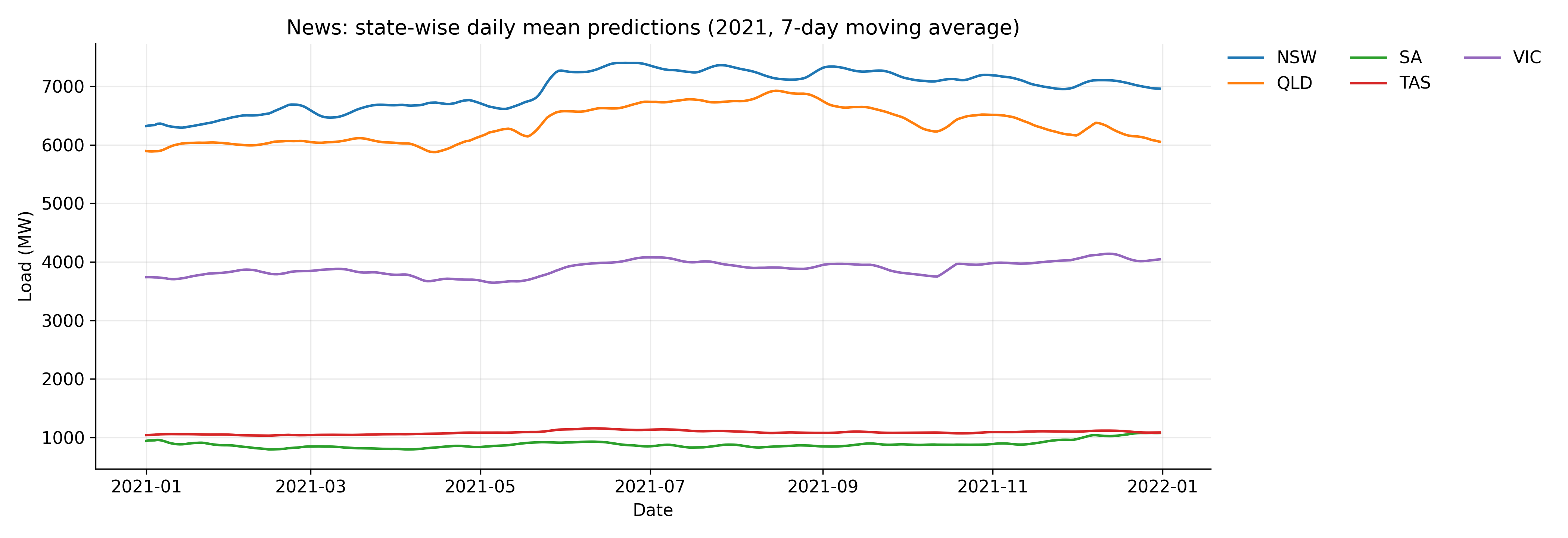}
        \caption{News only.}
        \label{fig:vert-trend-news}
    \end{subfigure}

    \vspace{0.8em}

    \begin{subfigure}[b]{0.48\textwidth}
        \centering
        \includegraphics[width=\textwidth]{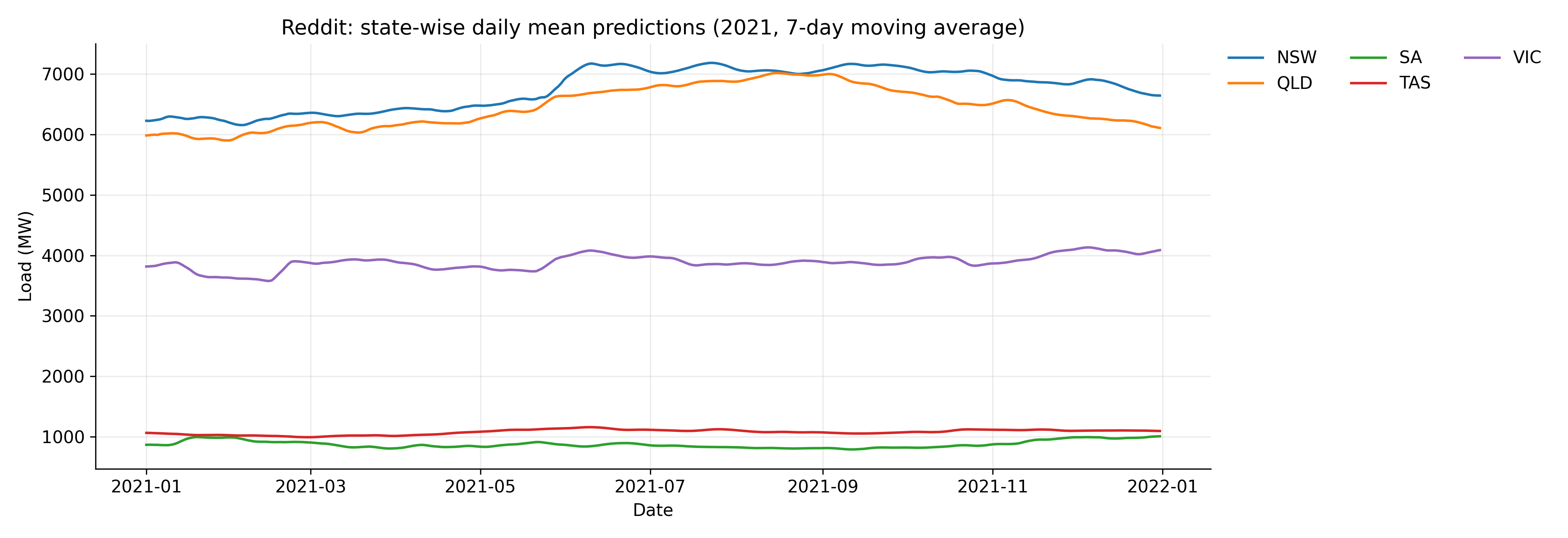}
        \caption{Reddit only.}
        \label{fig:vert-trend-reddit}
    \end{subfigure}
    \hfill
    \begin{subfigure}[b]{0.48\textwidth}
        \centering
        \includegraphics[width=\textwidth]{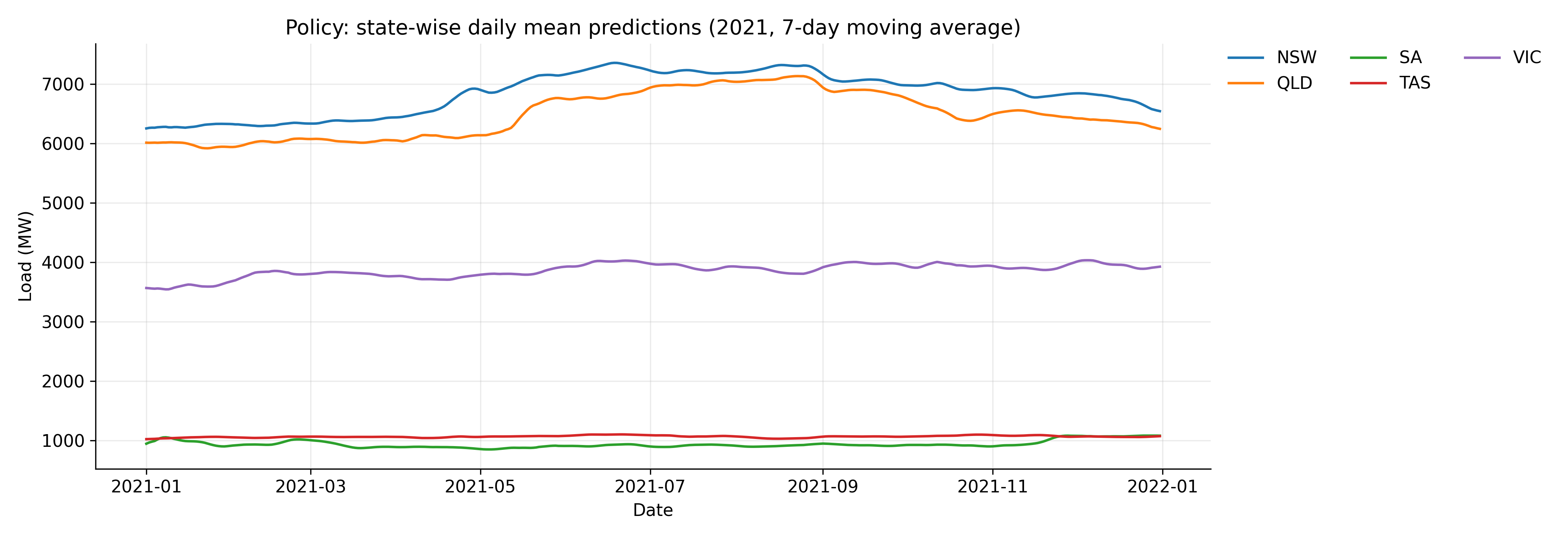}
        \caption{Policy only.}
        \label{fig:vert-trend-policy}
    \end{subfigure}

    \vspace{0.8em}

    \begin{subfigure}[b]{0.48\textwidth}
        \centering
        \includegraphics[width=\textwidth]{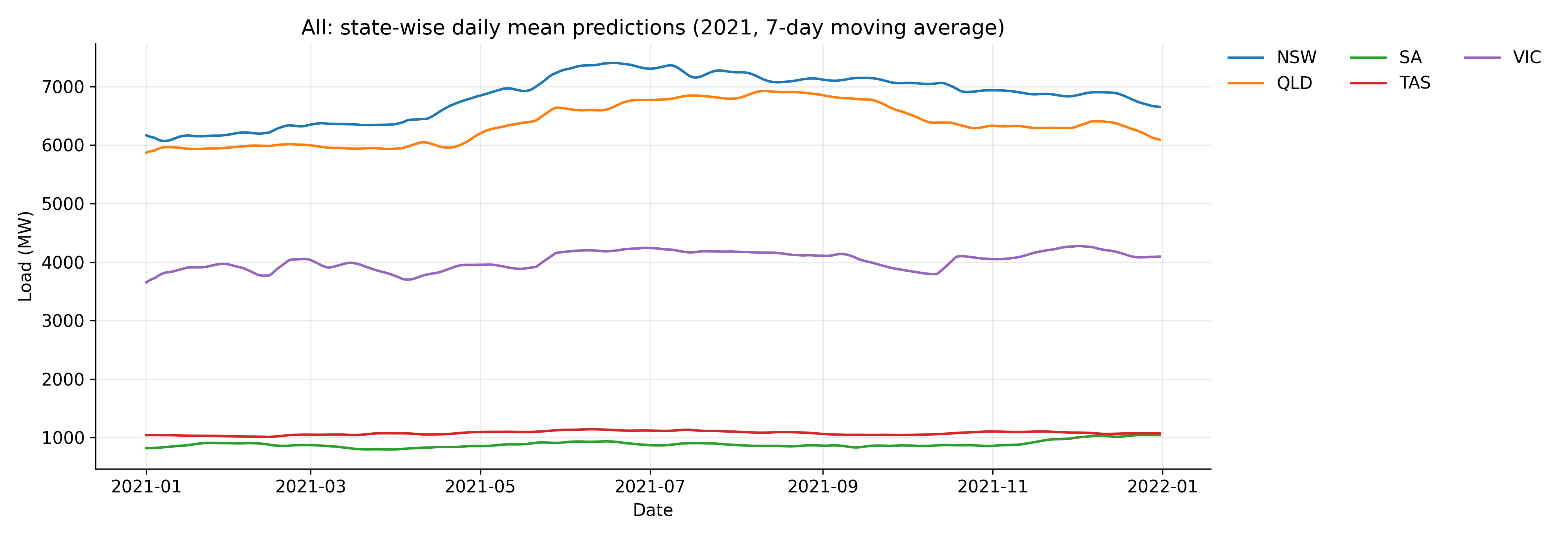}
        \caption{All (News + Reddit + Policy).}
        \label{fig:vert-trend-all}
    \end{subfigure}

    \caption{State-wise daily mean load predictions under different external text configurations in 2021 (7-day moving average). Each panel shows the predicted trajectories for five NEM regions (NSW, QLD, SA, TAS, VIC) for a given text-source configuration.}
    \label{fig:appendix-vertical-trends}
\end{figure*}

\FloatBarrier

\section*{Appendix B. Theoretical Results and Proof Sketches for STanHop}
\addcontentsline{toc}{section}{Appendix B. Theoretical Results and Proof Sketches for STanHop}

This appendix provides the mathematical justification for the three theorems (Theorems~1--3) in Section~3.2 of the main text:
\begin{itemize}
  \item Theorem~1 in the main text (EntMax--conjugate-gradient correspondence);
  \item Theorem~2 in the main text (Lyapunov monotonicity and convergence);
  \item Theorem~3 in the main text (retrieval error bounds and capacity trend).
\end{itemize}

Define the probability simplex
\[
    \Delta_M = \Bigl\{p \in \mathbb{R}^M_{\ge 0} \,\big|\, \sum_{\mu=1}^M p_\mu = 1 \Bigr\},
\]
and the Tsallis $\alpha$-entropy
\begin{equation}
\Psi_\alpha(p) =
\begin{cases}
\dfrac{1}{\alpha(\alpha-1)}\displaystyle\sum_{\mu=1}^M\bigl(p_\mu - p_\mu^\alpha\bigr), & \alpha \neq 1,\\[0.75em]
-\displaystyle\sum_{\mu=1}^M p_\mu \ln p_\mu, & \alpha = 1,
\end{cases}
\end{equation}
with convex conjugate denoted by $\Psi_\alpha^\star$.
Let the memory matrix be $\Xi = [\xi_1,\dots,\xi_M]\in\mathbb{R}^{d\times M}$,
the current state be $x\in\mathbb{R}^d$, and the inverse temperature be $\beta>0$.
The GSH energy function is
\begin{equation}
    H(x) = - \Psi_\alpha^\star\!\bigl(\beta \Xi^\top x \bigr)
    + \frac{1}{2}\langle x,x\rangle + C,
\end{equation}
where $C$ is independent of $x$.
The sparse allocation mapping $\alpha$-EntMax is defined as
\begin{equation}
    \alpha\text{-EntMax}(z)
    = \arg\max_{p\in\Delta_M}\,\bigl[\langle p,z\rangle - \Psi_\alpha(p)\bigr].
\end{equation}

\subsection*{B.1 EntMax--Conjugate-Gradient Correspondence (Main-Text Theorem~1; from STanHop Lemma~3.1)}

\textbf{Lemma B.1 (EntMax--conjugate-gradient correspondence).}
For any $z\in\mathbb{R}^M$, we have
\begin{equation}
    \nabla \Psi_\alpha^\star(z)
    = \alpha\text{-EntMax}(z).
\end{equation}

\emph{Proof sketch.}
Let
\[
    F(p,z) := \langle p,z\rangle - \Psi_\alpha(p),\quad p\in\Delta_M.
\]
Since the Tsallis entropy is continuous and bounded on $\Delta_M$,
$F$ is continuous on the compact set $\Delta_M\times\mathbb{R}^M$.
For any fixed $p$, $F(p,\cdot)$ is affine in $z$, hence convex in $z$.
Therefore, the convex conjugate can be written as
\[
    \Psi_\alpha^\star(z) = \max_{p\in\Delta_M} F(p,z).
\]
If the variational problem
\[
    p^\star(z)
    = \arg\max_{p\in\Delta_M} F(p,z)
    = \alpha\text{-EntMax}(z)
\]
has a unique maximizer, then by Danskin's theorem,
the gradient of $\Psi_\alpha^\star$ with respect to $z$ equals
the gradient of the inner objective at the optimum:
\[
    \nabla_z \Psi_\alpha^\star(z)
    = \nabla_z F(p^\star(z),z)
    = p^\star(z)
    = \alpha\text{-EntMax}(z).
\]
This proves the claim.\hfill$\square$

This lemma shows that mapping a score vector $z$ to a sparse probability distribution via $\alpha$-EntMax
can be viewed as the gradient of the convex conjugate of the Tsallis entropy,
which underpins the unified ``energy function + gradient-based retrieval'' formulation used later.

\subsection*{B.2 Lyapunov Monotonicity and Convergence (Main-Text Theorem~2; from STanHop Lemmas~3.2 and~3.3)}

Using Lemma~B.1, the one-step retrieval iteration of GSH can be written as
\begin{equation}
    x_{t+1} = T(x_t)
    = \nabla_x \Psi_\alpha^\star\!\bigl(\beta \Xi^\top x_t\bigr)
    = \alpha\text{-EntMax}\!\bigl(\beta \Xi^\top x_t\bigr),
\end{equation}
which corresponds to Theorem~2 in the main text.

\textbf{Lemma B.2 (Lyapunov monotonicity; corresponding to STanHop Lemma~3.2).}
For the energy function $H(x)$ and the iteration $T$ defined above, along the sequence $\{x_t\}$,
\begin{equation}
    H(x_{t+1}) \le H(x_t), \quad \forall t,
\end{equation}
i.e., $H$ is monotonically non-increasing during retrieval.

\emph{Proof sketch.}
Using
\[
    H(x) = -\Psi_\alpha^\star\!\bigl(\beta \Xi^\top x\bigr)
    + \frac{1}{2}\langle x,x\rangle + C,
\]
a first-order approximation around $x_t$ gives
\[
    H(x_{t+1}) - H(x_t)
    \approx \Bigl\langle \nabla_x H(x_t),\, x_{t+1}-x_t\Bigr\rangle
    + \frac{1}{2}\|x_{t+1}-x_t\|^2.
\]
Expanding $\nabla_x H$ and substituting
$x_{t+1} = \alpha\text{-EntMax}(\beta \Xi^\top x_t)$,
the difference can be rewritten as a Bregman divergence term
associated with the Tsallis entropy and its convex conjugate.
Its non-negativity implies $H(x_{t+1})\le H(x_t)$.
A complete derivation follows the argument for Lemma~3.2 in the original STanHop paper.

\textbf{Lemma B.3 (Convergence to stationary points; corresponding to STanHop Lemma~3.3).}
Let $\{x_t\}_{t=0}^\infty$ be generated by the iteration $x_{t+1}=T(x_t)$.
Then every limit point of this sequence is a stationary point of the energy function $H$.

\emph{Proof sketch.}
By Lemma~B.2, $\{H(x_t)\}$ is monotonically non-increasing and bounded below, hence convergent.
Combining this with Zangwill's global convergence theorem,
one can verify that the retrieval map $T$ satisfies the closed-graph property and compactness conditions,
which ensures any limit point is a (generalized) fixed point of $T$.
Using an auxiliary lemma establishing the equivalence between generalized fixed points and energy stationary points
(corresponding to Lemma~C.1 in the STanHop appendix),
we conclude that all limit points are stationary points of $H$,
providing a rigorous basis for the statement in Theorem~2 of the main text:
``if the limit exists, it is an energy stationary point.''\hfill$\square$

\subsection*{B.3 Retrieval Error Bounds and Capacity Trend (Main-Text Theorem~3; from STanHop Theorem~3.1 and related results)}

To characterize the error and capacity properties of sparse Hopfield retrieval,
define the pattern separation radius
\begin{equation}
    R_{\min}
    = \frac{1}{2}\min_{\mu,\nu}\|\xi_\mu - \xi_\nu\|_2,
\end{equation}
and the basin of attraction for each pattern
\begin{equation}
    S_\mu := \bigl\{x\in\mathbb{R}^d \ \big|\ 
    \text{the retrieval iteration }T \text{ starting from } x \text{ converges to } \xi_\mu\bigr\}.
\end{equation}

\textbf{Theorem B.1 (Retrieval error bound; corresponding to STanHop Theorem~3.1).}
Let $T_{\mathrm{dense}}$ be the retrieval operator of the dense modern Hopfield model,
and let $T$ be the GSH retrieval operator.
For any $x\in S_\mu$, we have
\begin{equation}
    \|T(x)-\xi_\mu\|_2 \le \|T_{\mathrm{dense}}(x)-\xi_\mu\|_2,
\end{equation}
i.e., under the same target pattern and initialization, the one-step retrieval error of the sparse model
is no larger than that of the dense model.

Moreover, under suitable parameter regimes (e.g., $1\le\alpha\le 2$ or $\alpha\ge 2$),
one can derive exponential or sublinear upper bounds for $\|T(x)-\xi_\mu\|_2$
in terms of the similarity $\langle\xi_\mu,x\rangle$
and the inter-pattern correlation $\max_\nu\langle\xi_\mu,\xi_\nu\rangle$,
indicating that stronger pattern separation and stronger sparsity lead to smaller retrieval errors.

\emph{Proof sketch.}
The key step is to express the sparse retrieval operator as a thresholded $\alpha$-EntMax weighted sum
and compare it with the softmax-form dense retrieval.
Exploiting the differences between EntMax and softmax in tail truncation and temperature scaling,
one can show that, for the same memory set and score vector,
sparse retrieval assigns higher probability mass to dominant components and lower mass to irrelevant patterns.
This yields an $\ell_2$-distance upper bound to the target pattern $\xi_\mu$
that is no worse than that of the dense model.
More detailed derivations can be found in the original proof of STanHop Theorem~3.1.

When noise is introduced via $x\mapsto x+\eta$ or $\xi_\mu\mapsto\xi_\mu+\eta$,
the above bounds further imply a robustness result:
for EntMax-based sparse retrieval, the error grows linearly with the noise norm,
whereas for dense retrieval (or when $\alpha$ lies in certain regimes),
the error may grow exponentially.
This explains the stability advantage of sparse Hopfield models under high-noise conditions.

\textbf{Corollary B.1 (Capacity trend; corresponding to the capacity lower bound in STanHop).}
Let the success probability of storage and retrieval be $1-p$.
Given the dimension $d$, radius $m$, separation radius $R_{\min}$, and temperature $\beta$,
the number of patterns $M$ that GSH can store and reliably retrieve admits a lower bound of the form
\begin{equation}
    M \;\gtrsim\; \sqrt{p}\;C\, d^{-1/2},
\end{equation}
where the constant $C$ is determined by an equation involving the Lambert-$W$ function;
see Lemma~3.4 in the original STanHop paper for the explicit expression.
This result indicates that, when pattern separation and noise level are fixed,
as the dimension $d$ increases, the capacity of sparse modern Hopfield models grows at least at a polynomial rate,
in contrast to the linear capacity of the classical Hopfield model.

\end{document}